\documentclass[iicol,pdflatex,sn-nature]{sn-jnl}
\usepackage{rotating}
\usepackage{graphicx}
\usepackage{multirow}
\usepackage{amsmath,amssymb,amsfonts}
\usepackage{amsthm}
\usepackage{mathrsfs}
\usepackage[title]{appendix}
\usepackage{textcomp}
\usepackage{manyfoot}
\usepackage{booktabs}
\usepackage{algorithm}
\usepackage{algorithmicx}
\usepackage{algpseudocode}
\usepackage{listings}
\usepackage{stfloats}
\usepackage{etoolbox}
\usepackage{arydshln}
\usepackage{hhline}
\usepackage[table]{xcolor}
\usepackage{makecell}

\AtBeginDocument{\setlength{\parindent}{2em}\setlength{\parskip}{0pt}}
\theoremstyle{thmstyleone}

\theoremstyle{thmstyletwo}

\theoremstyle{thmstylethree}

\definecolor{tblhead}{RGB}{206,203,186}   
\definecolor{tblsub}{RGB}{235,233,224}    
\definecolor{tblalt}{RGB}{244,243,238}    
\definecolor{tblrule}{RGB}{140,136,120}   
\begin{document}

\title{A generative foundation model for computational brain imaging}


\author[1]{\fnm{Yuting} \sur{He}}\email{yuting.he4@case.edu}
\equalcont{These authors contributed equally to this work.}

\author[3]{\fnm{Xu} \sur{Wen}}\email{wenxuyyds666@163.com}
\equalcont{These authors contributed equally to this work.}

\author[3]{\fnm{Hongxing} \sur{Lan}}\email{220252361@seu.edu.cn}

\author[7]{\fnm{Chengliang} \sur{Liu}}\email{liucl1996@163.com}

\author[8]{\fnm{Yuxing} \sur{Liu}}\email{liuyuxin264@gmail.com}

\author[3]{\fnm{Yan} \sur{Xi}}\email{yanxi@first-imaging.com}

\author[3]{\fnm{Wenxue} \sur{Yu}}\email{ywx.list@seu.edu.cn}

\author[2]{\fnm{Jindi} \sur{Kong}}\email{jindi.kong@case.edu}


\author[5]{\fnm{Zhiqiang} \sur{Zhang}}\email{zhangzq2001@126.com}

\author*[6]{\fnm{Liang} \sur{Jiang}}\email{jiangliang0402@163.com}

\author*[4]{\fnm{Rongjun} \sur{Ge}}\email{rongjun\_ge@seu.edu.cn}

\author*[3]{\fnm{Yang} \sur{Chen}}\email{chenyang.list@seu.edu.cn}

\author[\empty]{Multi-Center Cooperative Clinical Study Group for ``Lesion-Phenotype Brain Mapping in Acute Stroke''}

\affil[1]{Department of Biomedical Engineering, Case Western Reserve University, Cleveland, Ohio, USA}
\affil[2]{Department of Computer and Data Sciences, Case Western Reserve University, Cleveland, Ohio, USA}
\affil[3]{School of Computer Science and Engineering, Southeast University, Nanjing, China}
\affil[4]{School of Instrument Science and Engineering, Southeast University, Nanjing, China}
\affil[5]{Department of Radiology, Affiliated Jinling Hospital, Medical School of Nanjing University, Nanjing, China}
\affil[6]{Department of Radiology, Nanjing First Hospital, Nanjing Medical University, Nanjing, China}
\affil[7]{Department of Computer and Information Science, University of Macau, Macau, China}
\affil[8]{Department of Electronic and Electrical Engineering, Southern University of Science and Technology, Shenzhen, China}

\abstract{
Computational brain imaging is constrained by a mismatch between increasingly diverse tasks and scarce, heterogeneous clinical data. Although existing brain foundation models transfer pretrained representations, their knowledge remains entangled within model-specific features and fixed architectures. This paper proposes BrainGFM, a generative foundation model that extends knowledge transfer beyond representation reuse through a reusable and reconfigurable brain-imaging prior. By pretraining on a corpus of about 1.3 million MRI and CT slices, BrainGFM supported task-aligned data synthesis for low-data predictive learning and task-adapted post-training for image reconstruction. Across 48 downstream tasks, BrainGFM improved segmentation and classification under limited supervision and achieved high-fidelity image enhancement across heterogeneous imaging conditions. Multicenter clinical validation involving 4,796 stroke patients across two internal and four external centers demonstrated its applicability to three clinically relevant tasks around stroke cohorts. Blinded reader assessments showed that BrainGFM-generated FLAIR images approached acquired FLAIR in overall diagnostic quality. In a human-AI collaboration study, BrainGFM assistance yielded a 14\% relative improvement in mean radiologist stroke-subtyping accuracy. These findings support generative priors as a reusable foundation for heterogeneous computational brain-imaging applications.
}




\maketitle

\section{Introduction}\label{sec:introduction}
\begin{figure*}
    \centering
    \includegraphics[width=\linewidth]{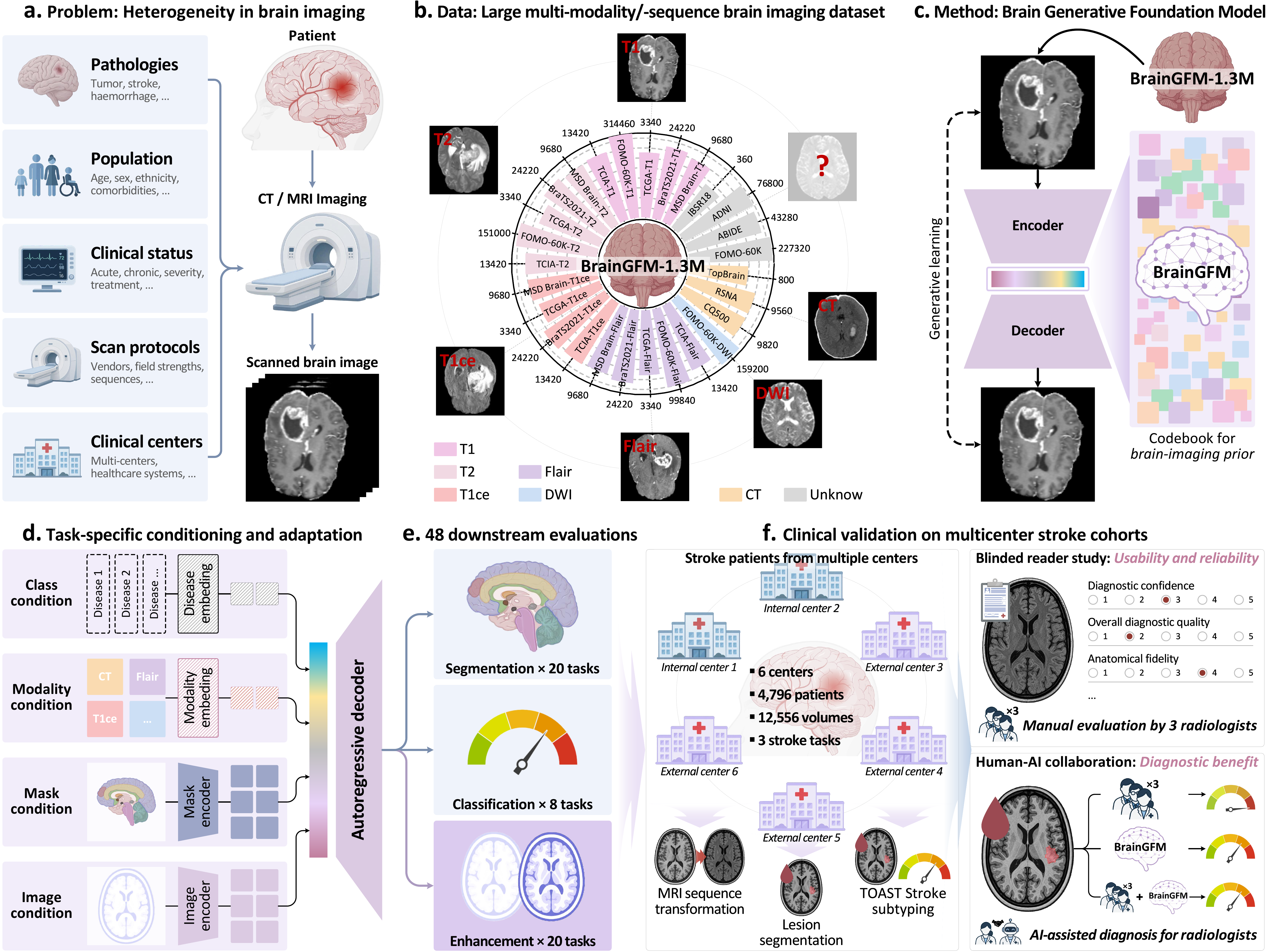}
    \caption{\textbf{Overview of BrainGFM for Computational brain imaging.} \textbf{a.} Heterogeneity in brain diseases, patient populations, clinical status, acquisition protocols, and clinical centers produces heterogeneous brain imaging distributions and limits the generalization of DL models. \textbf{b.} Our BrainGFM-1.3M comprises about 1.3 million axial slices spanning major MRI sequences and CT, providing a large-scale dataset for brain generative pretraining. \textbf{c.} Our BrainGFM learns a reusable brain-imaging prior from generative pretraining on our BrainGFM-1.3M dataset. \textbf{d.} Task-specific conditioning controls the pretrained generator where class labels are used for class-conditioned synthesis, segmentation masks for spatially conditioned synthesis, and source images for brain image reconstruction. These conditions are combined with task-specific adaptation as detailed in Fig.~\ref{fig:method_supp}b,c. \textbf{e.} Evaluation spans 48 downstream tasks, comprising 20 segmentation, 8 classification, and 20 image-enhancement tasks. \textbf{f.} Clinical validation in a multicenter stroke cohort assesses BrainGFM for TOAST stroke subtyping, lesion segmentation, and MRI sequence transformation, as well as its clinical utility in reader studies and radiologist-AI collaboration.}
    \label{fig:fig1}
\end{figure*}

Deep learning (DL) is reshaping computational brain imaging, enabling increasingly accurate models for anatomical segmentation, disease characterization, prognosis, and image enhancement \cite{miller2016multimodal,litjens2017survey,isensee2021nnunet}. However, its further progress is limited by a fundamental imbalance between the growing diversity of brain imaging tasks and the availability of representative data. Large-scale magnetic resonance imaging (MRI) and computed tomography (CT) datasets are costly to assemble, while variation across clinical centers, scan protocols, patient populations, patient clinical status, brain diseases, etc., compromises real-world generalization \cite{willemink2020preparing,cheplygina2019notsosupervised,geirhos2020shortcut} (Fig.~\ref{fig:fig1}a). In parallel, voxel-level annotation requires substantial expert effort, further limiting the amount of task-specific supervision available for model development. Consequently, despite major methodological advances, DL in brain imaging remains constrained by the persistent mismatch between increasingly complex downstream demands and scarce, heterogeneous clinical data \cite{tajbakhsh2016convolutional,raghu2019transfusion}.

Data-efficient learning has emerged as an important response to this bottleneck. Weakly supervised and semi-supervised paradigms reduce reliance on dense annotation by exploiting sparse labels, coarse supervision or unlabeled data \cite{zhou2018wsl,cheplygina2019notsosupervised,adadi2021survey}. These approaches have lowered annotation costs and improved label efficiency in settings where fully supervised training is impractical. However, their success remains conditioned on the scale, diversity, and reliability of the available supervisory signal. When supervision is incomplete, noisy, or weakly aligned with the target task, optimization will be driven toward unstable or misleading objectives, degrading downstream performance \cite{karimi2020noisy}. This vulnerability is particularly consequential in brain imaging, where fine anatomical boundaries, subtle tissue contrast, and heterogeneous pathology leave limited tolerance for erroneous supervision \cite{zhou2018wsl,cheplygina2019notsosupervised}. Moreover, many data-efficient methods remain tied to predefined task formulations and annotation formats, limiting their transferability across imaging protocols, institutions, and clinical applications \cite{cheplygina2019notsosupervised,ouyang2023fewshot}.

These limitations have motivated the development of representation-centered foundation models, which learn transferable knowledge from large-scale data and reuse it through pretrained feature representations \cite{bommasani2021foundation,moor2023generalist,tak2026brainiac,zhou2023foundation}. In computational brain imaging, this paradigm has improved model initialization and enabled label-efficient adaptation across downstream tasks \cite{zhou2023foundation,mazher2026towards,tak2026brainiac,kondepudi2026health}. However, pretrained knowledge is transferred through feature representations embedded within particular model architectures. Consequently, its downstream reuse has to depend on task-specific adaptation of these representations, which constrains the forms in which pretrained knowledge can be exploited when task objectives, supervision requirements, and imaging conditions vary \cite{isensee2021nnunet,karani2018lifelong,tak2026brainiac}. This limitation is amplified in brain imaging, where heterogeneous tasks require different forms of supervision and model functionality, ranging from spatially aligned annotations for segmentation to discriminative information for classification or fine-grained information for reconstruction.

Generative foundation models (GFMs) learn image generative distributions, enabling the generalization of varied downstream requirements in computational brain imaging \cite{ho2020ddpm,rombach2022ldm,zhang2023controlnet,wang2025minim}. This capacity is particularly relevant to brain imaging, where the brain has relatively stable anatomical organization \cite{evans19933d,willemink2020preparing,shinohara2014intensity}. Therefore, with large-scale generative pretraining on brain images, GFMs will learn a brain-imaging prior that describes brain organization and is able to generate diverse brain images. By adapting a brain-imaging prior to task-specific conditions \cite{zhang2023controlnet,wang2025minim}, pretrained knowledge can be expressed as synthetic image-mask pairs or class-conditioned images for predictive learning, while the imaging prior itself in GFMs can be adapted for image reconstruction. This enables pretrained brain-imaging knowledge to be reused in task-specific forms beyond architecture-bound feature representations, allowing more flexible adaptation across heterogeneous computational tasks.

In this paper, we propose BrainGFM, a generative foundation model that converts large-scale brain imaging pretraining into reusable generative capabilities for computational brain imaging (Fig.~\ref{fig:fig1}b,c). BrainGFM is pretrained for brain imaging using the BrainGFM-1.3M corpus, comprising 1.3 million axial slices from 64,754 three-dimensional (3D) MRI and CT volumes. The corpus spans six brain imaging modalities or sequences, including T1, T2, contrast-enhanced T1 (T1ce), FLAIR, DWI, and CT, and covers heterogeneous situations, e.g., neurodevelopmental disorders, brain tumours, intracranial haemorrhage, vessels, etc. Large-scale generative pretraining enables BrainGFM to learn a reusable brain-imaging prior that supports two complementary forms of downstream transfer. 1) For predictive tasks, this prior can be expressed as task-aligned synthetic data, enabling scarce supervision to be amplified into effective training data for segmentation and classification. 2) For brain image reconstruction tasks, the brain-imaging prior can be adapted through parameter-efficient post-training to support image enhancement and transformation. BrainGFM transfers pretrained brain-imaging knowledge either through generated task-specific data or through direct model adaptation, providing a unified foundation for heterogeneous brain predictive and image-generation tasks.

We conducted two levels of validation (Fig.~\ref{fig:fig1}d-f) to evaluate the methodological breadth and clinical applicability of our BrainGFM. First, we evaluated BrainGFM across 48 downstream tasks spanning brain image segmentation, classification, and enhancement, assessing its generality and performance advantages through task-aligned data synthesis for low-data predictive learning and task-adapted post-training for brain image reconstruction (Fig.~\ref{fig:fig1}d,e). Across heterogeneous datasets, imaging modalities, and task objectives, BrainGFM improved segmentation and classification performance under limited supervision, while achieving high-fidelity image enhancement. Second, we assessed its real-world clinical applicability in a multicenter stroke-imaging study involving 4,796 patients from six clinical centers, covering lesion segmentation, stroke subtyping, and MRI sequence transformation, complemented by a blinded reader study and radiologist-AI collaborative diagnosis (Fig.~\ref{fig:fig1}f). These evaluations provide broad empirical evidence that BrainGFM can transfer a shared brain-imaging prior across diverse brain computational tasks while supporting clinically relevant applications in heterogeneous real-world settings.

\section{Results}\label{sec:results}
\subsection{Overview of brain generative foundation model (BrainGFM)}\label{sec:overview}
We constructed BrainGFM-1.3M, a large multi-modality / -sequence brain imaging corpus, to support the generative pretraining of our BrainGFM (Fig.~\ref{fig:fig1}b). The corpus consists of 64,754 3D volumes collected from 11 public resources, comprising T1, T2, T1ce, FLAIR, DWI, and CT, and covering heterogeneous situations. For each volume, empty extremal slices were removed, and 20 uniformly spaced axial slices were sampled within the brain extent of each volume, yielding 1,295,080 slices. Skull stripping was performed with HD-BET for MRI images \cite{isensee2019hdbet} and SynthStrip for CT images \cite{hoopes2022synthstrip}, followed by brain-region-aware intensity normalization \cite{shinohara2014intensity,nyul2000standardization} and canvas-based standardization to $256 \times 256$. Dataset construction and preprocessing are detailed in Sec.~\ref{sec:methods_dataset}. Dataset composition and preprocessing are detailed in Sec.~\ref{sec:methods_dataset} and Supplementary Tab.~\ref{tab:supp_pretraining_sources} and~\ref{tab:supp_pretraining_distribution}.

BrainGFM was pretrained on BrainGFM-1.3M using two optimized components (Fig.~\ref{fig:method_supp}a and Sec.~\ref{sec:methods_pretraining}). First, a multi-scale vector-quantized visual tokenizer learned to represent brain images and segmentation masks within a shared discrete visual vocabulary \cite{vandenOord2017vqvae,razavi2019vqvae2,esser2021vqgan}. The learned representations convert the knowledge in the pretraining corpus into discrete token sequences ordered from lower to higher spatial resolution, achieving a brain-imaging prior. Second, an autoregressive transformer \cite{tian2024var} learned to predict these discrete representations from coarse to fine, modeling brain-image distributions under modality- and sequence-specific conditions. The conditional generative model distinguished T1, T2, T1ce, FLAIR, DWI, and CT images, as well as segmentation masks and images without standardized sequence information. These capabilities enabled the pretrained model to generate different brain-image appearances and spatial structures, providing the foundation for task-aligned data synthesis and task-adapted image reconstruction. The architecture and pretraining objectives are detailed in Sec.~\ref{sec:methods_pretraining}.

\subsection{Task-aligned data synthesis enables low-data predictive learning}\label{sec:results_predictive}
\begin{figure*}[!t]
    \centering
    \includegraphics[width=1\linewidth]{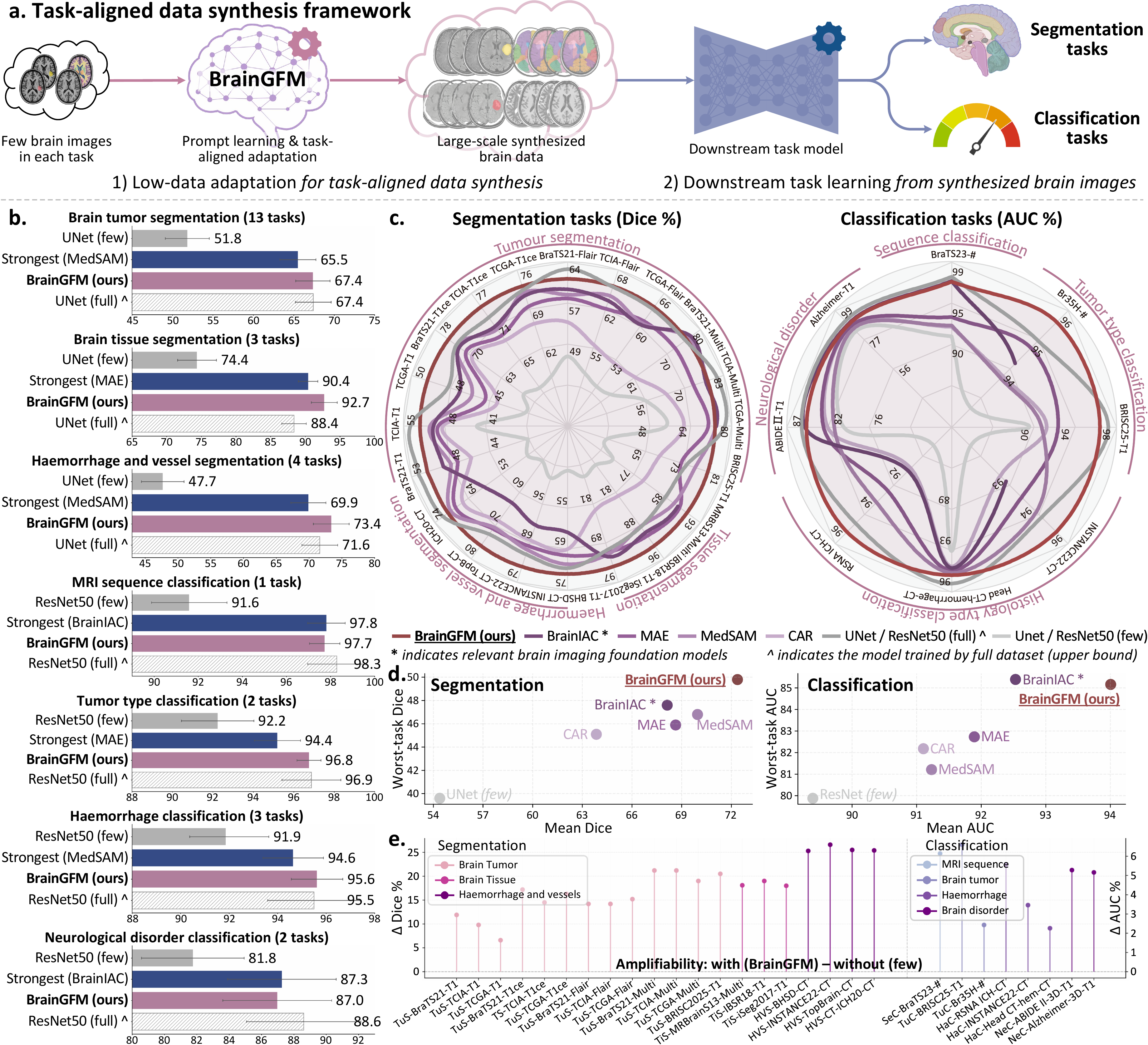}
    \caption{\textbf{Task-aligned data synthesis enables low-data predictive learning.} \textbf{a.} Overview of the task-aligned data synthesis framework. BrainGFM is adapted with few brain images and synthesizes large-scale task-aligned data to train downstream segmentation and classification models. \textbf{b.} Task-family-level performance. BrainGFM is compared with direct low-data training, the strongest competing low-data method, and full-data training references. BrainGFM achieves strong Dice and AUC across heterogeneous prediction targets. \textbf{c.} Performance across 20 segmentation and 8 classification tasks. BrainGFM maintains strong performance across heterogeneous datasets, imaging modalities, and prediction targets. \textbf{d.} Task robustness measured by mean and worst-task performance. BrainGFM achieves a favourable balance between overall and lower-bound performance, indicating its robustness and broad applicability. \textbf{e.} Amplifiability of low-data supervision. The lollipop plot shows BrainGFM's performance gain compared with direct low-data training (with $-$ without BrainGFM), measured as $\Delta$Dice or $\Delta$AUC. BrainGFM achieves positive gains across all tasks, demonstrating its powerful amplifiability for the value of few supervision. Superscript $\wedge$ denotes models trained using the full dataset, and superscript $*$ denotes brain-imaging foundation models.}
    \label{fig:fig2}
\end{figure*}

BrainGFM is able to convert its pretrained generative prior into task-specific training data under limited supervision (Fig.~\ref{fig:fig2}a). The evaluation comprised 28 predictive tasks, including 20 segmentation and eight classification tasks. These tasks require different forms of supervision: segmentation relies on spatially aligned image-mask pairs, whereas classification depends on images carrying discriminative class-specific characteristics. BrainGFM was adapted using one annotated case for each segmentation task or one case per class for each classification task, with 20 axial slices used for adaptation in volumetric settings. It then generated synthetic image-mask pairs or class-conditioned images to train downstream predictors. By evaluating the resulting models against low-data training and pretrained competing approaches, we assessed whether generative synthesis could amplify scarce annotations across distinct predictive objectives. The adaptation procedures are detailed in Sec.~\ref{sec:methods_synthesis}.

\subsubsection{Segmentation learning through image-mask synthesis}\label{sec:results_segmentation}
We examined BrainGFM's ability to convert few brain images and annotations into effective supervision for segmentation. The evaluation covered 20 tasks spanning brain-tumour segmentation (TuS, 13 tasks), brain-tissue segmentation (TiS, three tasks), and intracranial haemorrhage and vessel segmentation tasks (HVS, four tasks) (Supplementary Tab.~\ref{tab:segmentation_multidataset}).  For each task, BrainGFM was adapted using 20 annotated slices from one target case and generated 2,000 synthetic image-mask pairs to train a downstream UNet. The comparison included low-data training, generative and representation-based methods, and a full-data UNet reference. Performance was assessed using Dice, with the experimental protocol and comparison methods described in Secs.~\ref{sec:methods_synthesis} and~\ref{sec:experimental_comparison}.

Across all 20 tasks, BrainGFM achieved a mean Dice score of 72.4\%, compared with 54.4\% for direct low-data training with UNet (few) and 70.0\% for MedSAM, the strongest competing method on average (Fig.~\ref{fig:fig2}b-d). BrainGFM ranked first among the competing methods on 18 of the 20 tasks and achieved an average margin of 2.0\% over the strongest competing result. Its mean Dice score also exceeded that of UNet (full) (71.4\%), and it surpassed the full-data reference on 14 tasks despite using annotations from only one target case. The profiles further showed that this advantage was broadly distributed across datasets and imaging settings. Consistent with this observation, BrainGFM combined the highest mean Dice with the strongest worst-task performance among the evaluated methods, placing it in the favourable upper-right region of the mean-versus-worst-task analysis.

The improvements were observed across all three segmentation task families. For TuS tasks, BrainGFM achieved a mean Dice score of 67.4\% across 13 tasks, compared with 51.8\% for UNet (few), 65.5\% for MedSAM, and 67.4\% for UNet (full) (Fig.~\ref{fig:fig2}b). It ranked first among the competing methods on 11 tasks and exceeded the full-data reference on eight, demonstrating that one-case adaptation supported image-mask synthesis across heterogeneous tumour cohorts, MRI sequences, and input configurations, including both single-sequence and multimodal settings. For TiS tasks, BrainGFM achieved a mean Dice score of 92.7\%, compared with 74.4\% for UNet (few), 90.4\% for MAE and 88.4\% for UNet (full). It ranked first and exceeded the full-data reference on all three tasks, indicating that the pretrained generative prior captured spatial information relevant to distributed tissue organization and anatomical boundaries. For HVS tasks, BrainGFM achieved a mean Dice score of 73.4\% across four CT tasks, compared with 47.7\% for UNet (few), 69.9\% for MedSAM, and 71.6\% for UNet (full). It ranked first among the methods on all four tasks and exceeded the full-data reference on three, showing that the same synthesis strategy transferred effectively to CT images with distinct intensity distributions and heterogeneous haemorrhage morphologies.

The per-task gain analysis quantified the amplification of limited supervision (Fig.~\ref{fig:fig2}e). Compared with UNet (few), BrainGFM produced positive Dice improvements on all 20 tasks, with a mean gain of 18.0\%. The gains were present across tumour, tissue and haemorrhage tasks, including tasks with different modalities, MRI sequences, and segmentation targets. This consistency demonstrates broad amplification of limited pixel-level supervision across the segmentation tasks.

These results show that BrainGFM is able to reconfigure a shared brain-imaging prior into task-specific image-mask generators using annotations from only one target case. The resulting synthetic data supported segmentation at multiple spatial and semantic scales, from distributed neuroanatomical tissues and boundaries to focal tumours and heterogeneous haemorrhagic lesions. By approaching or surpassing full-data training on most tasks while improving over direct low-data learning, task-aligned image-mask synthesis provides an effective mechanism for amplifying scarce annotations.

\subsubsection{Classification learning through class-conditioned synthesis}\label{sec:results_classification}
We evaluated BrainGFM's ability to generate discriminative class-conditioned training data across eight classification tasks, spanning MRI sequence identification (SeC), brain-tumour classification (TuC), intracranial haemorrhage classification and detection (HaC), and neuroimaging-based disorder classification (NeC) (Supplementary Tab.~\ref{tab:classification_multidataset}). Under a case-level $N$-way 1-shot setting, one target case per class, i.e., 20 slices, was used to adapt BrainGFM, followed by the generation of 1,000 synthetic images per class for downstream classification. Volumetric tasks (ABIDE II and Alzheimer-3D) were evaluated using case-level predictions. Performance was compared with low-data classification, pretrained competing methods, and a full-data ResNet50 reference using AUC, F1, and ACC. The adaptation and evaluation procedures are detailed in Secs.~\ref{sec:methods_synthesis} and~\ref{sec:experimental_metrics}.

Across the eight tasks, BrainGFM achieved the highest mean AUC among the evaluated methods, reaching 94.0\%, compared with 89.4\% for ResNet50 (few), 91.2\% for MedSAM, 91.9\% for MAE, and 91.1\% for CAR (Fig.~\ref{fig:fig2}b-d). Across the seven tasks on which BrainIAC was applicable, BrainGFM achieved a mean AUC of 93.6\%, compared with 92.5\% for BrainIAC. BrainGFM ranked first among the low-data methods on five tasks and second in the remaining three, placing it among the two strongest low-data methods in every evaluated task. Its mean AUC also approached the 94.5\% obtained by ResNet50 (full), despite using only one target case per class for adaptation. The performance profiles further showed that this advantage was maintained across heterogeneous modalities, disease categories, prediction targets, and dimensionalities. BrainGFM also combined the highest mean AUC among the low-data methods with a worst-task AUC of 85.2\%, demonstrating strong lower-bound performance across the classification tasks.

The performance gains extended across all four classification families. For SeC task on BraTS2023, BrainGFM achieved an AUC of 97.7\%, compared with 91.6\% for ResNet50 (few), 97.8\% for BrainIAC, and 98.3\% for ResNet50 (full), indicating that generated images preserved sequence-dependent contrast characteristics (Fig.~\ref{fig:fig2}b). For TuC tasks, BrainGFM achieved a mean AUC of 96.8\%, compared with 92.2\% for ResNet50 (few) and 96.9\% for ResNet50 (full). BrainGFM ranked first among the low-data methods on both tasks, demonstrating that class-conditioned synthesis retained tumour-related phenotypic variation across different cohorts. For HaC tasks, BrainGFM achieved a mean AUC of 95.7\%, compared with 91.9\% for ResNet50 (few), 94.6\% for MedSAM, and 95.5\% for ResNet50 (full). For NeC tasks, BrainGFM achieved a mean AUC of 87.0\%, matching BrainIAC while exceeding all other competing methods. The strong performance on these tasks indicates that the generated samples retained spatially distributed imaging characteristics associated with neurodevelopmental and neurodegenerative disorders.

The per-task gain analysis directly quantified the amplification of limited class-level supervision (Fig.~\ref{fig:fig2}e). Compared with ResNet50 (few), BrainGFM produced positive AUC gains on all eight tasks, with a mean improvement of 4.6\%. These gains were observed across MRI and CT, binary and multi-class prediction, and both 2D and 3D settings. This consistency demonstrates that class-conditioned synthesis amplifies limited supervision across heterogeneous classification targets.

These results show that BrainGFM can reconfigure a shared brain-imaging prior into task-specific class-conditioned generators using only one support case per class. The resulting synthetic data retained discriminative information at multiple semantic scales, ranging from acquisition-related sequence contrast and focal tumour or haemorrhage phenotypes to spatially distributed disorder-related patterns. By achieving top-two performance among the low-data methods on every task, approaching the overall full-data reference, and improving over direct low-data training, class-conditioned synthesis provides an effective mechanism for amplifying scarce supervision in brain image classification.

\subsection{Task-adapted post-training enables high-fidelity brain image reconstruction}\label{sec:results_reconstruction}
\begin{figure*}
    \centering
    \includegraphics[width=1\linewidth]{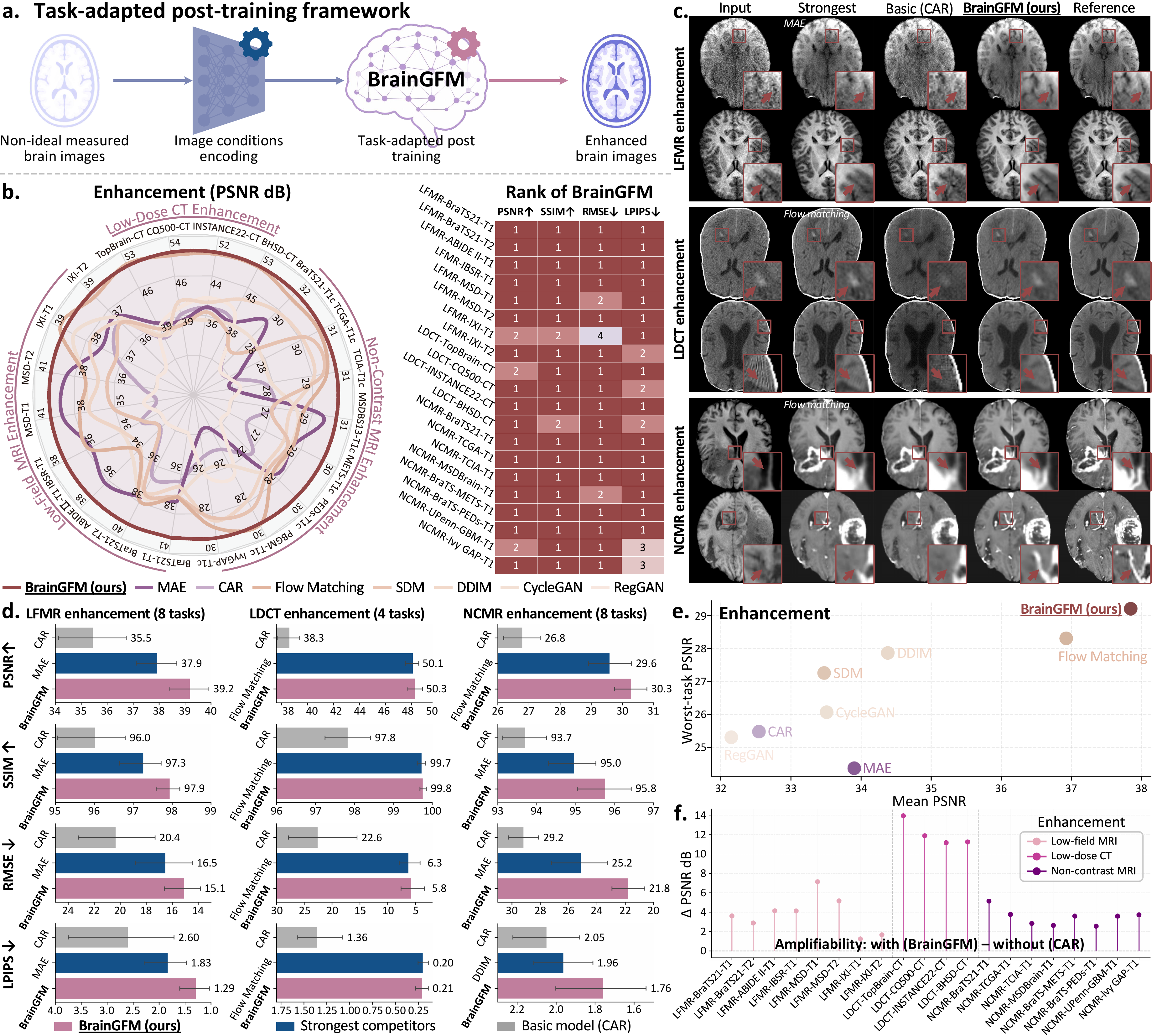}
    \caption{\textbf{Task-adapted post-training enables high-fidelity brain-image enhancement.} \textbf{a.} Overview of the task-adapted post-training framework. BrainGFM is adapted with paired task-specific brain images and directly enhances non-ideal measurements to enhanced brain images. \textbf{b.} The radar plot shows PSNR across tasks, and the heat map reports BrainGFM's rank for PSNR, SSIM, RMSE, and LPIPS. BrainGFM maintains strong performance across heterogeneous datasets, modalities, and enhancement targets. \textbf{c.} Representative qualitative comparisons. BrainGFM reduces image degradation while preserving anatomical structures and local details. \textbf{d.} Task-family-level performance. BrainGFM outperforms CAR and the strongest competitors across the three task families and four evaluation metrics. \textbf{e.} Robustness between mean and worst-task PSNR. BrainGFM achieves a favourable balance between overall and lower-bound performance. \textbf{f.} Amplifiability of task-adapted post-training. The lollipop plot reports BrainGFM's PSNR gain over CAR as $\Delta$PSNR. Positive gains across all tasks demonstrate that task-adapted post-training amplifies the enhancement capability of the model without brain-image pretraining.}
    \label{fig:fig3}
\end{figure*}

BrainGFM extends its pretrained brain-imaging prior to heterogeneous image reconstruction through task-adapted post-training (Fig.~\ref{fig:fig3}a and Sec.~\ref{sec:methods_posttraining}). The evaluation comprised 20 enhancement tasks
across three families, including low-field brain MRI enhancement (LFMR), low-dose CT enhancement (LDCT), and non-contrast MRI enhancement (NCMR). These tasks involve distinct reconstruction requirements across degraded MRI signals, reduction of CT acquisition noise, and generation of contrast-enhanced MRI appearances from non-contrast images. They provide complementary settings for assessing the transferability of the pretrained brain-imaging prior. The evaluation use paired task-specific source and reference images (Supplementary Tab.~\ref{tab:imaging_multidataset}). BrainGFM was compared with established adversarial, masked-image-modeling, diffusion, flow-based, and autoregressive approaches. Reconstruction performance was assessed using PSNR, SSIM, RMSE, and LPIPS. Implementation, comparison methods, and metric definitions are detailed in Secs.~\ref{sec:experimental_implementation}, \ref{sec:experimental_comparison}, and~\ref{sec:experimental_metrics}.

\paragraph{Low-field brain MRI enhancement} BrainGFM achieved consistent enhancement across heterogeneous LFMR conditions. We evaluated eight simulated LFMR tasks covering both T1- and T2-weighted images. Paired low- and high-quality images were synthesized using a k-space-based degradation protocol, in which complex Gaussian noise was added to the k-space representation of each reference image, followed by inverse Fourier transformation and magnitude reconstruction \cite{lin2023low}. The original images were retained as reconstruction targets. Across the eight tasks, BrainGFM achieved 39.2\,dB PSNR, 97.9\% SSIM, 15.1 RMSE, and 1.29 LPIPS, outperforming the strongest competitor MAE across all four metrics (Fig.~\ref{fig:fig3}d). BrainGFM ranked first in 27 of the 32 task-metric comparisons and achieved the best results across all four metrics on five tasks (Fig.~\ref{fig:fig3}b). These results demonstrate stable enhancement across heterogeneous datasets and MRI contrasts.

Task-adapted post-training amplified LFMR restoration and improved visual fidelity. BrainGFM increased PSNR over CAR on all eight tasks, with an average gain of 3.8\,dB and individual gains ranging from 1.2 to 7.1\,dB (Fig.~\ref{fig:fig3}f). Representative results showed reduced image noise and clearer tissue boundaries with improved correspondence to the reference anatomy (Fig.~\ref{fig:fig3}c). The consistent quantitative and visual improvements support effective transfer of the pretrained brain-imaging prior to low-SNR MRI restoration.

\paragraph{Low-dose brain CT enhancement} BrainGFM achieved strong and consistent restoration across heterogeneous LDCT conditions. We evaluated four simulated LDCT tasks using paired low-dose and reference CT images, with low-dose degradation simulated by introducing dose-dependent Poisson noise in the image domain. The original CT images were retained as reconstruction targets. BrainGFM achieved 50.3\,dB, 99.8\% SSIM, 5.8 RMSE, and 0.21 LPIPS on average, exceeding Flow Matching on three task-family-level metrics (Fig.~\ref{fig:fig3}d). At the task level, BrainGFM ranked first in 12 of the 16 task-metric comparisons and remained within the top two in all 16 comparisons (Fig.~\ref{fig:fig3}b). These results demonstrate stable restoration across different CT cohorts and degradation conditions.

Task-adapted post-training amplified the LDCT enhancement capability of the underlying CAR model. BrainGFM improved PSNR over CAR on all four tasks, with an average improvement of 12.1\,dB and individual gains ranging from 11.2 to 13.9\,dB (Fig.~\ref{fig:fig3}f). The gains reached 13.9\,dB on TopBrain, 11.9\,dB on CQ500, 11.2\,dB on INSTANCE22, and 11.2\,dB on BHSD. Representative images showed suppression of low-dose degradation and improved delineation of anatomical structures with closer correspondence to the reference CT images (Fig.~\ref{fig:fig3}c). These findings demonstrate strong adaptation of the pretrained brain-imaging prior to low-dose CT restoration.

\paragraph{Non-contrast brain MRI enhancement} BrainGFM generated contrast-enhanced T1-weighted MRI appearances from non-contrast T1-weighted inputs, achieving high agreement with paired reference images. We evaluated eight NCMR tasks constructed from paired T1 and T1ce images, with non-contrast T1 images as inputs and the corresponding T1ce images as reconstruction targets. BrainGFM achieved 30.3\,dB PSNR, 95.8\% SSIM, 21.8 RMSE, and 1.76 LPIPS on average (Fig.~\ref{fig:fig3}d). Compared with the strongest competitor for each metric, BrainGFM improved PSNR by 0.7\,dB over Flow Matching and SSIM by 0.8\% over MAE, while reducing RMSE by 3.3 relative to MAE and LPIPS by 0.20 to DDIM. BrainGFM ranked first in 28 of the 32 task-metric comparisons, including the highest SSIM on all eight tasks, the best PSNR and RMSE on seven tasks, and the best LPIPS on six tasks (Fig.~\ref{fig:fig3}b). These results demonstrate effective recovery of contrast-related image information across heterogeneous cohorts.

Task-adapted post-training amplified NCMR enhancement across all eight tasks. BrainGFM achieved positive PSNR gains over CAR on all tasks, with an average improvement of 3.5\,dB and individual gains ranging from 2.6 to 5.1\,dB (Fig.~\ref{fig:fig3}f). Representative results showed improved recovery of contrast-enhancement patterns while preserving anatomical correspondence with the reference T1ce images (Fig.~\ref{fig:fig3}c). These findings show that the pretrained brain-imaging prior can be adapted to infer subject-specific contrast-enhanced MRI appearances from non-contrast acquisitions.

\paragraph{Cross-task robustness} 
BrainGFM maintained the strongest overall and lower-bound performance across the complete enhancement task suite. Across all 20 tasks, BrainGFM achieved a mean PSNR of 37.8\,dB and a worst-task PSNR of 29.2\,dB, compared with 36.9\,dB and 28.3\,dB for Flow Matching, the strongest competing method in the cross-task analysis (Fig.~\ref{fig:fig3}e). BrainGFM occupied the favourable upper-right region of the mean-versus-worst-task space, indicating stable enhancement performance across heterogeneous imaging conditions. These results demonstrate that task-adapted post-training consistently transfers the pretrained brain-imaging prior across MRI restoration, CT restoration, and contrast-enhanced MRI generation.

\subsection{Clinical validation on real-world multicenter stroke cohorts}\label{sec:results_clinical}
We evaluated the clinical applicability of BrainGFM in multicenter stroke imaging through three analyses in 2 internal (Jinling Hospital (HLH) and Nanjing First Hospital (NFH)) and 4 external clinical (Nanjing Jiangning Hospital (NJH), Wuxi People’s Hospital (WXPH), Northern Jiangsu People’s Hospital (NJPH), and Changzhou Second People’s Hospital (CZSPH)) cohorts. First, multicenter validation assessed BrainGFM across MRI sequence transformation, lesion segmentation, and TOAST stroke subtyping (5 categories). Second, a blinded reader study evaluated the clinical quality of BrainGFM-generated images. Third, a human-AI collaboration study evaluated the contribution of BrainGFM to radiologist diagnosis.

\subsubsection{Multicenter validation across stroke cohorts} \label{sec:multicenter_threetasks}
\begin{table}[!t]
    \centering
    \resizebox{\linewidth}{!}
    {
    \begin{tabular}{clcc}
         \hline
         \rowcolor{tblhead}
         & 
         & \textbf{Internal cohort}
         & \textbf{External cohort} \\
         \rowcolor{tblhead}
         \multirow{-2}{*}{\textbf{Task}}
         & \multirow{-2}{*}{\textbf{Method}}
         & $_{\text{JLH, NFH}}$
         & $_{\text{NJH, WXPH, NJPH, CZSPH}}$ \\
         \hline
         \cellcolor{tblsub}& \cellcolor{tblalt}UNet (few)                 & \cellcolor{tblalt}58.0            & \cellcolor{tblalt}50.8 \\
         \cellcolor{tblsub}& \cellcolor{tblalt}\textit{60 cases}          & \cellcolor{tblalt}(57.3-58.7)    & \cellcolor{tblalt}(48.3-53.3) \\
         \cellcolor{tblsub}& CAR                                          & 69.6                              & 67.2 \\
         \cellcolor{tblsub}&                                              & (68.8-70.3)                      & (66.4-67.9) \\
         \cellcolor{tblsub}& \cellcolor{tblalt}MedSAM                     & \cellcolor{tblalt}72.0            & \cellcolor{tblalt}70.3 \\
         \cellcolor{tblsub}& \cellcolor{tblalt}                           & \cellcolor{tblalt}(71.2-72.9)    & \cellcolor{tblalt}(68.2-72.5) \\
         \cellcolor{tblsub}& MAE                                          & 70.5                              & 69.9 \\
         \cellcolor{tblsub}&                                              & (69.6-71.3)                      & (67.6-72.2) \\
         \cellcolor{tblsub}& \cellcolor{tblalt}BrainIAC                   & \cellcolor{tblalt}69.9            & \cellcolor{tblalt}70.6 \\
         \cellcolor{tblsub}& \cellcolor{tblalt}                           & \cellcolor{tblalt}(69.2-70.6)    & \cellcolor{tblalt}(68.4-72.8) \\
         \cellcolor{tblsub}& \textbf{BrainGFM}                            & \textbf{78.1}                     & \textbf{76.9} \\
         \cellcolor{tblsub}&                                              & \textbf{(77.4-78.8)}             & \textbf{(74.7-79.0)} \\
         \cdashline{2-4}[0.8pt/2pt]
         \cellcolor{tblsub}& \cellcolor{tblalt}{\color{gray}UNet (full)}  & \cellcolor{tblalt} {\color{gray}69.3}   & \cellcolor{tblalt} {\color{gray}65.1} \\
         \cellcolor{tblsub}
         \multirow{-14}{*}{\rotatebox{90}{\textbf{\makecell{Lesion segmentation\\ Dice (\%) $\uparrow$}}}}
         & \cellcolor{tblalt}\color{gray}{\textit{200 cases}}             & \cellcolor{tblalt}\color{gray}{(68.5-70.2)}    & \cellcolor{tblalt}\color{gray}{(62.7-67.6)} \\
         \hline
         \cellcolor{tblsub}& \cellcolor{tblalt}ResNet50 (few)             & \cellcolor{tblalt}62.6              & \cellcolor{tblalt}61.6 \\
         \cellcolor{tblsub}& \cellcolor{tblalt}\textit{60 cases}          & \cellcolor{tblalt}(58.8-66.4)       & \cellcolor{tblalt}(58.4-64.8) \\
         \cellcolor{tblsub}& CAR                                          & 81.5                                & 77.5 \\
         \cellcolor{tblsub}&                                              & (79.7-83.4)                         & (74.7-80.3) \\
         \cellcolor{tblsub}& \cellcolor{tblalt}MedSAM                     & \cellcolor{tblalt}78.4              & \cellcolor{tblalt}70.8 \\
         \cellcolor{tblsub}& \cellcolor{tblalt}                           & \cellcolor{tblalt}(75.4-81.4)       & \cellcolor{tblalt}(67.8-73.7) \\
         \cellcolor{tblsub}& MAE                                          & 82.5                                & 79.6 \\
         \cellcolor{tblsub}&                                              & (79.9-85.1)                         & (76.1-83.1) \\
         \cellcolor{tblsub}& \cellcolor{tblalt}BrainIAC                   & \cellcolor{tblalt}79.7              & \cellcolor{tblalt}74.7 \\
         \cellcolor{tblsub}& \cellcolor{tblalt}                           & \cellcolor{tblalt}(77.6-81.7)       & \cellcolor{tblalt}(72.4-77.0) \\
         \cellcolor{tblsub}& \textbf{BrainGFM}                            & \textbf{87.4}                       & \textbf{ 84.5} \\
         \cellcolor{tblsub}&                                              & \textbf{(85.0-89.8)}                & \textbf{(82.3-86.6)} \\
         \cdashline{2-4}[0.8pt/2pt]
         \cellcolor{tblsub}& \cellcolor{tblalt}{\color{gray}ResNet50 (full)}   & \cellcolor{tblalt}\color{gray}{77.5}         & \cellcolor{tblalt} \color{gray}{75.6} \\
         \cellcolor{tblsub}
         \multirow{-14}{*}{\rotatebox{90}{\textbf{\makecell{TOAST stroke subtyping\\ AUC (\%) $\uparrow$}}}}
         & \cellcolor{tblalt}\color{gray}{\textit{200 cases}}             & \cellcolor{tblalt}\color{gray}{(75.2-79.8)} & \cellcolor{tblalt}\color{gray}{(72.4-78.9)} \\
         \hline
         \cellcolor{tblsub}& \cellcolor{tblalt}RegGAN                     & \cellcolor{tblalt}23.95              & \cellcolor{tblalt}22.92 \\
         \cellcolor{tblsub}& \cellcolor{tblalt}                           & \cellcolor{tblalt}(21.24-26.66)      & \cellcolor{tblalt}(20.40-25.44) \\
         \cellcolor{tblsub}& CycleGAN                                     & 24.14                                & 23.29 \\
         \cellcolor{tblsub}&                                              & (20.48-27.80)                        & (19.34-27.24) \\
         \cellcolor{tblsub}& \cellcolor{tblalt}MAE                        & \cellcolor{tblalt}25.34              & \cellcolor{tblalt}24.52 \\
         \cellcolor{tblsub}& \cellcolor{tblalt}                           & \cellcolor{tblalt}(22.07-28.61)      & \cellcolor{tblalt}(21.02-28.02) \\
         \cellcolor{tblsub}& DDIM                                         & 18.42                                & 17.12 \\
         \cellcolor{tblsub}&                                              & (15.74-21.10)                        & (14.39-19.83) \\
         \cellcolor{tblsub}& \cellcolor{tblalt}SDM                        & \cellcolor{tblalt}23.35              & \cellcolor{tblalt}21.55 \\
         \cellcolor{tblsub}& \cellcolor{tblalt}                           & \cellcolor{tblalt}(19.79-26.90)      & \cellcolor{tblalt}(17.87-25.21) \\
         \cellcolor{tblsub}& Flow Matching                                & 23.63                                & 24.10 \\
         \cellcolor{tblsub}&                                              & (20.34-26.92)                        & (20.74-27.46) \\
         \cellcolor{tblsub}& \cellcolor{tblalt}CAR                        & \cellcolor{tblalt}21.35              & \cellcolor{tblalt}20.83 \\
         \cellcolor{tblsub}& \cellcolor{tblalt}                           & \cellcolor{tblalt}(18.72-23.99)      & \cellcolor{tblalt}(18.25-23.41) \\
         \cellcolor{tblsub}& \textbf{BrainGFM}                            & \textbf{26.63}                       & \textbf{26.90} \\
         \multirow{-16}{*}{\rotatebox{90}{\textbf{\makecell{MRI sequence transformation\\ PSNR (dB) $\uparrow$}}}}
         \cellcolor{tblsub}&                                              & \textbf{(23.83-29.43)}               & \textbf{(23.93-29.87)} \\
         \hline
    \end{tabular}
    }
    \caption{\textbf{Multicenter validation across stroke cohorts.} Performance was evaluated on pooled eligible cases from the two internal centers and four external centers for MRI sequence transformation, lesion segmentation, and TOAST stroke subtyping.}
    \label{tab:stroke_multicenter}
\end{table}

The multicenter evaluation included 4,796 unique patients from six clinical centers, including two internal centers (JLH and NFH) and four external centers (NJH, WXPH, NJPH, and CZSPH) (Tab.~\ref{tab:stroke_cohort}). The evaluation covered clinically relevant tasks around stroke cohorts, i.e., stroke lesion segmentation, TOAST stroke subtyping, and DWI-to-FLAIR MRI sequence transformation. The internal cohorts supported model development and internal evaluation, whereas the external cohorts were reserved for assessing cross-center generalization. For each task, 200 internal cases formed the training cohort. Lesion segmentation and TOAST stroke subtyping used a 60-case low-data setting, with the corresponding 200-case models serving as full-data references. Segmentation was performed on DWI, whereas stroke subtyping incorporated DWI, ADC, and predicted lesion masks. Sequence transformation used paired DWI and FLAIR images to reconstruct FLAIR from DWI. The complete training and validation protocols are described in Sec.~\ref{sec:methods_stroke_protocol}. Performance is reported in Tab.~\ref{tab:stroke_multicenter}, with additional metrics in Supplementary Tab.~\ref{tab:stroke_multicenter_supp}.

BrainGFM maintained strong performance across both lesion analysis and stroke characterization in the multicenter cohorts. For lesion segmentation, BrainGFM achieved Dice scores of 78.1\% (95\% CIs, 77.4-78.8) in the internal cohorts and 76.9\% (74.7-79.0) in the external cohorts. The preservation of performance across centers indicates that the adapted generative prior remained effective for delineating stroke lesions despite differences in clinical populations and imaging environments. This is clinically relevant because reliable lesion localization provides the spatial basis for subsequent assessment of lesion burden and downstream diagnostic analysis. For TOAST stroke subtyping, BrainGFM achieved AUCs of 87.4\% (85.0-89.8) internally and 84.5\% (82.3-86.6) externally. In this task, DWI and ADC were combined with the segmented lesion masks, allowing patient-level classification to incorporate both diffusion characteristics and explicit lesion localization. The strong internal and external performance therefore supports the utility of the BrainGFM-derived lesion representation beyond segmentation itself and indicates that the same framework can contribute to higher-level clinical characterization of stroke.

BrainGFM also showed stable cross-center performance for DWI-to-FLAIR transformation, achieving PSNR values of 26.63\,dB (95\% CI, 23.83-29.43) internally and 26.90\,dB (23.93-29.87) externally, together with SSIM values of 90.7\% and 89.6\%. The consistency between internal and external cohorts indicates that the pretrained generative prior transferred across centers for subject-specific cross-sequence transformation while preserving anatomical correspondence with the source DWI. Together with the segmentation and subtyping results, these findings show that BrainGFM can support three clinically distinct levels of stroke imaging analysis, spanning lesion delineation, patient-level disease characterization, and cross-sequence image transformation. These results demonstrate that BrainGFM can be adapted to three clinical tasks around stroke cohorts. Its performance across four external centers provides evidence of cross-center transferability under heterogeneous stroke-imaging workflows.

\subsubsection{Blinded reader study}\label{sec:results_blinded_reader}
\begin{figure*}[!t]
    \centering
    \includegraphics[width=1\linewidth]{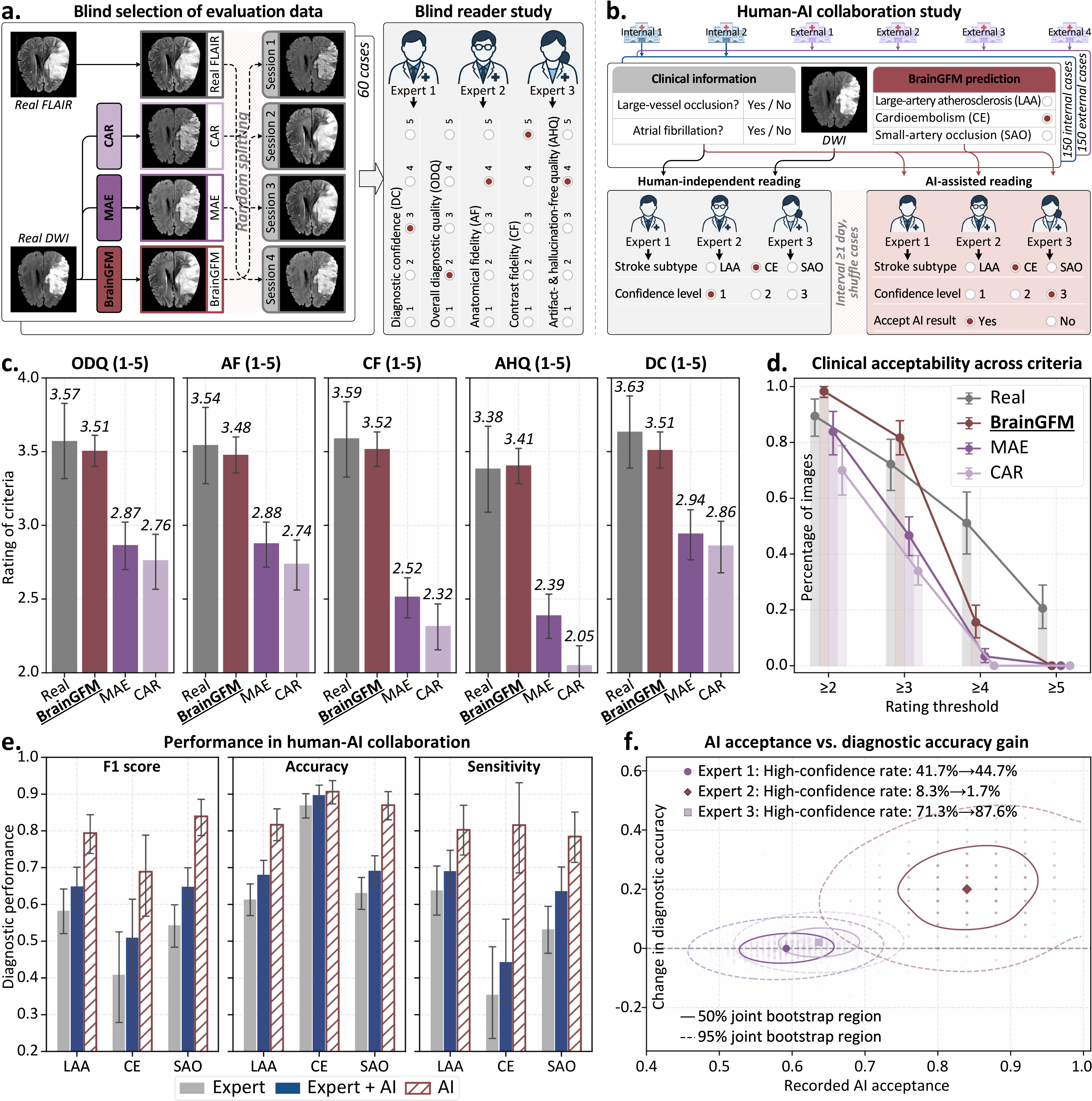}
    \caption{\textbf{Blinded reader study and human-AI collaboration study in stroke cohorts.} \textbf{a-b} Workflow of the blinded reader study for acquired and DWI-synthesized FLAIR images and the human-AI collaboration study for ischemic stroke subtyping. \textbf{c.} Radiologist ratings across five image-quality criteria: overall diagnostic quality (ODQ), anatomical fidelity (AF), FLAIR contrast fidelity (CF), artifact- and hallucination-free quality (AHQ), and diagnostic confidence (DC). \textbf{d.} Proportion of reader assessments meeting all five image-quality criteria at increasing Likert-score thresholds. \textbf{e.} F1 score, one-versus-rest accuracy, and sensitivity for experts, AI-assisted experts, and BrainGFM across LAA, CE, and SAO. \textbf{f.} Relationship between recorded AI recommendation acceptance and changes in diagnostic accuracy among initially high-confidence assessments (confidence score 3). Solid and dashed contours represent the 50\% and 95\% regions of the bootstrap distributions. Error bars in panels \textbf{c-e} indicate 95\% CIs.}
    \label{fig:clinical_reader_study}
\label{fig:clinical_reader_study}
\end{figure*}

We conducted a blinded reader study to evaluate the clinical quality of BrainGFM-generated FLAIR images (Fig.~\ref{fig:clinical_reader_study}a). Three experienced radiologists assessed 60 stroke cases under four imaging conditions, i.e., acquired FLAIR and synthetic FLAIR generated using BrainGFM, MAE, and CAR. Image quality was evaluated across five dimensions covering diagnostic quality, anatomical fidelity, contrast, artifacts and hallucinations, and diagnostic confidence. The blinded reading protocol and five-point Likert scoring procedure \cite{likert1932technique} are described in Sec.~\ref{sec:methods_blinded_reader}, with reader-specific results in Supplementary Tab.~\ref{tab:supp_reader_study}.

Individual quality dimensions revealed the advantages of BrainGFM over competing synthesis methods (Fig.~\ref{fig:clinical_reader_study}c). BrainGFM achieved a mean overall diagnostic quality score of 3.51, compared with 2.87 for MAE and 2.76 for CAR, approaching the score of 3.57 for acquired FLAIR. The largest differences were observed in FLAIR contrast fidelity (3.52 vs. 2.52 and 2.32 for MAE and CAR) and artifact- and hallucination-free quality (3.41 vs. 2.39 and 2.05). BrainGFM also approached acquired FLAIR in anatomical fidelity (3.48 vs. 3.54) and diagnostic confidence (3.51 versus 3.63), with similar artifact-related quality ratings (3.41 vs. 3.38). These results indicate that BrainGFM can produce FLAIR images with substantially improved radiologist-rated quality relative to the evaluated synthesis baselines.

We examined multidimensional image quality by calculating the proportion of reader assessments in which all five quality dimensions met or exceeded higher Likert-score thresholds (Fig.~\ref{fig:clinical_reader_study}d). BrainGFM satisfied all five quality criteria at a threshold of at least 3 in 81.7\% of reader assessments, compared with 72.2\% for acquired FLAIR. However, at the stricter threshold of at least 4, the corresponding proportions were 15.6\% and 51.1\%. Thus, although BrainGFM-generated FLAIR approached acquired FLAIR in mean overall diagnostic quality and performed favourably at moderate quality thresholds, a substantial difference remained at more stringent thresholds. These findings support further evaluation of generated FLAIR as opportunistic imaging information when FLAIR acquisitions are unavailable.

\subsubsection{Human-AI collaboration study}\label{sec:results_human_ai}
We conducted a human-AI collaboration study involving 300 stroke patients to evaluate the contribution of BrainGFM to radiologist diagnosis of ischemic stroke subtypes, including LAA, CE, and SAO (Fig.~\ref{fig:clinical_reader_study}b). Three radiologists assigned stroke subtypes using DWI and clinical information on large-vessel occlusion and atrial fibrillation. Following an interval of more than one day, they reassessed the reshuffled cases with access to BrainGFM subtype predictions. The study compared unaided, AI-assisted, and standalone BrainGFM performance, while examining diagnostic confidence and acceptance of AI recommendations. The complete protocol is described in Sec.~\ref{sec:methods_human_ai}, with reader-specific results in Supplementary Tab.~\ref{tab:human_ai_reader_subtype_performance}.

BrainGFM achieved higher diagnostic accuracy than the participating radiologists and improved their mean performance under AI assistance (Fig.~\ref{fig:clinical_reader_study}e). Standalone BrainGFM achieved an overall accuracy of 79.7\%. With access to its predictions, radiologists' mean accuracy increased from 55.6\% to 63.4\%, corresponding to a 14.0\% relative improvement. Mean F1 score and sensitivity increased from 51.1\% to 60.1\% and from 50.8\% to 59.0\%. However, the benefits varied substantially across readers. Expert 2 showed the largest accuracy improvement, from 53.0\% to 75.0\%, whereas Experts 1 and 3 showed changes from 57.3\% to 56.0\% and from 56.4\% to 59.2\%.

Greater acceptance of AI recommendations was accompanied by both diagnostic benefits and potential risks. When BrainGFM predictions were incorrect, Experts 1, 2, and 3 recorded acceptance of 41.0\%, 82.0\%, and 50.0\% recommendations. For Expert 2, 80 of 102 initially incorrect diagnoses were corrected when BrainGFM was correct, whereas 15 of 22 initially correct diagnoses became incorrect when BrainGFM was wrong. Thus, the reader who benefited most from correct AI predictions was also more susceptible to erroneous recommendations, highlighting the trade-off between diagnostic rescue and inappropriate reliance on AI.

Diagnostic confidence provided insight into reader-specific responses (Fig.~\ref{fig:clinical_reader_study}f). Expert 2 had the lowest initial high-confidence rate (8.3\%) and the highest overall AI recommendation acceptance rate (90.3\%), compared with 41.7\% and 63.0\% for Expert 1, and 71.3\% and 61.3\% for Expert 3. Among initially high-confidence diagnoses (confidence score 3), AI acceptance remained highest for Expert 2 (84.0\%), whose accuracy increased from 52.0\% to 72.0\%. In contrast, Experts 1 and 3 showed smaller changes in accuracy (64.8\% to 64.8\% and 62.2\% to 64.2\%). These exploratory findings suggest that the clinical benefits of AI assistance depend not only on model accuracy but also on how individual radiologists respond to AI recommendations.


\subsection{Ablation study and model analysis}\label{sec:results_ablation}
\begin{figure*}[!t]
\centering
\includegraphics[width=\linewidth]{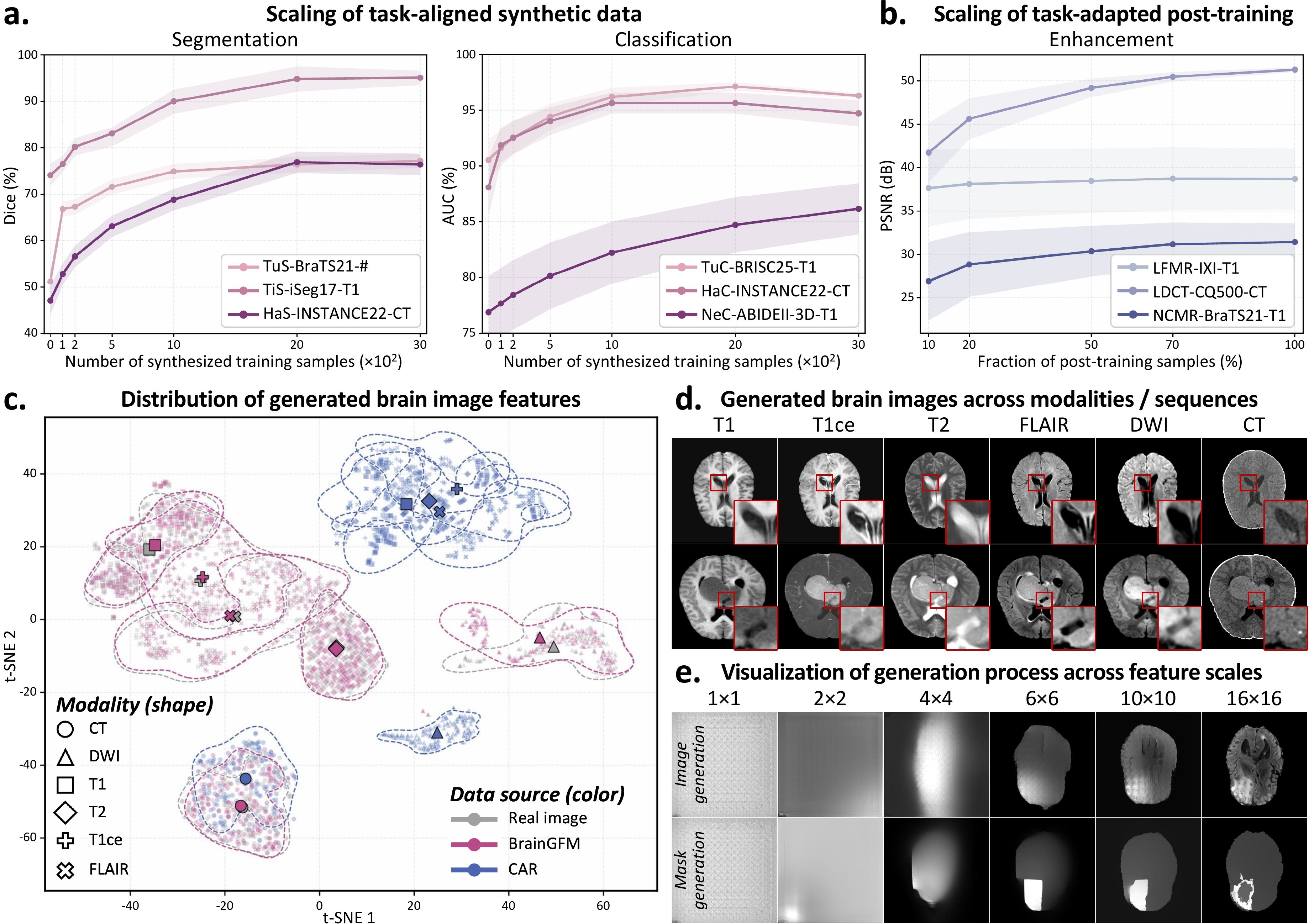}
\caption{\textbf{Ablation study and model analysis of BrainGFM.} \textbf{a.} Synthetic-data scaling across representative segmentation and classification tasks shows consistent performance gains followed by task-dependent saturation, demonstrating that task-aligned synthesis effectively amplifies limited supervision. \textbf{b.} Data scaling across three enhancement tasks shows progressive improvement with increasing post-training data, with task-dependent saturation indicating efficient adaptation from limited paired data. \textbf{c.} t-SNE analysis of real, BrainGFM-generated, and CAR-generated features shows greater overlap between BrainGFM and real-data distributions, supporting improved distribution alignment. \textbf{d.} Representative BrainGFM-generated images across modalities and sequences preserve modality-specific contrast, anatomical structure, and local image details, demonstrating multimodal generation fidelity. \textbf{e.} Intermediate image and mask predictions across successive feature scales reveal progressive formation of global structure and fine spatial details, illustrating the coarse-to-fine next-scale generation process.}
\label{fig:fig4}
\end{figure*}
Task-aligned synthesis showed a clear data-scaling effect and progressively saturated as synthetic supervision increased. We varied the number of synthesized training samples from 0 to 3,000 for three representative segmentation tasks and three classification tasks while keeping the support data and downstream training protocol fixed (Fig.~\ref{fig:fig4}a). For segmentation, Dice increased from 51.2 to 77.2 on TuS-BraTS21, from 74.1 to 95.1 on TiS-iSeg17-T1, and from 47.1 to 76.9 on HVS-INSTANCE22-CT. Classification showed a similar trend, with AUC increasing from 90.54 to 97.14 on TuC-BRISC25-T1, from 88.09 to 95.65 on HaC-INSTANCE22-CT, and from 76.88 to 86.16 on NeC-ABIDEII-3D-T1. Most curves approached a plateau between 1,000 and 3,000 synthesized samples, indicating that task-aligned generation efficiently amplifies limited supervision while exhibiting diminishing returns at larger synthetic-data scales.

Task-adapted post-training exhibited strong data efficiency, with the magnitude of scaling varying across enhancement tasks. We post-trained BrainGFM using 10\%, 20\%, 50\%, 70\%, and 100\% of the paired training data for representative LFMR, LDCT, and NCMR tasks (Fig.~\ref{fig:fig4}b). LFMR performance increased rapidly from 37.64 to 38.11 dB between 10\% and 20\% of the data and subsequently remained near saturation, reaching 38.69 dB with the full training set. LDCT improved from 41.72 to 51.28 dB and NCMR from 26.91 to 31.42 dB as the available data increased. These results show that the pretrained representation can be adapted with limited paired supervision, while more challenging enhancement settings continue to benefit from additional data.

BrainGFM-generated images showed stronger alignment with real brain-image feature distributions across modalities. Features extracted by a fixed MAE encoder were jointly projected with t-SNE for real, BrainGFM-generated, and CAR-generated images from T1, T1ce, T2, FLAIR, DWI, and CT (Fig.~\ref{fig:fig4}c). Across the MRI sequences, BrainGFM features more consistently occupied the real-data regions, with their distribution contours and centroids positioned closer to those of real images than CAR. CT showed substantial overlap among all three sources, indicating a smaller overall domain discrepancy. Nevertheless, the BrainGFM centroid remained more closely aligned with the real-data centroid, suggesting better preservation of the central tendency of the target distribution. The overlap of distribution support and the reduced centroid displacement indicate that BrainGFM captures both the diversity and modality-specific characteristics of real brain images more faithfully.

Generated samples demonstrated that this distributional alignment corresponds to realistic multimodal image content. We visualized representative BrainGFM samples from T1, T1ce, T2, FLAIR, DWI, and CT and enlarged local regions to inspect fine image characteristics (Fig.~\ref{fig:fig4}d). The generated images retained coherent brain anatomy together with sequence-specific tissue contrast, structural boundaries, and local texture patterns. These observations demonstrate that the shared generative representation supports visually consistent synthesis across heterogeneous brain-imaging modalities and sequences.

BrainGFM exhibited a hierarchical mask-to-image generation process across feature scales. We visualized intermediate predictions at $1\times1$, $2\times2$, $4\times4$, $6\times6$, $10\times10$, and $16\times16$ scales (Fig.~\ref{fig:fig4}e). During direct mask generation, coarse scales established the approximate lesion location and extent, while finer scales progressively refined its shape and boundaries. The generated mask was subsequently used as spatial guidance for brain-image generation, where global anatomical and intensity structure emerged first, and local tissue contrast and lesion-associated details were progressively resolved. This coarse-to-fine progression shows that BrainGFM first establishes task-relevant spatial structure and then generates anatomically coherent image content consistent with the synthesized mask.

\section{Discussion}\label{sec:discussion}
A persistent challenge in computational brain imaging is that advances in model capability have not removed the dependence of downstream development on task-specific data. New diseases, imaging conditions, and computational objectives continue to require dedicated cohorts and annotations, making progress difficult in settings where representative supervision is scarce. Although representation-centered foundation models reduce this burden, pretrained knowledge is transferred through model-specific feature representations, which constrain its flexible reuse across heterogeneous downstream requirements \cite{tak2026brainiac}. Our results suggest that this limitation is consequential. Across tasks with different targets and supervision requirements, BrainGFM demonstrated the utility of pretrained knowledge through both task-aligned data synthesis and task-adapted post-training. This indicates that scalable brain-imaging intelligence requires pretrained knowledge to remain reusable after the downstream task and its required form of supervision have changed.

Generative pretraining provides a way to extend pretrained knowledge reuse beyond model-specific feature representations \cite{wang2025minim}. BrainGFM learns a brain-imaging distribution that can be reconfigured through limited supervision and expressed as images, masks, image-mask pairs, or conditioned image transformations. This changes the role of a foundation model in downstream development. The pretrained model can provide task-aligned supervision that is consumed by conventional segmentation or classification models, while the same prior can also be adapted directly for image reconstruction. Therefore, the value of BrainGFM lies in the ability to externalize a shared brain-imaging prior into forms required by different computational tasks. BrainGFM-1.3M provides the multimodal imaging basis for learning this prior, while controllable generation provides the mechanism for converting it into task-specific utility.

The broad performance gains in this study support the generality of this formulation. Task-aligned synthesis improved downstream learning across every evaluated segmentation and classification task under limited supervision, while task-adapted post-training transferred the pretrained prior across MRI restoration, CT restoration, and contrast-related MRI reconstruction. The scaling experiments showed that additional synthetic samples produced progressive gains before reaching task-dependent saturation, indicating that the benefit arises from useful variation represented by the adapted generative distribution. The alignment between generated and real-image feature distributions provides additional evidence that BrainGFM captures image characteristics resembling those of real brain imaging data. Multicenter stroke validation further extends this observation beyond curated tasks by applying the same generative framework to image reconstruction, lesion analysis, and diagnostic classification in clinical cohorts. These findings support generative foundation modeling as a flexible mechanism for transferring one pretrained brain-imaging prior across different downstream requirements.

The clinical evaluations provide complementary evidence for the potential utility of BrainGFM. Multicenter validation demonstrated the transferability of the framework across clinical centers and distinct stroke-imaging tasks. The blinded reader study further showed that BrainGFM-generated FLAIR images achieved favourable perceived diagnostic quality, although their performance at stringent quality thresholds remained below that of acquired FLAIR. In the human--AI collaboration study, access to BrainGFM predictions improved mean radiologist performance, but individual responses to assistance varied. These findings support the clinical relevance of a shared generative brain-imaging prior while highlighting the need for further validation of image interchangeability and the consistency of AI-assisted diagnostic benefits in prospective clinical settings.

This study has several limitations. First, BrainGFM relies predominantly on 2D axial modeling, which limits the extent to which volumetric anatomical structure can be captured during pretraining. Extending the generative prior to 3D modeling may improve the representation of spatially continuous brain anatomy. Second, BrainGFM-1.3M is constructed from existing brain-imaging resources and remains constrained by the population, disease, modality and acquisition coverage represented in these datasets. Expanding the scale and diversity of pretraining data across institutions and clinical settings will strengthen the generality of the learned prior. Third, the current adaptation strategy is based on a fixed set of conditioning signals and parameter-efficient LoRA tuning. More flexible adaptation mechanisms that incorporate richer clinical, acquisition, or task-specific information will improve the efficiency and controllability of downstream transfer. Finally, the scaling results indicate diminishing gains as more synthetic samples are generated, suggesting that future work should improve the diversity and informativeness of generated data. 

BrainGFM suggests a broader role for generative foundation models in computational brain imaging. Large-scale pretraining can provide more than a reusable model initialization when the learned distribution can be adapted and expressed as new task-specific data. This capability offers a route to reuse brain-imaging knowledge across changing diseases, modalities, and computational objectives while reducing the dependence of downstream applications on extensive task-specific data. BrainGFM and BrainGFM-1.3M provide an initial foundation for this direction, in which generative models serve as reusable data-generating infrastructure for the development of computational brain-imaging systems.

\section{Methods}\label{sec:method}
\subsection{BrainGFM-1.3M Dataset Construction and Preprocessing}\label{sec:methods_dataset}
We constructed BrainGFM-1.3M from 11 publicly available or access-controlled, de-identified brain imaging resources, comprising 64,754 3D volumes and 1,295,080 axial slices (Fig.~\ref{fig:fig1}b). The dataset includes ADNI \cite{mueller2005adni}, FOMO-60K \cite{Cerri2026large}, ABIDE II \cite{dimartino2017abide2}, TCGA \cite{weinstein2013tcga}, TCIA \cite{clark2013tcia}, IBSR18 \cite{ibsr18}, MSD Brain \cite{simpson2019msd}, BraTS2021 \cite{bakas2021brats}, CQ500 \cite{chilamkurthy2018cq500}, the RSNA Intracranial Hemorrhage Detection Challenge \cite{flanders2020rsna}, and TopBrain \cite{topbrain_challenge_site}. In total, 63,745 MRI volumes and 1,009 CT volumes were included, covering T1, T2, T1ce, FLAIR, DWI, and CT, together with scans without standardized sequence information. The corpus spans normal brain anatomy, brain tumours, intracranial haemorrhage, neurodevelopmental disorders and neurodegenerative conditions. Before pretraining, a subset of the corpus was reserved as a held-out validation partition and excluded from optimization of BrainGFM during generative pretraining. Samples from this held-out partition were used to construct the downstream evaluations.

All volumes were processed using a unified brain-focused pipeline. Empty slices at the superior and inferior ends of each volume were first removed, after which 20 axial slices were uniformly sampled within the brain-containing extent. Skull stripping was performed using HD-BET for MRI images \cite{isensee2019hdbet} and SynthStrip for CT images \cite{hoopes2022synthstrip}. To avoid interpolation-induced alterations of anatomical boundaries and image intensities, no interpolation-based resizing was applied. Instead, we cropped redundant background according to the brain extent and placed the retained image on a $256 \times 256$ canvas using zero-padding where necessary. Corresponding segmentation masks underwent the same spatial transformation to preserve pixel-level alignment.

A modality-specific intensity normalization strategy was adopted for MRI and CT images. For MRI images, Z-score normalization was performed using the mean and standard deviation calculated within the brain foreground of each case. For CT images, Z-score normalization was performed using the corresponding dataset-level mean and standard deviation. The normalized intensities of both modalities were subsequently rescaled to $[-1,1]$ using predefined intensity bounds, ensuring a consistent numerical range for generative pretraining. For datasets containing pixel-level anatomical or pathological annotations, including brain-tumour, intracranial-haemorrhage, and brain-tissue labels, the corresponding segmentation masks were retained alongside the brain foreground masks. For datasets without lesion annotations, such as neurodevelopmental cohorts, only the brain foreground masks were used.

\begin{figure*}
    \centering
    \includegraphics[width=1\linewidth]{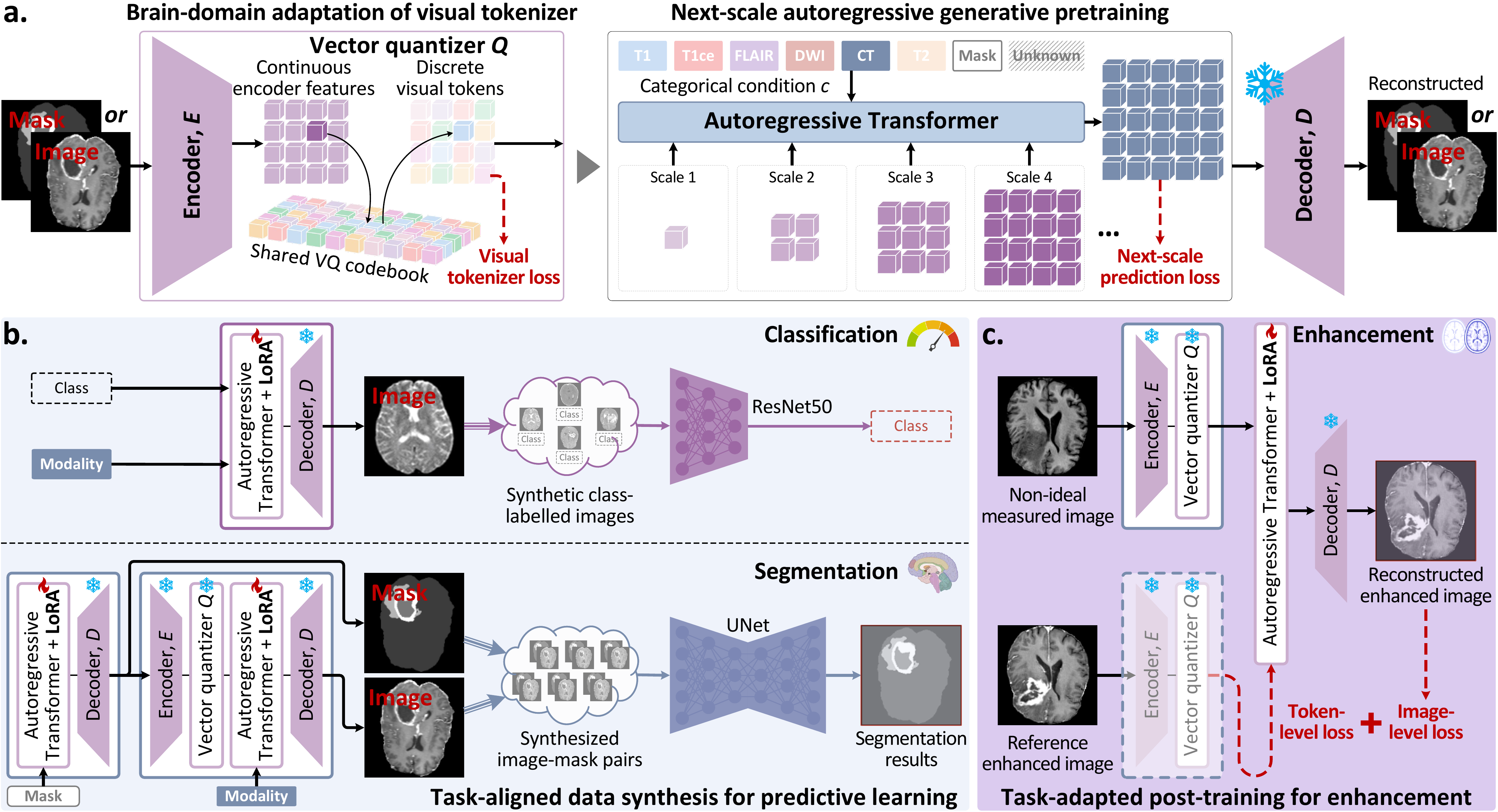}
    \caption{\textbf{BrainGFM architecture, generative pretraining and downstream adaptation.} \textbf{a.} BrainGFM pretraining consists of brain-domain adaptation of a multi-scale visual tokenizer and next-scale autoregressive generative pretraining. A shared encoder ($E$) and vector quantizer ($Q$) map brain images and segmentation masks into multi-scale discrete tokens using a shared visual codebook. The autoregressive Transformer learns the distribution of tokens through coarse-to-fine next-scale prediction under categorical modality and sequence conditions. The pretrained decoder ($D$) reconstructs brain images or segmentation masks from generated tokens. \textbf{b.} Task-aligned data synthesis adapts BrainGFM with limited supervision using LoRA. For classification, task-specific class conditioning enables the synthesis of class-labelled images for training ResNet50. For segmentation, BrainGFM generates task-specific masks and synthesizes corresponding images, producing image-mask pairs for training UNet. \textbf{c.} Task-adapted post-training adapts BrainGFM for image enhancement using paired source-target images. Source-image tokens condition autoregressive prediction of target-image tokens, while LoRA is optimized using token-level prediction and image-level reconstruction losses. }
    \label{fig:method_supp}
\end{figure*}

\subsection{BrainGFM Architecture and Generative Pretraining}\label{sec:methods_pretraining}
BrainGFM was developed to learn a reusable and controllable generative prior from heterogeneous brain imaging data. Generative pretraining consisted of two sequential stages: brain-domain adaptation of a multi-scale vector-quantized visual tokenizer, followed by next-scale autoregressive generative pretraining (Fig.~\ref{fig:method_supp}a). The first stage established a discrete visual representation shared across brain images and masks, whereas the second stage learned their conditional multi-scale distributions under different imaging and structural conditions.

\subsubsection{Brain-domain Adaptation of the Visual Tokenizer}\label{sec:methods_tokenizer}
A multi-scale vector-quantized visual tokenizer initialized from natural-image pretraining was optimized on BrainGFM-1.3M. This stage adapted the pretrained discrete representation to brain-specific anatomical structures, pathological appearances, and imaging characteristics before autoregressive generative pretraining. Brain images and their corresponding segmentation masks were represented within a shared discrete visual vocabulary. Single-channel masks were replicated across three channels so that images and masks could be processed using the same encoder and codebook. Given a brain image $x$ and its corresponding mask $m$, the shared encoder $E$ and vector quantizer $Q$ produced two hierarchies of multi-scale discrete tokens:
\begin{equation}
\begin{aligned}
\{r_k^{x}\}_{k=1}^{K} &= Q(E(x)),\\
\{r_k^{m}\}_{k=1}^{K} &= Q(E(m)).
\end{aligned}
\label{eq:vq_tokens}
\end{equation}

The tokenizer used a codebook containing 4,096 entries with a latent dimension of 32. Ten token scales, $\{1,2,3,4,5,6,8,10,13,16\}$, were used to represent brain images at progressively increasing spatial resolutions. Lower-resolution token maps captured coarse anatomical organization, whereas progressively finer token maps represented local tissue boundaries, structural details, and pathological appearances. During brain-domain adaptation, the tokenizer was jointly optimized using paired brain images and masks with reconstruction, perceptual, vector-quantization, and adversarial objectives \cite{esser2021taming}:
\begin{equation}
\mathcal{L}_{\mathrm{tok}}=\mathcal{L}_{\mathrm{rec}}+\lambda_{\mathrm{perc}}\mathcal{L}_{\mathrm{perc}}  +\lambda_{\mathrm{VQ}}\mathcal{L}_{\mathrm{VQ}}+\lambda_{\mathrm{adv}}\mathcal{L}_{\mathrm{adv}}.
\label{eq:tokenizer_loss}
\end{equation}
Here, $\mathcal{L}_{\mathrm{rec}}$ denotes the reconstruction objective, $\mathcal{L}_{\mathrm{perc}}$ denotes perceptual supervision, $\mathcal{L}_{\mathrm{VQ}}$ denotes the vector-quantization objective, and $\mathcal{L}_{\mathrm{adv}}$ denotes adversarial supervision. The reconstruction, perceptual, and vector-quantization terms were jointly computed for the image and its paired mask, with LPIPS used for perceptual supervision \cite{zhang2018unreasonable}. After convergence, the visual tokenizer was frozen and used throughout subsequent generative pretraining and downstream adaptation.

\subsubsection{Next-scale Autoregressive Generative Pretraining}\label{sec:methods_autoregressive}
The autoregressive Transformer was initialized from pretrained ControlVAR weights \cite{li2024controlvar} and fully optimized on BrainGFM-1.3M. The model learned the joint distribution of brain-image appearance and spatial structure through next-scale prediction \cite{tian2024visual}. For paired images and segmentation masks, the mask and image tokens at each spatial scale were arranged as
\begin{equation}
s_k=\left[r_k^{m};r_k^{x}\right],
\label{eq:paired_tokens}
\end{equation}
and progressively predicted from coarse to fine according to
\begin{equation}
\begin{aligned}
p_{\theta}(s_{1:K}\mid c)=\prod_{k=1}^{K}p_{\theta}\left(s_k\mid s_{<k},c\right),
\end{aligned}
\label{eq:next_scale}
\end{equation}
where $c$ denotes the categorical generation condition. The autoregressive Transformer was optimized using token-level cross-entropy:
\begin{equation}
\begin{aligned}
\mathcal{L}_{\mathrm{AR}}=-\sum_{k=1}^{K}\log p_{\theta}\left(s_k\mid s_{<k},c\right).
\end{aligned}
\label{eq:ar_loss}
\end{equation}
The next-scale formulation established global anatomical organization at coarse resolutions and progressively refined tissue boundaries, local anatomical structures, and pathological appearances at finer resolutions.

The categorical generation conditions included T1, T2, T1ce, FLAIR, DWI, CT, scans for which standardized sequence information was unavailable, and a dedicated mask category. Modality- and sequence-specific conditions enabled the model to learn distinct image distributions within the shared generative representation. The dedicated mask condition enabled segmentation masks to be modelled as generative targets. For datasets containing paired images and segmentation masks, joint image-mask modelling further established correspondence between spatial structure and image appearance.

During mask-conditioned image generation, the input segmentation mask was first encoded by the frozen visual tokenizer into multi-scale mask tokens. At each spatial scale, these mask tokens were treated as observed conditioning tokens and remained fixed, while BrainGFM autoregressively predicted the corresponding image tokens from coarse to fine. This enabled the generated image to follow the spatial structure specified by the input mask. In parallel, the modality or sequence condition was represented by a categorical embedding and injected into the autoregressive Transformer through adaptive layer normalization, specifying the desired image category, such as T1, T2, T1ce, FLAIR, DWI, or CT. Random dropping of the categorical condition during pretraining was used to enable classifier-free guidance at inference.

\subsection{Adaptation of BrainGFM to Downstream Tasks}\label{sec:methods_adaptation}
Downstream benchmarks were constructed from held-out partitions of datasets shared with pretraining and from additional datasets not included in the pretraining corpus. These samples were excluded from generative pretraining and were used only for downstream adaptation and evaluation, preventing sample-level information leakage between foundation-model pretraining and downstream experiments.

BrainGFM was adapted to downstream tasks through two complementary strategies. For predictive tasks, including segmentation and classification, task-aligned data synthesis adapted the pretrained generative prior using limited supervision and generated additional task-specific training samples for downstream predictors (Fig.~\ref{fig:method_supp}b). For brain image reconstruction, task-adapted post-training directly adapted BrainGFM using paired source-target images for image-to-image mapping learning (Fig.~\ref{fig:method_supp}c).

For all downstream adaptations, the pretrained visual tokenizer and the original autoregressive Transformer weights remained frozen. The Transformer was adapted using LoRA \cite{hu2022lora}. LoRA modules were inserted into the query, key, value, and attention output projections of the Transformer, as well as its feed-forward network and adaptive linear layers, with rank $r=8$, scaling factor $\alpha=16$, and dropout of $0.05$. For a pretrained weight matrix $W$, the adapted transformation was defined as
\begin{equation}
W'=W+\Delta W,\ \Delta W=\frac{\alpha}{r}BA,
\label{eq:lora}
\end{equation}
where $A$ and $B$ denote trainable low-rank matrices. Additional task-specific conditioning parameters were optimized when required by the corresponding downstream task.

\subsubsection{Task-aligned Data Synthesis for Predictive Learning}\label{sec:methods_synthesis}
\textbf{Segmentation} used two successive conditioning steps. BrainGFM first generated task-specific segmentation masks using the dedicated mask condition learned during pretraining. Each generated mask was then encoded by the frozen visual tokenizer into multi-scale mask tokens and used as the spatial condition for mask-conditioned image generation. The corresponding modality or sequence condition specified the image type to be generated. For single-sequence tasks, the generated mask was paired with the corresponding MRI-sequence or CT condition. For multimodal MRI tasks, the same generated mask was combined with the required modality conditions to generate spatially aligned multimodal images. For each segmentation task, one annotated case was used for adaptation, from which 20 axial slices were uniformly sampled. The LoRA parameters were optimized using these image-mask pairs to adapt the pretrained generative distribution to the target segmentation task. A total of 2,000 synthetic image-mask pairs were generated for each segmentation task. A standard 2D U-Net \cite{ronneberger2015unet} was trained using these synthetic pairs and evaluated on the corresponding test set.

\textbf{Classification} replaced the pretrained imaging-category condition with task-specific class conditioning. A learnable embedding was introduced for each target class and optimized together with the LoRA parameters, while the pretrained visual tokenizer and Transformer weights remained frozen. For each classification task, one case was selected for each class, and 20 axial slices were uniformly sampled from each case for adaptation. After adaptation, the class embedding was used as the generation condition, and BrainGFM generated 1,000 class-conditioned images for each class. These synthetic images were used to train a ResNet50 classifier \cite{he2016resnet}. For the ABIDE II and Alzheimer-3D tasks, which require prediction at the volume level, consecutive axial slices were organized into case-level inputs for volumetric classification.

\subsubsection{Task-adapted Post-training for Brain Image Reconstruction}\label{sec:methods_posttraining}
For brain image reconstruction, BrainGFM was adapted from conditional image synthesis to paired image-to-image mapping learning. Each training sample consisted of a source image and its corresponding target image. The source image represented the observed image to be enhanced or transformed, whereas the target image represented the desired reconstruction. Source-image conditioning replaced the mask conditioning used for spatially controlled image generation. The frozen visual tokenizer encoded the source image into multi-scale source-image tokens. At each spatial scale, these source-image tokens were provided as conditioning information to the autoregressive Transformer, which predicted the corresponding target-image tokens through next-scale prediction. The target-image tokens were subsequently decoded by the frozen visual tokenizer to reconstruct the target image. In this way, the conditioning signal was changed from a segmentation mask specifying spatial structure to a source brain image specifying subject-specific anatomy and image content.

For each enhancement task, all available paired training samples were used for post-training. The pretrained visual tokenizer and original Transformer weights remained frozen, whereas the LoRA parameters, task-specific condition embedding, and parameters required for source-image conditioning were optimized. The post-training objective combined autoregressive prediction of the target-image tokens with image-level reconstruction:
\begin{equation}
\mathcal{L}_{\mathrm{enh}}=\mathcal{L}_{\mathrm{token}}+\lambda_{1}\mathcal{L}_{1},
\label{eq:enh_loss}
\end{equation}
where $\mathcal{L}_{\mathrm{token}}$ denotes cross-entropy over the target-image tokens and $\mathcal{L}_{1}$ denotes the $L_{1}$ reconstruction loss between the reconstructed target image and its reference.

Task-specific source-target pairs were constructed according to the three enhancement settings. For low-field brain MRI enhancement, each low-quality source image was generated from the corresponding reference MRI image using a k-space-based degradation protocol \cite{lin2023low}. Complex Gaussian noise was added to the k-space representation followed by inverse Fourier transformation and magnitude reconstruction, and the original MRI image was retained as the target image. For low-dose brain CT enhancement, each low-dose source image was generated from the corresponding reference CT image by introducing dose-dependent Poisson noise in the image domain, with the original CT image retained as the reconstruction target. For non-contrast brain MRI enhancement, spatially registered T1 and T1ce images were used directly as paired source-target data, with T1 serving as the source image and T1ce as the target image.

\section*{Data Availability}
The publicly available and access-controlled brain imaging datasets used for BrainGFM pretraining and downstream evaluation were obtained from their original data providers, subject to the respective licenses and data-use agreements. Dataset sources
and access information are provided in Supplementary Tab.~\ref{tab:supp_pretraining_sources}. The multicenter clinical stroke imaging data are subject to institutional ethical and data-governance requirements and are not publicly distributed. Detailed results from the downstream benchmarks, multicenter clinical validation, and reader studies are provided in the Supplementary Information.

\section*{Code Availability}
All source code developed for this study, including BrainGFM generative pretraining, task-aligned data synthesis, task-adapted post-training, data preprocessing, downstream evaluation, and pretrained BrainGFM weights, have been made publicly available at \url{https://github.com/YutingHe-list/BrainGFM}. The repository has also provided the training configurations and scripts required to reproduce the reported experiments and statistical analyses.

\section*{Ethics Approval}
The multicenter clinical validation was conducted in accordance with the Declaration of Helsinki and applicable institutional and regulatory requirements. The study was approved by the institutional review boards of all participating hospitals (No. 2024DZKY-019-01), and informed consent was waived. The blinded reader study and human-AI collaboration study were conducted under the same approved protocol. All clinical data were de-identified before research analysis, and access to patient-level information was restricted to authorized research personnel.

All third-party datasets used for generative pretraining, downstream model adaptation, training, validation, and testing were obtained from their original data providers and used in accordance with the applicable ethical approvals, informed-consent provisions, data-use agreements, licenses, and access conditions established by the respective data providers. The sources and access information for these datasets are detailed in Supplementary Tab.~\ref{tab:supp_pretraining_sources}.

\section*{Competing Interests}
The authors declare no competing interests.

\section*{Author Contributiond}
Y.H., X.W., R.G., and Y.C. conceived and designed the study. Y.H. and X.W. developed the BrainGFM framework and led its implementation and experimental evaluation. H.L., C.L., Y.L., W.Y., and J.K. contributed to data processing, model development, and benchmark experiments. Z.Z. and L.J. contributed to multicenter clinical data collection, clinical study design, and clinical evaluation. Y.X. contributed to the design of imaging experiments and interpretation of imaging results. Y.H. and X.W. performed data analysis, prepared the figures, and drafted the manuscript. L.J., R.G., and Y.C. supervised the study, interpreted the findings, and critically revised the manuscript. All authors reviewed and approved the final manuscript.

\section*{Acknowledgments}
We thank the members of the Multi-Center Cooperative Clinical Study Group for ``Lesion-phenotype brain mapping in acute stroke'': Xindao Yin, phD, Wusheng Zhu, phD, Xiangming Fang, phD, Jing Ye, phD, Changjie Pan, phD, Zhen Jiang, phD, Caifeng Guo, phD, Lulu Xiao, MD, Zhiwen Geng, phD, Zhen Jiang, phD, Zheng Dai, phD, Min Zhang, phD, Zhiliang Guo, MD, Caifeng Guo, MD, Ting Hu, MD.

\bibliography{sn-bibliography}
\appendix
\section{Details of Experiments}
\subsection{Downstream Experimental Protocols}\label{sec:experimental_protocols}
\subsubsection{Implementation and computational resources}
\label{sec:experimental_implementation}
BrainGFM and the downstream experiments were implemented in PyTorch and are implemented on four NVIDIA GeForce RTX 3090 GPUs, each with 24 GB of memory.

\textbf{For generative pretraining}, two stages are implemented. In the first stage, the multi-scale visual tokenizer was initialized from pretrained weights and adapted to brain images and segmentation masks for 20 epochs. Optimization used Adam with $\beta_1=0.5$ and $\beta_2=0.9$, an initial learning rate of $4.5\times10^{-6}$, a weight decay of $1\times10^{-4}$, and linear learning-rate decay. The tokenizer was optimized using reconstruction, perceptual, and adversarial objectives, with an $L_1$ reconstruction loss and LPIPS-based perceptual loss. Adversarial optimization was introduced after 150,000 optimization steps, and the mask-mixing probability was set to 0.02. In the second stage, the autoregressive Transformer was initialized from pretrained ControlVAR weights and trained for 100 epochs using full-parameter optimization with AdamW ($\beta_1=0.9$, $\beta_2=0.95$). The base learning rate was set to $1\times10^{-4}$, with a linear decay-to-zero schedule and a 1\% warmup period, while weight decay decreased from 0.02 to zero. Gradient clipping with a maximum norm of 2.0 and BF16 mixed precision were employed. Both pretraining stages used a per-GPU batch size of two, corresponding to a global batch size of eight without gradient accumulation.

\textbf{For downstream adaptation}, the pretrained visual tokenizer and the original autoregressive Transformer weights were frozen. LoRA modules were introduced into the query, key, value, and attention output projections, as well as the feed-forward network and adaptive linear layers, with rank $r=8$, scaling factor $\alpha=16$, and dropout 0.05. Task-specific conditioning parameters were additionally optimized where required. LoRA adaptation used AdamW ($\beta_1=0.9$, $\beta_2=0.95$) with a weight decay of 0.02, a linear decay-to-zero learning-rate schedule, and a 1\% warmup period. Gradient clipping was applied with a maximum norm of 2.0, and training was conducted using BF16 mixed precision with a per-GPU batch size of eight. Predictive tasks used conditional next-scale token prediction to adapt BrainGFM for segmentation-mask and class-conditioned image synthesis. For brain image reconstruction, the adaptation objective combined target-token prediction with image-level reconstruction, as defined in Eq.~\ref{eq:enh_loss}. 

The resulting synthetic image-mask pairs or class-conditioned images were used to train downstream UNet \cite{ronneberger2015unet} or ResNet50 \cite{he2016resnet} predictors of the generative model. For segmentation, the UNet was trained for up to 100 epochs with a batch size of 128 using RMSprop, a weight decay of $1\times10^{-8}$, and a momentum of 0.999. A ReduceLROnPlateau scheduler was employed in maximization mode with a patience of five epochs, together with gradient clipping at a maximum norm of 1.0 and early stopping. For classification, ResNet50 was initialized from pretrained weights and trained for up to 100 epochs with a batch size of 128 using Adam and a cosine annealing learning-rate schedule, with early stopping applied during training.

\subsubsection{Comparison methods}\label{sec:experimental_comparison}
To evaluate BrainGFM across heterogeneous computational brain-imaging tasks, we compared it with established task-specific models, generative models, and representation-based foundation models under matched experimental settings. Comparisons were organized into three task categories: segmentation, classification, and image enhancement. Within each task, all methods were evaluated on the same test data, using consistent input configurations and evaluation metrics. The low-data and full-data supervised references used the same model architectures and initialization strategies within each task, differing in the amount of available annotated training data.

\textbf{For segmentation}, the comparison methods included UNet \cite{ronneberger2015unet}, CAR \cite{li2024controlvar}, the MedSAM \cite{medsam}, MAE \cite{he2022masked}, and BrainIAC \cite{tak2026brainiac}. UNet served as a conventional supervised segmentation baseline, with UNet (few) and UNet (full) representing low-data and full-data references. Both were trained from scratch using the same standard UNet architecture. CAR is a controllable autoregressive generative model and followed the same synthesis-based prediction framework as BrainGFM, generating synthetic image-mask pairs to train the same downstream UNet architecture. For representation-based methods, MAE was implemented using a TransUNet \cite{chen2021transunet} architecture incorporating a pretrained MAE ViT-B encoder, with the encoder and segmentation decoder jointly fine-tuned. MedSAM employed a pretrained image encoder and mask decoder, both of which were fine-tuned while the prompt encoder remained frozen. BrainIAC provided brain-imaging-specific pretrained representations through a frozen encoder, with a trainable UNet decoder used for segmentation prediction.

\textbf{For classification}, ResNet50 \cite{he2016resnet} was used as the conventional supervised classification baseline, with ResNet50 (few) and ResNet50 (full) serving as the corresponding low-data and full-data references. Both used the same architecture and pretrained initialization strategy. Following the setting in segmentation tasks, CAR generated class-conditioned synthetic images to train the same downstream ResNet50 architecture as BrainGFM. For representation-based comparisons, MAE and MedSAM employed their respective pretrained image encoders together with task-specific classification heads, with the encoders and classification heads jointly fine-tuned. In contrast, BrainIAC used a frozen pretrained encoder for feature extraction, followed by a trainable ResNet-based classification head. These methods represented distinct approaches to transferring pretrained visual or brain-imaging representations for downstream classification.

\textbf{For image enhancement}, the comparison methods included RegGAN \cite{reggan}, CycleGAN \cite{yi2019cycleganmed}, MAE \cite{he2022masked}, DDIM \cite{song2020denoising}, SDM \cite{rombach2022ldm}, Flow Matching \cite{lipman2023flow}, and CAR \cite{li2024controlvar}. RegGAN incorporates registration-aware generative adversarial learning, whereas CycleGAN uses cycle-consistent adversarial learning for image translation. MAE is implemented using a TransUNet \cite{chen2021transunet} architecture incorporating a pretrained MAE ViT-B encoder, while DDIM and SDM represent diffusion-based generative modeling. Flow Matching learns continuous generative transformations through flow-based modeling, and CAR provides an autoregressive generative baseline. Regarding model initialization, RegGAN, CycleGAN, DDIM, and Flow Matching were trained from scratch, whereas MAE was initialized with pretrained ViT weights across all MRI and CT enhancement tasks. SDM was initialized from pretrained Stable Diffusion v1.4 weights, and CAR was initialized using pretrained visual tokenizer and autoregressive Transformer weights. All methods were evaluated on the same task-specific test data against the corresponding reference images, while retaining their respective model architectures and training strategies.

\subsubsection{Evaluation metrics and statistical methods}\label{sec:experimental_metrics}
Segmentation performance was evaluated using the Dice similarity coefficient (Dice) and the 95th-percentile Hausdorff distance (HD95). Dice measures the spatial overlap between the predicted region $P$ and reference region $G$:
\begin{equation}
\mathrm{Dice} = \frac{2|P\cap G|}{|P|+|G|}.
\end{equation}
HD95 was calculated on 2D axial slices as the 95th percentile of the bidirectional surface distances between the predicted and reference regions \cite{isensee2021nnunet}. Euclidean surface distances were computed using the actual in-plane voxel spacing, and HD95 values were reported in millimeters (mm). For datasets with multiple segmentation targets, Dice scores were averaged across the evaluated target regions.

Classification performance was assessed using accuracy (ACC), F1 score, and area under the receiver operating characteristic curve (AUC). Accuracy was defined as the proportion of correctly classified samples, and F1 score as the harmonic mean of precision and recall:
\begin{equation}
\begin{aligned}
\mathrm{ACC} &= \frac{N_{\mathrm{correct}}}{N},\\
\mathrm{F1} &= \frac{2\,\mathrm{Precision}\,\mathrm{Recall}}{\mathrm{Precision}+\mathrm{Recall}}.
\end{aligned}
\end{equation}
For multiclass classification, F1 scores were macro-averaged across classes. AUC was calculated by integrating the receiver operating characteristic curve, which plots the true-positive rate against the false-positive rate across classification thresholds. For multiclass tasks, one-versus-rest AUC was computed separately for each class and macro-averaged across classes. For the two volumetric classification benchmarks, consecutive axial slices were used to construct pseudo-3D samples for data augmentation, and predictions were evaluated at the case level.

Image enhancement performance was evaluated using peak signal-to-noise ratio (PSNR), structural similarity index (SSIM), root-mean-square error (RMSE), and learned perceptual image patch similarity (LPIPS). RMSE and PSNR were calculated for each evaluated image as
\begin{equation}
\begin{aligned}
\mathrm{RMSE} &=
\sqrt{\frac{1}{M}\sum_{i=1}^{M}
(\hat{x}_i-x_i)^2},\\
\mathrm{PSNR} &=
20\log_{10}\left(\frac{L}{\mathrm{RMSE}}\right),
\end{aligned}
\end{equation}
where $M$ denotes the number of evaluated pixels, $\hat{x}$ and $x$ represent the reconstructed and reference images, and $L$ denotes the reference-image intensity range. For MRI, the PSNR dynamic range was determined separately for each reference volume as $L_v=\max(x_v)-\min(x_v)$ and was kept consistent across its evaluated slices. For CT, the PSNR dynamic range was defined by the evaluated Hounsfield unit (HU) intensity range. PSNR was reported in decibels. SSIM was calculated using local luminance, contrast, and structural comparisons \cite{wang2004ssim}, with images normalized to $[-1,1]$ and a data range of 2.0. LPIPS measured perceptual differences using deep visual features \cite{zhang2018unreasonable}, with normalized images represented as three-channel inputs. For reporting, both SSIM and LPIPS values were multiplied by 100.

For segmentation and image enhancement, evaluation metrics were first calculated on individual 2D images or axial slices. When multiple evaluated slices belonged to the same case, slice-level scores were averaged within each case before aggregation across the test set. For native 2D datasets, individual test images were used as the evaluation units. Symmetric 95\% confidence intervals (CIs) for segmentation and image enhancement metrics were estimated using Student's $t$ distribution \cite{student1908probable}:
\begin{equation}
\mathrm{CI}_{95\%} =
\bar{m}\pm t_{0.975,n-1}\frac{s_m}{\sqrt{n}},
\end{equation}
where $\bar{m}$ and $s_m$ denote the sample mean and standard deviation of the metric, $n$ is the number of evaluation units, and $t_{0.975,n-1}$ is the corresponding Student's $t$ critical value.

For classification, symmetric 95\% CIs were estimated using nonparametric bootstrap resampling combined with a normal approximation. A total of 2,000 bootstrap iterations were performed on the held-out test set. In each iteration, test samples were randomly resampled with replacement, and ACC, macro-averaged F1, and AUC were recalculated from the resampled predictions. The bootstrap standard error was estimated as the standard deviation of the 2,000 bootstrap metric estimates. The symmetric 95\% CI was then constructed as
\begin{equation}
\mathrm{CI}_{95\%}^{\mathrm{boot}} =
\hat{m}\pm 1.96\,\widehat{\mathrm{SE}}_{\mathrm{boot}},
\end{equation}
where $\hat{m}$ is the metric calculated from the original test set and $\widehat{\mathrm{SE}}_{\mathrm{boot}}$ is its bootstrap-estimated standard error. For volumetric classification benchmarks, resampling was performed at the case level, whereas individual images were used for native 2D classification datasets.

\subsection{Multicenter Clinical Validation Protocols}
\label{sec:methods_clinical_validation}

The multicenter clinical evaluation included de-identified brain MRI data comprising 4,829 imaging examinations from 4,796 patients collected at six clinical centers. Jinling Hospital (JLH) and Nanjing First Hospital (NFH) served as the two internal centers for model development and internal evaluation. Nanjing Jiangning Hospital (NJH), Wuxi People's Hospital (WXPH), Northern Jiangsu People's Hospital (NJPH), and Changzhou Second People's Hospital (CZSPH) served as four external validation centers. Reference ischemic lesion annotations and TOAST subtype labels were provided by the participating clinical centers together with the imaging data. DWI, ADC, and FLAIR sequences were spatially registered to a common image space before task-specific processing.

\subsubsection{Multicenter validation across stroke cohorts}
\label{sec:methods_stroke_protocol}
We evaluated BrainGFM on three clinical stroke-imaging tasks: ischemic stroke lesion segmentation, TOAST stroke subtyping, and DWI-to-FLAIR MRI sequence transformation. The first two tasks used task-aligned data synthesis to support predictive learning under limited supervision, whereas the third used task-adapted post-training for paired image reconstruction. The task-specific cohorts comprised 4,318 cases for ischemic stroke lesion segmentation, 3,635 cases for TOAST stroke subtyping, and 3,842 cases for DWI-to-FLAIR MRI sequence transformation. For each task, 200 cases from the two internal centers, including 100 from JLH and 100 from NFH, constituted the training cohort, without a separate validation split. The remaining eligible internal cases and the cases from the four external centers were used for evaluation. Accordingly, the internal test cohorts comprised 3,608, 3,118, and 3,259 cases for the three tasks, while the corresponding external test cohorts comprised 510, 317, and 383 cases.

\textbf{Stroke lesion segmentation.} DWI images and their corresponding ischemic lesion masks were used to adapt BrainGFM through task-aligned data synthesis. A subset of 60 cases was randomly selected from the 200-case training cohort for low-data learning. All axial slices containing annotated lesion regions were retained for training. Following task-specific adaptation, BrainGFM generated synthetic lesion masks and corresponding DWI images through mask-conditioned generation, producing 3,000 synthetic image-mask pairs at the slice level. The resulting synthetic pairs were used to train a 2D UNet for lesion segmentation. For comparison, the conventional UNet and competing low-data methods used the same 60 annotated cases, whereas the full-data UNet reference was trained using all 200 training cases. During inference, the trained UNet predicted ischemic lesion masks from DWI images, and segmentation performance was assessed against the corresponding reference annotations.

\textbf{TOAST stroke subtyping.} BrainGFM was adapted for five-category TOAST stroke subtyping using DWI and ADC images together with the corresponding ischemic lesion masks predicted by the segmentation model. A subset of 60 cases was randomly selected from the task-specific 200-case training cohort for low-data learning. The segmentation model used to generate lesion masks was trained on cases disjoint from the classification test cohorts. BrainGFM was adapted using the corresponding stroke subtype labels and slice-level lesion-mask supervision to generate subtype-conditioned pseudo-3D training data. A total of 500 synthetic pseudo-3D cases were generated for each of the five stroke subtypes, yielding 2,500 synthetic cases in total. A ResNet50 classifier was subsequently trained using the generated data, with DWI, ADC, and lesion information incorporated into the classification inputs. Stroke subtype predictions were evaluated at the patient level. The competing low-data methods used the same 60 training cases, whereas the full-data ResNet50 reference was trained using all 200 cases.

\textbf{DWI-to-FLAIR MRI sequence transformation.} BrainGFM was adapted to reconstruct FLAIR images from DWI through task-adapted post-training. The task-specific 200-case training cohort provided paired DWI and FLAIR images, with DWI serving as the source condition and acquired FLAIR as the reconstruction target. The DWI and FLAIR sequences were spatially registered to a common image space to establish spatial correspondence between the source and target images. Paired axial slices with a brain-foreground area covering at least 5\% of the image were retained for training. During post-training, the pretrained visual tokenizer and autoregressive Transformer backbone were frozen, while LoRA and source-image conditioning parameters were optimized to predict FLAIR image tokens from the corresponding DWI inputs. The generated tokens were decoded into synthetic FLAIR images. Competing image-reconstruction methods were trained using the same paired training data and evaluated against the corresponding acquired FLAIR references.

\subsubsection{Blinded reader study}\label{sec:methods_blinded_reader}
The blinded reader study was conducted to assess the clinical usability and reliability of FLAIR images synthesized from DWI using BrainGFM (Fig.\ref{fig:clinical_reader_study}a). The study included 60 stroke cases, each evaluated under four imaging conditions: the acquired FLAIR image and synthetic FLAIR images generated by BrainGFM, MAE, and CAR from the corresponding DWI. Three senior radiologists, including one with more than 15 years of clinical experience and two with more than 10 years, reviewed the images. The images were randomly distributed across reading sessions, with the image sources concealed from the readers.

Each radiologist evaluated five dimensions of image quality using a five-point Likert scale \cite{likert1932technique}: overall diagnostic quality (ODQ), anatomical fidelity (AF), FLAIR contrast fidelity (CF), artifact- and hallucination-free quality (AHQ), and diagnostic confidence (DC). These dimensions assessed overall interpretability, preservation of anatomical structures, fidelity of FLAIR-specific contrast, absence of imaging artifacts or spurious anatomical features, and confidence in image-based interpretation. Higher scores indicated better perceived quality. For each radiologist and imaging condition, mean scores were calculated separately across the 60 cases for each dimension. To assess the synthesized images across multiple clinical quality criteria, we performed a threshold-based analysis across all five dimensions. For each reader-case assessment, an image was considered to satisfy a given quality threshold only when all five dimension scores were greater than or equal to that threshold. The proportion of qualifying assessments was calculated separately for each imaging condition using the 180 assessments from three radiologists and 60 cases.

\subsubsection{Human-AI collaboration study}\label{sec:methods_human_ai}
The human-AI collaboration study was conducted to evaluate the effect of BrainGFM assistance on radiologists' diagnosis of stroke subtypes (Fig.\ref{fig:clinical_reader_study}b). The study included 300 patients from the internal and external cohorts. This study focused on three major etiological subtypes, including large-artery atherosclerosis (LAA), cardioembolism (CE), and small-artery occlusion (SAO). For each patient, radiologists were provided with DWI and clinical information regarding large-vessel occlusion and atrial fibrillation, reflecting the combined use of imaging and clinical findings in stroke subtype assessment.

The same three senior radiologists who participated in the blinded reader study assessed all 300 patients in two reading sessions. During the first session, each radiologist reviewed the available imaging and clinical information and assigned a stroke subtype without AI assistance. After an interval of more than one day, the cases were reshuffled and presented for a second reading with access to the corresponding BrainGFM subtype predictions. The interval and reshuffling were intended to reduce recall and ordering effects between sessions. During the AI-assisted reading, radiologists determined their final diagnoses after considering the model predictions. Diagnostic confidence was recorded for each reading, and acceptance of the AI recommendation was additionally recorded during the assisted session.

Diagnostic performance was assessed by comparing unaided and AI-assisted radiologist decisions, as well as standalone BrainGFM predictions, against reference stroke subtype labels. Overall accuracy was defined as the proportion of correctly classified patients. One-versus-rest accuracy, sensitivity, specificity, positive predictive value (PPV), and F1 score were calculated for each subtype, with macro-averaged metrics obtained across LAA, CE, and SAO. Individual responses to AI assistance were further characterized by diagnostic confidence, AI recommendation acceptance, and changes in diagnostic correctness. High confidence was defined as a score of 3, and AI acceptance was determined from the recorded responses during assisted reading.

\section{Details of Data}
\label{sec:details_data}
\subsection{Segmentation Task Data}
We constructed 20 downstream segmentation tasks covering brain tumour segmentation (TuS, 13 tasks), brain tissue segmentation (TiS, three tasks), and intracranial haemorrhage and vessel segmentation targets (HVS, four tasks). The TuS tasks were derived from BraTS2021 \cite{bakas2021brats}, TCIA \cite{clark2013tcia}, TCGA \cite{weinstein2013tcga}, and BRISC2025 \cite{brisc2025}, using T1, T1ce, FLAIR, or multi-sequence MRI inputs. The TiS tasks used MRBrainS13 \cite{mrbrains13}, IBSR18 \cite{ibsr18}, and iSeg2017 \cite{iseg2017} to evaluate segmentation of brain tissue structures. The HVS tasks were constructed from BHSD \cite{wu2023bhsd}, INSTANCE2022 \cite{instance2022}, TopBrain \cite{topbrain_challenge_site}, and CT-ICH2020 \cite{physionet_ct_ich}.

Each task was defined by its imaging input and corresponding pixel-level reference annotations. For volumetric datasets, training and testing partitions were defined at the case level, ensuring that slices from the same case did not appear across the corresponding partitions. Low-data training used annotated slices from the volumes, while full-data reference models used the designated supervised training partitions. For multi-sequence tasks, corresponding anatomical locations across available MRI sequences were used to construct aligned inputs. Detailed input modalities, segmentation targets, and case- and slice-level partitions are reported in Supplementary Tab.~\ref{tab:supp_downstream_partitions}. 

\subsection{Classification Task Data}
Eight classification tasks were constructed across four categories: MRI sequence classification (SeC, one task), brain tumour classification (TuC, two tasks), intracranial haemorrhage classification (HaC, three tasks), and neuroimaging-based disorder classification (NeC, two tasks). The tasks were derived from BraTS2023 \cite{brats2023}, Br35H \cite{br35h}, BRISC2025 \cite{brisc2025}, INSTANCE2022 \cite{instance2022}, Head CT-hemorrhage, RSNA ICH \cite{flanders2020rsna}, ABIDE II \cite{dimartino2017abide2}, and Alzheimer-3D \cite{10.3389/fnagi.2020.583212}.

The classification tasks included both 2D and 3D learning settings. For MRI sequence classification, the original 3D MRI volumes were converted into 2D axial slices for slice-level classification. Similarly, INSTANCE2022 CT volumes were decomposed into all axial slices to construct a binary haemorrhage classification task, distinguishing slices with and without haemorrhage. Tasks based on native 2D datasets were performed  using 2D classifiers. In contrast, the two neuroimaging disorder classification tasks, ABIDE II and Alzheimer-3D, employed 3D classifiers. For these tasks, BrainGFM generated class-conditioned 2D slices, which were stacked into synthetic 3D volumes to provide volumetric training data. For datasets originating from 3D volumes, data partitioning was performed before slice extraction to prevent overlap between training and testing cases. Detailed classification targets and sample counts are summarized in Supplementary Tab.~\ref{tab:supp_downstream_partitions}.

\subsection{Image Enhancement Task Data}
Twenty image-enhancement tasks were constructed across three settings: simulated low-field brain MRI enhancement (LFMR, eight tasks), simulated low-dose brain CT enhancement (LDCT, four tasks), and non-contrast MRI enhancement (NCMR, eight tasks). The LFMR and LDCT tasks used simulated degraded images paired with their original reference images, whereas NCMR used paired non-contrast and contrast-enhanced MRI sequences. Data partitions were defined at the case level to prevent overlap between training and testing. The datasets and numbers of paired training and testing samples are summarized in Supplementary Tab.~\ref{tab:supp_downstream_partitions}.

\textbf{Simulated low-field brain MRI} Eight LFMR tasks were constructed from BraTS2021 \cite{bakas2021brats}, ABIDE II \cite{dimartino2017abide2}, IBSR18 \cite{ibsr18}, MSD Brain \cite{simpson2019msd}, and IXI \cite{ixi_dataset}, covering T1- and T2-weighted MRI. To simulate reduced image quality associated with low-SNR MRI acquisition, we generated degraded images from the corresponding reference MRI images using a k-space noise model \cite{lin2023low}. For each reference image $x$, the Fourier transform was calculated to obtain its k-space representation. Independent complex Gaussian noise was added to the k-space coefficients, followed by inverse Fourier transformation and magnitude reconstruction:
\begin{equation}
\tilde{x}=|\mathcal{F}^{-1}(\mathcal{F}(x)+\epsilon)|,\ \epsilon\sim\mathcal{CN}(0,\sigma_{\mathrm{MRI}}^2).
\end{equation}
Here, $\sigma_{\mathrm{MRI}}$ controls the simulated noise level and was parameterized as $\sigma_{\mathrm{MRI}}=\sqrt{\delta_t\,r}$ with the degradation strength $\delta_t=0.10$ and a calibrated reduction rate $r=0.005$. The noise standard deviation was thus set to $\sigma_{\mathrm{MRI}}\approx 0.022$, in units of the per-slice foreground min--max normalized intensity range: brain tissue was linearly rescaled to $[0,1]$ (background fixed at $0$), noise was injected in this normalized domain, and the degraded image was rescaled back to the original intensity range. Since the discrete Fourier transform is linear and orthogonal, this formulation is equivalent to perturbing the real and imaginary channels of the complex image with two Gaussian components, and the resulting magnitude image follows a Rician distribution, the physically established noise model of magnitude MRI data \cite{gudbjartsson1995rician}. The same degradation configuration was used across all datasets. The resulting degraded images were used as model inputs, with their original MRI images retained as reconstruction targets. The simulation was applied to the training and testing partitions.

\textbf{Simulated low-dose brain CT}
Four LDCT tasks were constructed using reference CT images from TopBrain \cite{topbrain_challenge_site}, CQ500 \cite{chilamkurthy2018cq500}, INSTANCE2022 \cite{instance2022}, and BHSD \cite{wu2023bhsd}. Synthetic low-dose CT images were generated using an image-domain Poisson noise model to approximate dose-dependent quantum noise \cite{ge2024imaging}. For each reference CT image, a relative dose factor $\beta$ was sampled for each slice from a uniform distribution over $[0.3,0.6]$. Poisson noise was applied to the non-negative brain-foreground intensity representation according to
\begin{equation}
\tilde{x}_{\mathrm{fg}}=\frac{1}{\beta}\cdot\mathrm{Poisson}(\beta x_{\mathrm{fg}}),\ \beta\sim\mathcal{U}(0.3,0.6),
\end{equation}
where $x_{\mathrm{fg}}$ denotes the non-negative foreground intensity representation, and $\beta$ controls the relative dose factor in the image-domain approximation. This formulation preserves the expected foreground intensity while increasing the noise variance as $\beta$ decreases, approximating the increased quantum noise associated with reduced CT radiation dose. The background outside the brain mask was kept unchanged. The degraded images served as source inputs, while the corresponding original CT images served as reconstruction targets. The same simulation protocol was applied to the designated training and testing partitions.

\textbf{Non-contrast MRI enhancement} Eight NCMR tasks were constructed from BraTS2021 \cite{bakas2021brats}, TCGA \cite{weinstein2013tcga}, TCIA \cite{clark2013tcia}, MSD Brain \cite{simpson2019msd}, BraTS-METS \cite{brats_mets}, BraTS-PEDs \cite{brats_peds}, UPenn-GBM \cite{upenn_gbm}, and Ivy GAP \cite{ivy_gap}. Each task used paired non-contrast T1-weighted and contrast-enhanced T1-weighted (T1ce) MRI from the same examination. The T1 images served as source inputs and the spatially corresponding T1ce images as reconstruction targets. Paired images were retained within the same case-level data partition.

\subsection{Multicenter Clinical Cohorts}
The multicenter clinical dataset comprised 4,796 unique patients and 4,829 brain MRI examinations from six hospitals. Jinling Hospital (JLH) and Nanjing First Hospital (NFH) constituted the two internal centers, while Nanjing Jiangning Hospital (NJH), Wuxi People's Hospital (WXPH), Northern Jiangsu People's Hospital (NJPH), and Changzhou Second People's Hospital (CZSPH) served as external validation centers. The internal centers contributed 2,617 and 1,669 patients, while the four external centers contributed 116, 143, 125, and 126 patients. Patient characteristics varied across institutions, including age, sex distribution, baseline stroke severity, TOAST subtype composition, and three-month functional outcomes. Detailed center-specific characteristics, including NIHSS, modified Rankin Scale
(mRS), and TOAST subtype distributions, are reported in Tab~\ref{tab:stroke_cohort}.

\begin{figure*}
    \centering
    \includegraphics[width=\textwidth]{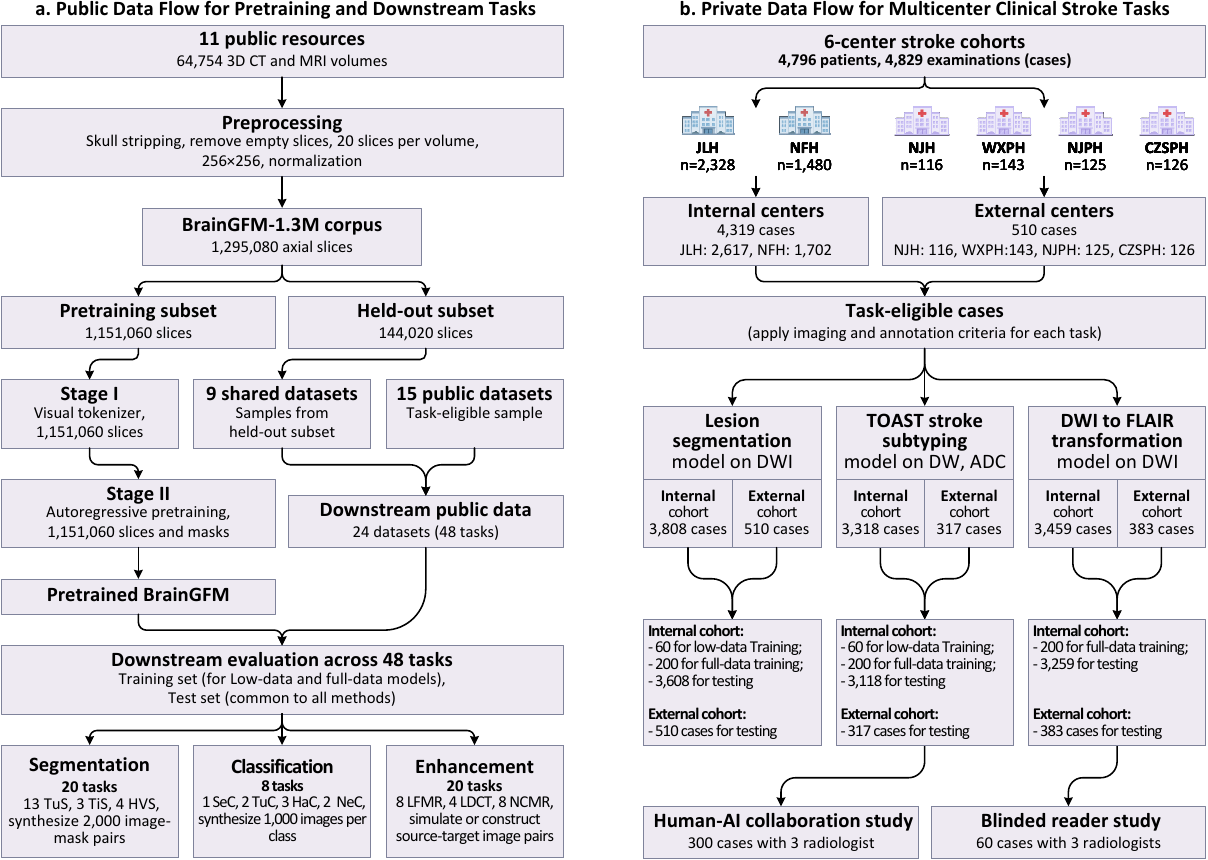}
    \caption{\textbf{Data flow and study design of BrainGFM.} \textbf{a.} Eleven public brain-imaging resources comprising 64,754 3D volumes and 1,295,080 axial slices were collected and partitioned into pretraining and held-out subsets. The pretraining subset was used for two-stage generative pretraining, including visual tokenizer training and autoregressive modeling.Held-out samples from nine shared datasets, together with 15 additional public datasets, supported downstream evaluation across 48 tasks, including segmentation (20 tasks), classification (8 tasks), and image enhancement (20 tasks). Task-specific training data were constructed through labeled image--mask pair generation, class-conditioned image synthesis, and source--target image pairing. \textbf{b.} Six clinical stroke cohorts comprising 4,796 patients and 4,829 cases from two internal centers (JLH and NFH) and four external centers (NJH, WXPH, NJPH, and CZSPH) were collected. Task-eligible cases were selected according to task-specific imaging and annotation criteria for lesion segmentation, TOAST subtyping, and DWI-to-FLAIR transformation. Internal-center data were used for task-specific training, including low-data settings, while separate internal and external test sets were used for clinical validation.}
    \label{fig:supp_data_flow}
\end{figure*}

The clinical dataset included DWI, ADC maps, and fluid-attenuated inversion recovery (FLAIR) images. Across all centers, 4,829 DWI volumes, 3,885 ADC volumes, and 3,842 FLAIR volumes were available. DWI was available for all included MRI examinations, whereas ADC and FLAIR availability varied by center and clinical examination. These differences in sequence availability, together with the availability of task-specific reference annotations and labels, determined the eligible cohorts for the three clinical evaluations. Ischemic lesion segmentation included 4,318 eligible cases, TOAST stroke subtyping included 3,635 patients, and DWI-to-FLAIR transformation included 3,842 paired examinations. For each task, 200 cases from the internal centers were reserved for model development, and the remaining eligible internal and external cases were used for evaluation. The corresponding data flow is illustrated in Fig.~\ref{fig:supp_data_flow}.

\clearpage

\begin{table*}
\caption{\textbf{Public or access-controlled brain imaging datasets used in this study.} The datasets used for BrainGFM pretraining and downstream evaluation are detailed below. The download link or access website is provided in the second column. * indicates the datasets in the BrainGFM-1.3M.}
\label{tab:supp_pretraining_sources}
\centering
\resizebox{\linewidth}{!}
{
\begin{tabular}{lll}
\hline
\rowcolor{tblhead}
\textbf{Dataset}                                    & \textbf{License or Use Terms} & \textbf{Link} \\
\hline
\rowcolor{tblalt}
*ADNI \cite{mueller2005adni}                         & Data Use Agreement            & \url{https://adni.loni.usc.edu/data-samples/adni-data/\#AccessData} \\
*FOMO-60K \cite{Cerri2026large}                      & CC BY-NC-SA 4.0               & \url{https://huggingface.co/datasets/FOMO-MRI/FOMO50K/tree/main} \\
\rowcolor{tblalt}
*ABIDE II \cite{dimartino2017abide2}                 & CC BY-NC-SA                   & \url{https://fcon_1000.projects.nitrc.org/indi/abide/abide_II.html} \\
*TCGA \cite{weinstein2013tcga}                       & Varies by data type           & \url{https://portal.gdc.cancer.gov/analysis_page?app=Downloads} \\
\rowcolor{tblalt}
*TCIA \cite{clark2013tcia}                           & Varies by data type           & \url{https://www.cancerimagingarchive.net/browse-collections/} \\
*IBSR18 \cite{ibsr18}                                & Non-commercial use only       & \url{https://www.nitrc.org/frs/?group_id=48} \\
\rowcolor{tblalt}
*MSD Brain \cite{simpson2019msd}                     & CC BY-SA 4.0                  & \url{https://msd-for-monai.s3-us-west-2.amazonaws.com/Task01_BrainTumour.tar} \\
*BraTS2021 \cite{bakas2021brats}                     & CC BY 4.0                     & \url{https://doi.org/10.7937/jc8x-9874} \\
\rowcolor{tblalt}
*CQ500 \cite{chilamkurthy2018cq500}                  & CC BY-NC-S                    & \url{https://headctstudy.qure.ai/dataset} \\
*RSNA ICH \cite{flanders2020rsna}                    & Non-commercial use only       & \url{https://imaging.rsna.org/dataset/1} \\
\rowcolor{tblalt}
*TopBrain \cite{topbrain_challenge_site}             & Custom data use terms         & \url{https://zenodo.org/records/16878417} \\
\hline
\hline
BHSD \cite{wu2023bhsd}                              & MIT License                   & \url{https://huggingface.co/datasets/Wendy-Fly/BHSD/tree/main} \\
\rowcolor{tblalt}
BraTS2023 \cite{brats2023}                          & Challenge-specific terms      & \url{https://www.synapse.org/Synapse:syn51514132} \\
BraTS-METS \cite{brats_mets}                        & Challenge-specific terms      & \url{https://www.synapse.org/Synapse:syn51514107} \\
\rowcolor{tblalt}
BraTS-PEDs \cite{brats_peds}                        & CC BY 4.0                     & \url{https://www.synapse.org/Synapse:syn51615054} \\
Br35H \cite{br35h}                                  & CC BY 4.0                     & \url{https://www.kaggle.com/datasets/ahmedhamada0/brain-tumor-detection/data} \\
\rowcolor{tblalt}
BRISC2025 \cite{brisc2025}                          & CC BY 4.0                     & \url{https://www.kaggle.com/datasets/briscdataset/brisc2025/data} \\
CT-ICH2020 \cite{physionet_ct_ich}                  & Data Use Agreement            & \url{https://physionet.org/content/ct-ich/1.3.1/} \\
\rowcolor{tblalt}
Head CT-hemorrhage                                  & CC0: Public Domain            & \url{https://www.kaggle.com/datasets/felipekitamura/head-ct-hemorrhage/data} \\
INSTANCE2022 \cite{instance2022}                    & Challenge-specific terms      & \url{https://instance.grand-challenge.org/Dataset/} \\
\rowcolor{tblalt}
iSeg2017 \cite{iseg2017}                            & Challenge-specific terms      & \url{https://iseg2017.web.unc.edu/download/} \\
Ivy GAP \cite{ivy_gap}                              & Controlled-access terms       & \url{https://www.cancerimagingarchive.net/collection/ivygap/} \\
\rowcolor{tblalt}
IXI \cite{ixi_dataset}                              & CC BY-SA 3.0                  & \url{https://biomedic.doc.ic.ac.uk/brain-development/downloads/IXI/} \\
MRBrainS13 \cite{mrbrains13}                        & Challenge-specific terms      & \url{https://doi.org/10.34894/645ZIN} \\
\rowcolor{tblalt}
UPenn-GBM \cite{upenn_gbm}                          & CC BY 4.0                     & \url{https://www.cancerimagingarchive.net/collection/upenn-gbm/} \\
Alzheimer-3D \cite{10.3389/fnagi.2020.583212}       & CC BY 4.0                     & \url{https://zenodo.org/records/3935636} \\

\hline
\end{tabular}
}
\end{table*}

\begin{table*}[!t]
\caption{\textbf{Composition of BrainGFM-1.3M.} The 11 source datasets comprise 64,754 3D volumes and 1,295,080 axial slices (20 slices per volume). Training and held-out partitions were constructed at the patient level using a ratio of 8:1, with all available volumes from the same patient assigned to the same partition. Owing to differences in the number of imaging sequences per patient, the resulting volume counts do not necessarily follow an exact 8:1 ratio. The table reports the corresponding volume and slice counts for each source.}
\label{tab:supp_pretraining_distribution}
\centering
\small
\resizebox{\linewidth}{!}
{%
\begin{tabular}{lccccccc}
\hline
\rowcolor{tblhead}
\textbf{Dataset} 
& \textbf{Modality}
& \textbf{Total volumes}
& \textbf{Total slices}
& \textbf{Train volumes}
& \textbf{Train slices}
& \textbf{Held-out volumes}
& \textbf{Held-out slices} \\
\hline
\rowcolor{tblalt}
ADNI       & MRI & 3,840  & 76,800  & 3,413  & 68,260  & 427  & 8,540 \\
FOMO-60K   & MRI & 47,591 & 951,820 & 42,303 & 846,060 & 5,288 & 105,760 \\
\rowcolor{tblalt}
ABIDE II   & MRI & 2,164  & 43,280  & 1,924  & 38,480  & 240  & 4,800 \\
TCGA       & MRI & 668    & 13,360  & 592    & 11,840  & 76   & 1,520 \\
\rowcolor{tblalt}
TCIA       & MRI & 2,684  & 53,680  & 2,384  & 47,680  & 300  & 6,000 \\
IBSR18     & MRI & 18     & 360     & 16     & 320     & 2    & 40 \\
\rowcolor{tblalt}
MSD Brain  & MRI & 1,936  & 38,720  & 1,720  & 34,400  & 216  & 4,320 \\
BraTS2021  & MRI & 4,844  & 96,880  & 4,304  & 86,080  & 540  & 10,800 \\
\rowcolor{tblalt}
CQ500      & CT  & 491    & 9,820   & 436    & 8,720   & 55   & 1,100 \\
RSNA ICH   & CT  & 478    & 9,560   & 425    & 8,500   & 53   & 1,060 \\
\rowcolor{tblalt}
TopBrain   & CT  & 40     & 800     & 36     & 720     & 4    & 80 \\
\hline
\rowcolor{tblsub}
\textbf{Total} & \textbf{MRI, CT}
& \textbf{64,754} & \textbf{1,295,080}
& \textbf{57,553} & \textbf{1,151,060}
& \textbf{7,201}  & \textbf{144,020} \\
\hline
\end{tabular}%
}
\end{table*}

\begin{table*}[!t]
\centering
\caption{\textbf{Training and testing data partitions for the 48 downstream BrainGFM tasks.} The table summarizes the data partitions for 20 segmentation, eight classification, and 20 image-enhancement tasks, with sample counts reported separately as cases/volumes and slices/images. For segmentation and classification, the low-data training partition denotes the few annotated samples used for task-specific adaptation, and the full-data training partition denotes the full setting reference for UNet (full) and ResNet50 (full). For image enhancement, the experiments are implemented in the full-data setting. For volumetric tumour-segmentation tasks, one case comprising 20 axial slices was reserved for low-data adaptation, and the remaining cases were divided approximately 5:1 into full-data training and testing partitions. Dashes indicate quantities that are not separately reported or are not applicable to the corresponding task. All sample counts represent real image data and exclude synthetically generated samples.}
\label{tab:supp_downstream_partitions}
\resizebox{\linewidth}{!}
{
\begin{tabular}{lccccccccl}
\hline
\rowcolor{tblhead}
& 
& 
& \multicolumn{2}{c}{\textbf{Low-data training}}
& \multicolumn{2}{c}{\color{gray}\textbf{Full-data training}}
& \multicolumn{2}{c}{\textbf{Testing}} 
& \\
\hhline{~~~------}
\rowcolor{tblhead}
\multirow{-2}{*}{\textbf{Task-Dataset-Modality}}
& \multirow{-2}{*}{\textbf{Native}}
& \multirow{-2}{*}{\textbf{Modality}}
& \textbf{Volumes}
& \textbf{Slices}
& \textbf{\color{gray}Volumes}
& \textbf{\color{gray}Slices}
& \textbf{Volumes}
& \textbf{Slices} 
& \multirow{-2}{*}{\textbf{Target}} \\
\hline
\rowcolor{tblsub}
\multicolumn{10}{l}{\textbf{TuS: Brain tumour segmentation (13 tasks)}} \\
\hline
\rowcolor{tblalt}
TuS-BraTS21-T1 & 3D & T1 & - & 20 & {\color{gray}-} & {\color{gray}2,240} & - & 440 & Tumour regions (ET, TC, WT)\\
TuS-TCIA-T1 & 3D & T1 & - & 20 & {\color{gray}-} & {\color{gray}1,240} & - & 240 & Brain tumour masks\\
\rowcolor{tblalt}
TuS-TCGA-T1 & 3D & T1 & - & 20 & {\color{gray}-} & {\color{gray}300} & - & 60 & Brain tumour masks\\
TuS-BraTS21-T1ce & 3D & T1ce & - & 20 & {\color{gray}-} & {\color{gray}2,240} & - & 440 & Tumour regions (ET, TC, WT)\\
\rowcolor{tblalt}
TuS-TCIA-T1ce & 3D & T1ce & - & 20 & {\color{gray}-} & {\color{gray}1,240} & - & 240 & Brain tumour masks\\
TuS-TCGA-T1ce & 3D & T1ce & - & 20 & {\color{gray}-} & {\color{gray}300} & - & 60 & Brain tumour masks\\
\rowcolor{tblalt}
TuS-BraTS21-FLAIR & 3D & FLAIR & - & 20 & {\color{gray}-} & {\color{gray}2,240} & - & 440 & Tumour regions (ET, TC, WT)\\
TuS-TCIA-FLAIR & 3D & FLAIR & - & 20 & {\color{gray}-} & {\color{gray}1,240} & - & 240 & Brain tumour masks\\
\rowcolor{tblalt}
TuS-TCGA-FLAIR & 3D & FLAIR & - & 20 & {\color{gray}-} & {\color{gray}300} & - & 60 & Brain tumour masks\\
TuS-BraTS21-Multi & 3D & T1/T1ce/T2/FLAIR & - & 20/mod. & {\color{gray}-} & {\color{gray}2,240/mod.} & - & 440/mod. & Tumour regions (ET, TC, WT)\\
\rowcolor{tblalt}
TuS-TCIA-Multi & 3D & T1/T1ce/T2/FLAIR & - & 20/mod. & {\color{gray}-} & {\color{gray}1,240/mod.} & - & 240/mod. & Brain tumour masks\\
TuS-TCGA-Multi & 3D & T1/T1ce/T2/FLAIR & - & 20/mod. & {\color{gray}-} & {\color{gray}300/mod.} & - & 60/mod. & Brain tumour masks\\
\rowcolor{tblalt}
TuS-BRISC25-T1 & 2D & T1 & - & 20 & {\color{gray}-} & {\color{gray}3,977} & - & 796 & Binary tumour masks\\
\hline
\rowcolor{tblsub}
\multicolumn{10}{l}{\textbf{TiS: Brain tissue segmentation (3 tasks)}} \\
\hline
TiS-MRBrainS13-Multi & 3D & T1/FLAIR & - & 20/mod. & {\color{gray}-} & {\color{gray}40/mod.} & - & 40/mod. & Brain tissue (GM, WM, CSF)\\
\rowcolor{tblalt}
TiS-IBSR18-T1 & 3D & T1 & - & 20 & {\color{gray}-} & {\color{gray}20} & - & 20 & Brain tissue and anatomical annotations\\
TiS-iSeg17-T1 & 3D & T1 & - & 20 & {\color{gray}-} & {\color{gray}140} & - & 40 & Infant brain tissue (GM, WM, CSF)\\
\hline
\rowcolor{tblsub}
\multicolumn{10}{l}{\textbf{HVS: haemorrhage and vessel segmentation (4 tasks)}} \\
\hline
\rowcolor{tblalt}
HVS-BHSD-CT & 3D & CT & - & 20 & {\color{gray}-} & {\color{gray}3180} & - & 640 & Intracranial haemorrhage masks\\
HVS-INSTANCE22-CT & 3D & CT & - & 20 & {\color{gray}-} & {\color{gray}1640} & - & 340 & Intracranial haemorrhage masks\\
\rowcolor{tblalt}
HVS-TopBrain-CT & 3D & CT & - & 20 & {\color{gray}-} & {\color{gray}60} & - & 60 & Brain vessel masks\\
HVS-CT-ICH20-CT & 3D & CT & - & 20 & {\color{gray}-} & {\color{gray}580} & - & 120 & Intracranial haemorrhage masks\\
\hline
\hline
\rowcolor{tblsub}
\multicolumn{10}{l}{\textbf{SeC: MRI sequence classification (1 task)}} \\
\hline
\rowcolor{tblalt}
SeC-BraTS23-\# & 3D & MRI sequences & - & 20/class & {\color{gray}-} & {\color{gray}2,220/class} & - & 440/class & MRI-sequence identification\\
\hline
\rowcolor{tblsub}
\multicolumn{10}{l}{\textbf{TuC: Brain tumour classification (2 tasks)}} \\
\hline
TuC-Br35H-Unknow & 2D & MRI & - & 20/class & {\color{gray}-} & {\color{gray}1,233/class} & - & 247/class & Tumour, normal\\
\rowcolor{tblalt}
TuC-BRISC25-T1 & 2D & T1 & - & 20/class & {\color{gray}-} & {\color{gray}4,933} & - & 987 & Glioma, meningioma, pituitary tumour, normal\\
\rowcolor{tblsub}
\multicolumn{10}{l}{\textbf{HaC: Haemorrhage classification (3 tasks)}} \\
\hline
HaC-INSTANCE22-CT & 3D & CT & - & 20/class & {\color{gray}-} & {\color{gray}727/class} & - & 146/class & Haemorrhage-based labels\\
\rowcolor{tblalt}
HaC-Head CT-hemorrhage-CT & 2D & CT & - & 20/class & {\color{gray}-} & {\color{gray}67/class} & - & 13/class & Haemorrhage, normal\\
HaC-RSNA ICH-CT & 2D & CT & - & 20/class & {\color{gray}-} & {\color{gray}817/class} & - & 163/class & EDH, IPH, IVH, SAH, SDH, ICH\\
\hline
\rowcolor{tblsub}
\multicolumn{10}{l}{\textbf{NeC: Neuroimaging disorder classification (2 tasks)}} \\
\hline
\rowcolor{tblalt}
NeC-ABIDE-II-3D-T1 & 3D & T1 & 1/class & - & {\color{gray}198} & {\color{gray}-} & 40 & - & Autism spectrum disorder, typically developing control\\
NeC-Alzheimer-3D-T1 & 3D & T1 & 1/class & - & {\color{gray}50} & {\color{gray}-} & 25 & - & AD, bipolar disorder, healthy control\\
\hline
\hline
\rowcolor{tblsub}
\multicolumn{10}{l}{\textbf{LFMR: Low-field MRI enhancement (8 tasks)}} \\
\hline
\rowcolor{tblalt}
LFMR-BraTS21-T1 & 3D & T1 & {\color{gray}-} & {\color{gray}-} & - & 13880 & - & 2708 & Normal field T1\\
LFMR-BraTS21-T2 & 3D & T2 & {\color{gray}-} & {\color{gray}-} & - & 13880 & - & 2708 & Normal field T2\\
\rowcolor{tblalt}
LFMR-ABIDE-II-T1 & 3D & T1 & {\color{gray}-} & {\color{gray}-} & - & 24218 & - & 4846 & Normal field T1 \\
LFMR-IBSR-T1 & 3D & T1 & {\color{gray}-} & {\color{gray}-} & - & 133 & - & 126 & Normal field T1 \\
\rowcolor{tblalt}
LFMR-MSD-T1 & 3D & T1 & {\color{gray}-} & {\color{gray}-} & - & 5455 & - & 1079 & Normal field T1 \\
LFMR-MSD-T2 & 3D & T2 & {\color{gray}-} & {\color{gray}-} & - & 5455 & - & 1079 & Normal field T2 \\
\rowcolor{tblalt}
LFMR-IXI-T1 & 3D & T1 & {\color{gray}-} & {\color{gray}-} & - & 41,162 & - & 8263 & Normal field T1 \\
LFMR-IXI-T2 & 3D & T2 & {\color{gray}-} & {\color{gray}-} & - & 41,162 & - & 8263 & Normal field T2 \\
\hline
\rowcolor{tblsub}
\multicolumn{10}{l}{\textbf{LDCT: Low-dose CT enhancement (4 tasks)}} \\
\hline
\rowcolor{tblalt}
LDCT-TopBrain-CT & 3D & CT & {\color{gray}-} & {\color{gray}-} & - & 514 & - & 635 & Normal dose CT \\
LDCT-CQ500-CT & 3D & CT & {\color{gray}-} & {\color{gray}-} & - & 4,833 & - & 866 & Normal dose CT \\
\rowcolor{tblalt}
LDCT-INSTANCE22-CT & 3D & CT & {\color{gray}-} & {\color{gray}-} & - & 1,916 & - & 413 &  Normal dose CT \\
LDCT-BHSD-CT & 3D & CT & {\color{gray}-} & {\color{gray}-} & - & 3,693 & - & 758 & Normal dose CT \\
\hline
\rowcolor{tblsub}
\multicolumn{10}{l}{\textbf{NCMR: Non-contrast MRI enhancement (8 tasks)}} \\
\hline
\rowcolor{tblalt}
NCMR-BraTS21-T1 & 3D & T1 & {\color{gray}-} & {\color{gray}-} & - & 13880 & - & 2780 & T1ce \\
NCMR-TCGA-T1 & 3D & T1 & {\color{gray}-} & {\color{gray}-} & - & 1932 & - & 334 & T1ce \\
\rowcolor{tblalt}
NCMR-TCIA-T1 & 3D & T1 & {\color{gray}-} & {\color{gray}-} & - & 7682 & - & 1418 & T1ce \\
NCMR-MSDBrain-T1 & 3D & T1 & {\color{gray}-} & {\color{gray}-} & - & 5,455 & - & 1079 & T1ce \\
\rowcolor{tblalt}
NCMR-BraTS-METS-T1 & 3D & T1 & {\color{gray}-} & {\color{gray}-} & - & 132,214 & - & 26,431 & T1ce \\
NCMR-BraTS-PEDs-T1 & 3D & T1 & {\color{gray}-} & {\color{gray}-} & - & 26,186 & - & 5,163 & T1ce \\
\rowcolor{tblalt}
NCMR-UPenn-GBM-T1 & 3D & T1 & {\color{gray}-} & {\color{gray}-} & - & 62,124 & - & 11,647 & T1ce \\
NCMR-Ivy GAP-T1 & 3D & T1 & {\color{gray}-} & {\color{gray}-} & - & 3,428 & - & 734 & T1ce \\
\hline
\rowcolor{tblsub}
\multicolumn{10}{l}{\textbf{Total: 48 downstream tasks (20 segmentation, 8 classification, 20 image enhancement)}} \\
\hline
\end{tabular}
}
\end{table*}

\begin{table*}[!t]
\centering
\caption{\textbf{Characteristics of the multicenter clinical stroke cohorts.} Data are reported as mean $\pm$ s.d., median (IQR), or $n$ (\%), as appropriate. Percentages are based on available cases. JLH, Jinling Hospital; NFH, Nanjing First Hospital; NJH, Nanjing Jiangning Hospital; WXPH, Wuxi People's Hospital; NJPH, Northern Jiangsu People's Hospital; CZSPH, Changzhou Second People's Hospital; NIHSS, National Institutes of Health Stroke Scale; TOAST, Trial of Org 10172 in Acute Stroke Treatment; mRS, modified Rankin Scale.}
\label{tab:stroke_cohort}
\resizebox{\linewidth}{!}
{
\begin{tabular}{llllllll}
\hline
\rowcolor{tblhead}
& \multicolumn{2}{l}{\textbf{Internal cohorts}}
& \multicolumn{4}{l}{\textbf{External cohorts}}
&
\\ \hhline{~------~}
\rowcolor{tblhead}
& \textbf{JLH}
& \textbf{NFH}
& \textbf{NJH}
& \textbf{WXPH}
& \textbf{NJPH}
& \textbf{CZSPH}
& \multirow{-2}{*}{\textbf{Total}}
\\
\hline
\rowcolor{tblsub}
\multicolumn{8}{l}{\textbf{Cohort characteristics}}
\\
\rowcolor{tblalt}
Patients, $n$
& 2,617 & 1,669 & 116 & 143 & 125 & 126 & 4,796 \\
Imaging examinations, $n$
& 2,617 & 1,702 & 116 & 143 & 125 & 126 & 4,829 \\
\rowcolor{tblalt}
Age, years
& 67.7$\pm12.3$, $n$=2,251 & 61.4$\pm11.3$, $n$=1,582 & 67.1$\pm13.2$, $n$=116 & 67.2$\pm13.3$, $n$=143 & 69.2$\pm11.8$, $n$=125 & 67.2$\pm12.3$, $n$=126 & 65.4$\pm12.4$, $n$=4,343 \\
Female sex, $n/N$ (\%)
& 710/2,251 (31.5\%) & 465/1,582 (29.4\%) & 31/116 (26.7\%) & 72/143 (50.3\%) & 55/125 (44.0\%) & 52/126 (41.3\%) & 1,385/4,343 (31.9\%) \\
\rowcolor{tblalt}
NIHSS, median (IQR)
& 4 (2-8) & 6 (3-9) & 6 (2-10) & 4 (2-8) & 4 (2-8) & 4 (2-5) & 5 (2-8) \\
\hline
\rowcolor{tblsub}
\multicolumn{8}{l}{\textbf{Imaging characteristics}} \\
\rowcolor{tblalt}
DWI volumes, $n$
& 2,617 & 1,702 & 116 & 143 & 125 & 126 & 4,829 \\
FLAIR volumes, $n$
& 1,806 & 1,653 & 89 & 129 & 91 & 74 & 3,842 \\
\rowcolor{tblalt}
ADC volumes, $n$
& 1,814 & 1,610 & 115 & 136 & 118 & 92 & 3,885 \\
All available MRI volumes, $n$
& 6,237 & 4,965 & 320 & 408 & 334 & 292 & 12,556 \\
\hline
\rowcolor{tblsub}
\multicolumn{8}{l}{\textbf{TOAST stroke subtype composition}} \\
\rowcolor{tblalt}
TOAST-labeled patients, $n$
& 2,251 & 1,582 & 116 & 78 & 55 & 117 & 4,199 \\
Large-artery atherosclerosis, $n$ (\%)
& 1,116 (49.6\%) & 1,219 (77.1\%) & 65 (56.0\%) & 41 (52.6\%) & 37 (67.3\%) & 62 (53.0\%) & 2,540 (60.5\%) \\
\rowcolor{tblalt}
Cardioembolism, $n$ (\%)
& 315 (14.0\%) & 132 (8.3\%) & 14 (12.1\%) & 19 (24.4\%) & 2 (3.6\%) & 7 (6.0\%) & 489 (11.6\%) \\
Small-vessel occlusion, $n$ (\%)
& 736 (32.7\%) & 120 (7.6\%) & 33 (28.4\%) & 15 (19.2\%) & 14 (25.5\%) & 48 (41.0\%) & 966 (23.0\%) \\
\rowcolor{tblalt}
Other determined aetiology, $n$ (\%)
& 26 (1.2\%) & 59 (3.7\%) & 1 (0.9\%) & 3 (3.8\%) & 1 (1.8\%) & 0 (0.0\%) & 90 (2.1\%) \\
Undetermined aetiology, $n$ (\%)
& 58 (2.6\%) & 52 (3.3\%) & 3 (2.6\%) & 0 (0.0\%) & 1 (1.8\%) & 0 (0.0\%) & 114 (2.7\%) \\
\hline
\rowcolor{tblsub}
\multicolumn{8}{l}{\textbf{Clinical outcome}} \\
\rowcolor{tblalt}
Patients with 3-month mRS, $n$
& 2,251 & 1,582 & 116 & 143 & 125 & 126 & 4,343 \\
3-month mRS, median (IQR)
& 1 (1-2) & 1 (1-3) & 1 (0-2) & 1 (0-3) & 1 (0-2) & 0 (0-1) & 1 (1-2) \\
\rowcolor{tblalt}
mRS 0-2, $n$ (\%)
& 1,849 (82.1\%) & 1,082 (68.4\%) & 93 (80.2\%) & 106 (74.1\%) & 98 (78.4\%) & 113 (89.7\%) & 3,341 (76.9\%) \\
mRS 3-6, $n$ (\%)
& 402 (17.9\%) & 500 (31.6\%) & 23 (19.8\%) & 37 (25.9\%) & 27 (21.6\%) & 13 (10.3\%) & 1,002 (23.1\%) \\
\hline
\rowcolor{tblsub}
\multicolumn{8}{l}{\textbf{Task-specific cohorts}} \\
\rowcolor{tblalt}
MRI sequence transformation cases, $n$
& 1,806 & 1,653 & 89 & 129 & 91 & 74 & 3,842 \\
Lesion segmentation cases, $n$
& 2,328 & 1,480 & 116 & 143 & 125 & 126 & 4,318 \\
\rowcolor{tblalt}
TOAST stroke subtyping patients, $n$
& 1,814 & 1,504 & 115 & 71 & 48 & 83 & 3,635 \\
\hline
\end{tabular}
}
\end{table*}

\begin{table*}
    \centering
    \caption{\textbf{Radiologist assessment of real and generated FLAIR images.} Expert-specific and pooled ratings across five image-quality domains. Values are presented as mean ratings with the reported CIs in parentheses. All images were assessed on a five-point Likert scale, with higher scores indicating better perceived quality. ODQ, overall diagnostic quality; AF, anatomical fidelity; CF, contrast fidelity; AHQ, artifact- and hallucination-free quality; DC, diagnostic confidence. The ``Mean'' row represents the arithmetic mean across the five quality domains. ``All readers'' denotes the pooled ratings from the three experts. \textbf{Bold} and \underline{underlined} values indicate the highest and second-highest mean scores, within each reader and evaluation metric.}
    \label{tab:supp_reader_study}
    \resizebox{0.55\textwidth}{!}
    {
    \begin{tabular}{lccccc}
        \hline\rowcolor{tblhead}
        \textbf{Experts} & \textbf{Metric} & Real  & \textbf{BrainGFM}  & MAE & CAR\\
        \hline
        \rowcolor{tblalt} \cellcolor{tblsub}& ODQ & \textbf{3.65} & \underline{3.35} & 2.70 & 2.48 \\
        \rowcolor{tblalt} \cellcolor{tblsub}& & \textbf{(3.44-3.86)} & \underline{(3.14-3.56)} & (2.49-2.91) & (2.27-2.69) \\
        \cellcolor{tblsub}&AF  & \textbf{3.50} & \underline{3.22} & 2.70 & 2.43 \\
        \cellcolor{tblsub}& & \textbf{(3.29-3.71)} & \underline{(3.01-3.43)} & (2.49-2.91) & (2.22-2.64) \\
        \rowcolor{tblalt} \cellcolor{tblsub}& CF  & \textbf{3.50} & \underline{3.23} & 2.70 & 2.47 \\
        \rowcolor{tblalt} \cellcolor{tblsub}& & \textbf{(3.29-3.71)} & \underline{(3.02-3.44)}& (2.49-2.91) & (2.26-2.68) \\
        \cellcolor{tblsub}& AHQ & \textbf{3.32} & \underline{3.18} & 2.45 & 1.95 \\
        \cellcolor{tblsub}& & \textbf{(3.09-3.55)} & \underline{(2.95-3.41)} & (2.22-2.68) & (1.72-2.18) \\
        \rowcolor{tblalt} \cellcolor{tblsub}& DC & \textbf{3.53} & \underline{3.25} & 2.70 & 2.47 \\
        \rowcolor{tblalt} \cellcolor{tblsub}& & \textbf{(3.32-3.74)} & \underline{(3.04-3.46)} & (2.49-2.91) & (2.26-2.68) \\
         \cdashline{2-6}[0.8pt/2pt]
        \cellcolor{tblsub}& Mean & \textbf{3.50} & \underline{3.25} & 2.65 & 2.36 \\
        \multirow{-12}{*}{\makecell{\textbf{Expert 1}\\\textbf{$\geq$15 years}}}
        \cellcolor{tblsub}& & \textbf{(3.29-3.71)} & \underline{(3.04-3.46)} & (2.44-2.86) & (2.15-2.57) \\
        \hline
        \rowcolor{tblalt} \cellcolor{tblsub}& ODQ & \textbf{3.92} & \underline{3.65} & 2.73 & 2.72 \\
        \rowcolor{tblalt} \cellcolor{tblsub}& & \textbf{(3.74-4.10)} & \underline{(3.47-3.83)} & (2.55-2.91) & (2.54-2.90) \\
        \cellcolor{tblsub}& AF & \textbf{3.88} & \underline{3.62} & 2.82 & 2.73 \\
        \cellcolor{tblsub}& & \textbf{(3.69-4.07)} & \underline{(3.43-3.81)} & (2.63-3.01) & (2.54-2.92) \\
        \rowcolor{tblalt} \cellcolor{tblsub}& CF & \textbf{4.08} & \underline{3.83} & 1.85 & 1.62 \\
        \rowcolor{tblalt} \cellcolor{tblsub}& & \textbf{(3.90-4.26)} & \underline{(3.65-4.01)} & (1.67-2.03) & (1.44-1.80) \\
        \cellcolor{tblsub}& AHQ & \textbf{3.72} & \underline{3.53} & 1.87 & 1.58 \\
        \cellcolor{tblsub}& & \textbf{(3.52-3.92)} & \underline{(3.33-3.73)} & (1.67-2.07) & (1.38-1.78) \\
        \rowcolor{tblalt} \cellcolor{tblsub}& DC & \textbf{3.92} & \underline{3.63} & 2.62 & 2.63 \\
        \rowcolor{tblalt} \cellcolor{tblsub}& & \textbf{(3.74-4.10)} & \underline{(3.45-3.81)} & (2.44-2.80) & (2.45-2.81) \\
         \cdashline{2-6}[0.8pt/2pt]
        \cellcolor{tblsub}& Mean & \textbf{3.90} & \underline{3.65} & 2.38 & 2.26 \\
        \multirow{-12}{*}{\makecell{\textbf{Expert 2}\\\textbf{$\geq$10 years}}}
        \cellcolor{tblsub}& & \textbf{(3.73-4.07)} & \underline{(3.48-3.82)} & (2.21-2.55) & (2.09-2.43) \\
        \hline
        \rowcolor{tblalt} \cellcolor{tblsub}& ODQ & 3.15 & \textbf{3.52} & \underline{3.17} & 3.08 \\
        \rowcolor{tblalt} \cellcolor{tblsub}& & \underline{(2.91-3.39)} & \textbf{(3.28-3.76)} & (2.93-3.41) & (2.84-3.32) \\
        \cellcolor{tblsub}& AF & \underline{3.25} & \textbf{3.60} & 3.12 & 3.05 \\
        \cellcolor{tblsub}& & \underline{(3.03-3.47)} & \textbf{(3.38-3.82)} & (2.90-3.34) & (2.83-3.27) \\
        \rowcolor{tblalt} \cellcolor{tblsub}& CF & \underline{3.18} & \textbf{3.48} & 3.00 & 2.87 \\
        \rowcolor{tblalt} \cellcolor{tblsub}& & \underline{(2.95-3.41)} & \textbf{(3.25-3.71)} & (2.77-3.23) & (2.64-3.10) \\
        \cellcolor{tblsub}& AHQ & \underline{3.12} & \textbf{3.50} & 2.85 & 2.62 \\
        \cellcolor{tblsub}& & \underline{(2.90-3.34)} & \textbf{(3.28-3.72)} & (2.63-3.07) & (2.40-2.84) \\
        \rowcolor{tblalt} \cellcolor{tblsub}& DC & \underline{3.45} & \textbf{3.65} & 3.52 & 3.48 \\
        \rowcolor{tblalt} \cellcolor{tblsub}& & \underline{(3.22-3.68)} & \textbf{(3.42-3.88)} & (3.29-3.75) &(3.25-3.71) \\
         \cdashline{2-6}[0.8pt/2pt]
        \cellcolor{tblsub}& Mean  & \underline{3.23} & \textbf{3.55} & 3.13 & 3.02 \\
        \multirow{-12}{*}{\makecell{\textbf{Expert 3}\\\textbf{$\geq$10 years}}}
        \cellcolor{tblsub}& & \underline{(3.01-3.45)} & \textbf{(3.33-3.77)} & (2.91-3.35) & (2.80-3.24) \\
        \hline
        \hline
        \rowcolor{tblalt} \cellcolor{tblsub}& ODQ & \textbf{3.57} & \underline{3.51} & 2.87 & 2.76 \\
        \rowcolor{tblalt} \cellcolor{tblsub}& & \textbf{(3.38-3.76)} & \underline{(3.32-3.70)} & (2.68-3.06) & (2.57-2.95) \\
        \cellcolor{tblsub}& AF & \textbf{3.54} & \underline{3.48} & 2.88 & 2.74 \\
        \cellcolor{tblsub}& & \textbf{(3.35-3.73)} & \underline{(3.29-3.67)} & (2.69-3.07) & (2.55-2.93) \\
        \rowcolor{tblalt} \cellcolor{tblsub}& CF & \textbf{3.59} & \underline{3.51} & 2.52 & 2.32 \\
        \rowcolor{tblalt} \cellcolor{tblsub}& & \textbf{(3.41-3.77)} & \underline{(3.33-3.69)} & (2.34-2.70) & (2.14-2.50) \\
        \cellcolor{tblsub}& AHQ & \underline{3.38} & \textbf{3.40} & 2.39 & 2.05 \\
        \cellcolor{tblsub}& & \underline{(3.19-3.57)} & \textbf{(3.21-3.59)} & (2.20-2.58) & (1.86-2.24) \\
        \rowcolor{tblalt} \cellcolor{tblsub}& DC & \textbf{3.63} & \underline{3.51} & 2.94 & 2.86 \\
        \rowcolor{tblalt} \cellcolor{tblsub}& & \textbf{(3.44-3.82)} &  \underline{(3.32-3.70)} & (2.75-3.13) & (2.67-3.05) \\
         \cdashline{2-6}[0.8pt/2pt]
        \cellcolor{tblsub}& Mean  & \textbf{3.54} & \underline{3.48} & 2.72 & 2.55 \\
        \multirow{-12}{*}{\makecell{\textit{\textbf{Average}}\\\textit{\textbf{experts}}}}
        \cellcolor{tblsub}&       & \textbf{(3.36-3.72)} & \underline{(3.30-3.66)} & (2.54-2.90) & (2.37-2.73) \\
        \hline
    \end{tabular}}
\end{table*}

\begin{table*}[!t]
    \centering
    \caption{\textbf{Human-AI collaboration study: Reader-specific diagnostic performance and response to BrainGFM assistance in internal and external stroke cohorts.} Diagnostic performance was compared before and after AI assistance, with standalone BrainGFM as a reference. Values in parentheses indicate absolute changes (\%) in percentage points following AI assistance. Diagnostic confidence, AI acceptance, and diagnostic corrections were summarized across both cohorts. Expert 2 exhibited the largest performance improvements, alongside the lowest baseline high-confidence rate (8.3\%), highest AI acceptance rate (90.3\%), and greatest proportion of incorrect diagnoses corrected (27.3\%). In contrast, Experts 1 and 3 showed higher baseline confidence, lower AI acceptance, and more limited or variable improvements. These findings highlight substantial inter-reader heterogeneity in the clinical benefits of BrainGFM assistance.}
    \label{tab:human_ai_reader_subtype_performance}
    \footnotesize
    \resizebox{\linewidth}{!}
    {
    \begin{tabular}{cllccccccc}
    \hline
    \rowcolor{tblhead}
        &  
        & 
        & \multicolumn{2}{c}{\textbf{Expert 1} ($\geq$15 yr)}
        & \multicolumn{2}{c}{\textbf{Expert 2} ($\geq$10 yr)} 
        & \multicolumn{2}{c}{\textbf{Expert 3} ($\geq$10 yr)} 
        & \\
    \hhline{~~~------~}
    \rowcolor{tblhead}
        \multirow{-2}{*}{\textbf{Cohort}}
        & \multirow{-2}{*}{\textbf{Metric}}
        & \multirow{-2}{*}{\textbf{Subtype}}
        & Human & Human + AI & Human & Human + AI & Human & Human + AI 
        & \multirow{-2}{*}{\textbf{BrainGFM}} \\
    \hline
        \rowcolor{tblalt} \cellcolor{tblhead}
        & \cellcolor{tblsub}
        & LAA    
        & 67.3 & 63.3 (-4.0) & 54.7 & 80.7 (+26.0) & 65.1 & 62.3 (-2.7) & 80.7 \\
        \cellcolor{tblhead}
        & \cellcolor{tblsub}          
        & CE          
        & 84.7 & 86.7 (+2.0) & 87.3 & 90.7 (+3.3) & 89.7 & 91.1 (+1.4) & 91.3 \\
        \rowcolor{tblalt} \cellcolor{tblhead}
        & \cellcolor{tblsub}          
        & SAO 
        & 73.3 & 67.3 (-6.0) & 60.7 & 82.0 (+21.3) & 67.1 & 67.1 (0.0) & 82.7 \\
        \cdashline{3-10}[0.8pt/2pt]
        \cellcolor{tblhead}
        & \cellcolor{tblsub} \multirow{-4}{*}{Accuracy}
        & \textbf{Macro mean} 
        & 75.1 & 72.4 (-2.7) & 67.6 & 84.4 (+16.9) & 74.0 & 73.5 (-0.5) & 84.9 \\
        \hhline{~---------}
        \rowcolor{tblalt}  \cellcolor{tblhead}
        & \cellcolor{tblsub} 
        & LAA   
        & 69.7 & 69.7 (0.0) & 90.9 & 90.9 (0.0) & 71.4 & 77.8 (+6.3) & 84.9 \\
        \cellcolor{tblhead}
        & \cellcolor{tblsub}         
        & CE          
        & 31.6 & 15.8 (-15.8) & 36.8 & 63.2 (+26.3) & 36.8 & 47.4 (+10.5) & 79.0 \\
        \rowcolor{tblalt} \cellcolor{tblhead}
        & \cellcolor{tblsub}         
        & SAO   
        & 64.6 & 60.0 (-4.6) & 15.4 & 66.2 (+50.8) & 57.8 & 46.9 (-10.9) & 69.2 \\
        \cdashline{3-10}[0.8pt/2pt]
        \cellcolor{tblhead}
        & \cellcolor{tblsub} \multirow{-4}{*}{Sensitivity}
        & \textbf{Macro mean}
        & 55.3 & 48.5 (-6.8) & 47.7 & 73.4 (+25.7) & 55.4 & 57.3 (+2.0) & 77.7 \\
        \hhline{~---------}
        \rowcolor{tblalt} \cellcolor{tblhead}
        & \cellcolor{tblsub}
        & LAA      
        & 65.3 & 62.6 (-2.7) & 63.8 & 80.5 (+16.7) & 63.8 & 64.1 (+0.2) & 79.4 \\
        \cellcolor{tblhead}
        & \cellcolor{tblsub}         
        & CE          
        & 34.3 & 23.1 (-11.2) & 42.4 & 63.2 (+20.7) & 48.3 & 58.1 (+9.8) & 69.8 \\
        \rowcolor{tblalt} \cellcolor{tblhead}
        & \cellcolor{tblsub}         
        & SAO   
        & 67.7 & 61.4 (-6.3) & 25.3 & 76.1 (+50.8) & 60.7 & 55.6 (-5.1) & 77.6 \\
        \cdashline{3-10}[0.8pt/2pt]
        \cellcolor{tblhead}
        & \cellcolor{tblsub} \multirow{-4}{*}{F1 score}
        & \textbf{Macro mean}
        & 55.8 & 49.0 (-6.7) & 43.9 & 73.3 (+29.4) & 57.6 & 59.2 (+1.6) & 75.6 \\
        \hhline{~---------}
        \rowcolor{tblalt}  \cellcolor{tblhead}
        & \cellcolor{tblsub}
        & LAA     
        & 65.5 & 58.3 (-7.1) & 26.2 & 72.6 (+46.4) & 60.2 & 50.6 (-9.6) & 77.4 \\
        \cellcolor{tblhead}
        & \cellcolor{tblsub}         
        & CE          
        & 92.4 & 97.0 (+4.6) & 94.7 & 94.7 (0.0) & 97.6 & 97.6 (0.0) & 93.1 \\
        \rowcolor{tblalt} \cellcolor{tblhead}
        & \cellcolor{tblsub}         
        & SAO   
        & 80.0 & 72.9 (-7.1) & 95.3 & 94.1 (-1.2) & 74.4 & 82.9 (+8.5) & 92.9 \\
        \cdashline{3-10}[0.8pt/2pt]
        \cellcolor{tblhead}
        & \cellcolor{tblsub} \multirow{-4}{*}{Specificity}
        & \textbf{Macro mean}
        & 79.3 & 76.1 (-3.2) & 72.1 & 87.1 (+15.1) & 77.4 & 77.1 (-0.4) & 87.8 \\
        \hhline{~---------}
        \rowcolor{tblalt}  \cellcolor{tblhead}
        & \cellcolor{tblsub}
        & LAA     
        & 61.3 & 56.8 (-4.5) & 49.2 & 72.3 (+23.1) & 57.7 & 54.4 (-3.2) & 74.7 \\
        \cellcolor{tblhead}
        & \cellcolor{tblsub}         
        & CE          
        & 37.5 & 42.9 (+5.4) & 50.0 & 63.2 (+13.2) & 70.0 & 75.0 (+5.0) & 62.5 \\
        \rowcolor{tblalt}   \cellcolor{tblhead}
        & \cellcolor{tblsub}         
        & SAO 
        & 71.2 & 62.9 (-8.3) & 71.4 & 89.6 (+18.2) & 63.8 & 68.2 (+4.4) & 88.2 \\
        \cdashline{3-10}[0.8pt/2pt]
        \cellcolor{tblhead}\multirow{-20}{*}{\rotatebox{90}{\textbf{Internal cohorts}}}
        & \cellcolor{tblsub} \multirow{-4}{*}{PPV}
        & \textbf{Macro mean}
        & 56.7 & 54.2 (-2.5) & 56.9 & 75.0 (+18.1) & 63.8 & 65.9 (+2.0) & 75.1 \\
    \hline
        \rowcolor{tblalt}  \cellcolor{tblhead}
        & \cellcolor{tblsub}
        & LAA     
        & 60.7 & 61.3 (+0.7) & 60.0 & 76.7 (+16.7) & 59.6 & 64.0 (+4.4) & 82.7 \\
        \cellcolor{tblhead}
        & \cellcolor{tblsub}         
        & CE          
        & 86.7 & 89.3 (+2.7) & 86.7 & 90.7 (+4.0) & 86.0 & 89.7 (+3.7) & 90.0 \\
        \rowcolor{tblalt} \cellcolor{tblhead}
        & \cellcolor{tblsub}         
        & SAO   
        & 56.7 & 56.0 (-0.7) & 62.7 & 79.3 (+16.7) & 57.4 & 62.5 (+5.1) & 91.3 \\
        \cdashline{3-10}[0.8pt/2pt]
        \cellcolor{tblhead}
        & \cellcolor{tblsub} \multirow{-4}{*}{Accuracy}
        & \textbf{Macro mean}
        & 68.0 & 68.9 (+0.9) & 69.8 & 82.2 (+12.4) & 67.7 & 72.1 (+4.4) & 88.0 \\
        \hhline{~---------}
        \rowcolor{tblalt} \cellcolor{tblhead}
        & \cellcolor{tblsub}
        & LAA      
        & 34.9 & 40.9 (+6.1) & 72.7 & 75.8 (+3.0) & 40.7 & 57.4 (+16.7) & 75.8 \\
        \cellcolor{tblhead}
        & \cellcolor{tblsub}         
        & CE          
        & 47.4 & 31.6 (-15.8) & 26.3 & 57.9 (+31.6) & 33.3 & 50.0 (+16.7) & 84.2 \\
        \rowcolor{tblalt} \cellcolor{tblhead}
        & \cellcolor{tblsub}        
        & SAO   
        & 70.8 & 72.3 (+1.5) & 44.6 & 75.4 (+30.8) & 65.6 & 60.9 (-4.7) & 87.7 \\
        \cdashline{3-10}[0.8pt/2pt]
        \cellcolor{tblhead}
        & \cellcolor{tblsub} \multirow{-4}{*}{Sensitivity}
        & \textbf{Macro mean}
        & 51.0 & 48.3 (-2.7) & 47.9 & 69.7 (+21.8) & 46.6 & 56.1 (+9.5) & 82.6 \\
        \hhline{~---------}
        \rowcolor{tblalt} \cellcolor{tblhead}
        & \cellcolor{tblsub}
        & LAA      
        & 43.8 & 48.2 (+4.4) & 61.5 & 74.1 (+12.5) & 44.4 & 55.9 (+11.4) & 79.4 \\
        \cellcolor{tblhead}
        & \cellcolor{tblsub}        
        & CE          
        & 47.4 & 42.9 (-4.5) & 33.3 & 61.1 (+27.8) & 38.7 & 56.3 (+17.5) & 68.1 \\
        \rowcolor{tblalt}  \cellcolor{tblhead}
        & \cellcolor{tblsub}        
        & SAO  
        & 58.6 & 58.8 (+0.2) & 50.9 & 76.0 (+25.1) & 59.2 & 60.5 (+1.3) & 89.8 \\
        \cdashline{3-10}[0.8pt/2pt]
        \cellcolor{tblhead}
        & \cellcolor{tblsub} \multirow{-4}{*}{F1 score}
        & \textbf{Macro mean}
        & 49.9 & 49.9 (0.0) & 48.6 & 70.4 (+21.8) & 47.4 & 57.5 (+10.1) & 79.1 \\
        \hhline{~---------}
        \rowcolor{tblalt} \cellcolor{tblhead}
        & \cellcolor{tblsub}
        & LAA      
        & 81.0 & 77.4 (-3.6) & 50.0 & 77.4 (27.4) & 72.0 & 68.3 (-3.7) & 88.1 \\
        \cellcolor{tblhead}
        & \cellcolor{tblsub}         
        & CE          
        & 92.4 & 97.7 (+5.3) & 95.4 & 95.4 (0.0) & 94.1 & 95.8 (+1.7) & 90.8 \\
        \rowcolor{tblalt} \cellcolor{tblhead}
        & \cellcolor{tblsub}         
        & SAO   
        & 45.9 & 43.5 (-2.4) & 76.5 & 82.4 (+5.9) & 50.0 & 63.9 (+13.9) & 94.1 \\
        \cdashline{3-10}[0.8pt/2pt]
        \cellcolor{tblhead}
        & \cellcolor{tblsub} \multirow{-4}{*}{Specificity}
        & \textbf{Macro mean} 
        & 73.1 & 72.9 (-0.2) & 74.0 & 85.1 (+11.1) & 72.0 & 76.0 (+4.0) & 91.0 \\
        \hhline{~---------}
        \rowcolor{tblalt}\cellcolor{tblhead}
        & \cellcolor{tblsub}
        & LAA               
        & 59.0 & 58.7 (-0.3) & 53.3 & 72.5 (+19.1) & 48.9 & 54.4 (+5.5) & 83.3 \\
        \cellcolor{tblhead}
        & \cellcolor{tblsub}
        & CE          
        & 47.4 & 66.7 (+19.3) & 45.5 & 64.7 (+19.3) & 46.2 & 64.3 (+18.1) & 57.1 \\
        \rowcolor{tblalt}\cellcolor{tblhead}
        & \cellcolor{tblsub}     
        & SAO 
        & 50.0 & 49.5 (-0.5) & 59.2 & 76.6 (+17.4) & 53.9 & 60.0 (+6.2) & 91.9 \\
        \cdashline{3-10}[0.8pt/2pt]
        \cellcolor{tblhead}
        \multirow{-20}{*}{\rotatebox{90}{\textbf{External cohorts}}}
        & \cellcolor{tblsub} \multirow{-4}{*}{PPV}
        & \textbf{Macro mean} 
        & 52.1 & 58.3 (+6.2) & 52.7 & 71.2 (+18.6) & 49.6 & 59.6 (+9.9) & 77.5 \\
    \hline
    \hline
    \rowcolor{tblalt}\multicolumn{3}{l}{\textbf{Overall accuracy (\%)}} & 57.3 & 56.0 (-1.3) & 53.0 & 75.0 (+22.0) & 56.4 & 59.2 (+2.8) & 79.7 \\
    \multicolumn{3}{l}{\textbf{High diagnostic confidence, score 3 (\%)}} & 41.7 & 44.7 & 8.3 & 1.7 & 71.3 & 87.6 & - \\
    \rowcolor{tblalt}\multicolumn{3}{l}{\textbf{AI recommendation acceptance (\%)}} & - & 63.0 & - & 90.3 & - & 61.3 & - \\
    \multicolumn{3}{l}{\textbf{Incorrect → Correct (\%)}} & - & 7.7 & - & 27.3 & - & 9.2 & - \\
    \rowcolor{tblalt}\multicolumn{3}{l}{\textbf{Correct → Incorrect (\%)}} & - & 9.0 & - & 5.3 & - & 6.4 & - \\
    \hline
    \end{tabular}
    }
\end{table*}

\begin{table*}[!t]
\centering
\caption{\textbf{Performance of BrainGFM in multicenter clinical stroke-imaging validation.} Performance was evaluated on pooled task-eligible cases from two internal and four external centers. Higher PSNR, SSIM, Dice, ACC, F1, and AUC indicate better performance, whereas lower LPIPS and HD95 indicate better performance.}
\label{tab:stroke_multicenter_supp}
\resizebox{\linewidth}{!}
{
\begin{tabular}{clcccccccccccccccccccccccc}
\hline\rowcolor{tblhead}
& & \multicolumn{12}{c}{\textbf{Internal 2 cohorts}} & \multicolumn{12}{c}{\textbf{External 4 cohorts}} \\
\hhline{~~------------------------}\rowcolor{tblhead}
\multirow{-2}{*}{\textbf{Task}}
& \multirow{-2}{*}{\textbf{Method}}
& \multicolumn{3}{c}{\textbf{PSNR $\uparrow$}}
& \multicolumn{3}{c}{\textbf{SSIM $\uparrow$}}
& \multicolumn{3}{c}{\textbf{RMSE $\downarrow$}}
& \multicolumn{3}{c}{\textbf{LPIPS $\downarrow$}}
& \multicolumn{3}{c}{\textbf{PSNR $\uparrow$}}
& \multicolumn{3}{c}{\textbf{SSIM $\uparrow$}}
& \multicolumn{3}{c}{\textbf{RMSE $\downarrow$}}
& \multicolumn{3}{c}{\textbf{LPIPS $\downarrow$}}
\\
\hline
 \rowcolor{tblalt}\cellcolor{tblsub} & RegGAN
& \multicolumn{3}{c}{23.95} & \multicolumn{3}{c}{87.1} & \multicolumn{3}{c}{16.31} & \multicolumn{3}{c}{14.02} & \multicolumn{3}{c}{22.92} & \multicolumn{3}{c}{87.3} & \multicolumn{3}{c}{18.28} & \multicolumn{3}{c}{14.57} \\
 \rowcolor{tblalt}\cellcolor{tblsub} &
& \multicolumn{3}{c}{(21.24-26.66)} & \multicolumn{3}{c}{(80.9-93.2)} & \multicolumn{3}{c}{(15.99-16.63)} & \multicolumn{3}{c}{(5.38-22.66)} & \multicolumn{3}{c}{(20.40-25.44)} & \multicolumn{3}{c}{(80.9-93.7)} & \multicolumn{3}{c}{(17.26-19.29)} & \multicolumn{3}{c}{(6.53-22.61)} \\
\cellcolor{tblsub} & CycleGAN
& \multicolumn{3}{c}{24.14} & \multicolumn{3}{c}{87.4} & \multicolumn{3}{c}{15.93} & \multicolumn{3}{c}{14.15} & \multicolumn{3}{c}{23.29} & \multicolumn{3}{c}{87.1} & \multicolumn{3}{c}{17.57} & \multicolumn{3}{c}{13.80} \\
\cellcolor{tblsub} &
& \multicolumn{3}{c}{(20.48-27.80)} & \multicolumn{3}{c}{(80.7-94.1)} & \multicolumn{3}{c}{(15.62-16.23)} & \multicolumn{3}{c}{(5.71-22.59)} & \multicolumn{3}{c}{(19.34-27.24)} & \multicolumn{3}{c}{(80.3-93.9)} & \multicolumn{3}{c}{(16.59-18.55)} & \multicolumn{3}{c}{(5.02-22.58)} \\
 \rowcolor{tblalt}\cellcolor{tblsub} & MAE
& \multicolumn{3}{c}{25.34} & \multicolumn{3}{c}{88.7} & \multicolumn{3}{c}{13.87} & \multicolumn{3}{c}{14.17} & \multicolumn{3}{c}{24.52} & \multicolumn{3}{c}{89.1} & \multicolumn{3}{c}{15.07} & \multicolumn{3}{c}{16.10} \\
 \rowcolor{tblalt}\cellcolor{tblsub} &
& \multicolumn{3}{c}{(22.07-28.61)} & \multicolumn{3}{c}{(82.1-95.3)} & \multicolumn{3}{c}{(13.60-14.14)} & \multicolumn{3}{c}{(7.85-20.49)} & \multicolumn{3}{c}{(21.02-28.02)} & \multicolumn{3}{c}{(82.5-95.6)} & \multicolumn{3}{c}{(14.28-15.87)} & \multicolumn{3}{c}{(10.22-21.98)} \\
\cellcolor{tblsub} & DDIM
& \multicolumn{3}{c}{18.42} & \multicolumn{3}{c}{85.3} & \multicolumn{3}{c}{30.61} & \multicolumn{3}{c}{26.93} & \multicolumn{3}{c}{17.12} & \multicolumn{3}{c}{84.6} & \multicolumn{3}{c}{35.43} & \multicolumn{3}{c}{27.57} \\
\cellcolor{tblsub} &
& \multicolumn{3}{c}{(15.74-21.10)} & \multicolumn{3}{c}{(79.0-91.6)} & \multicolumn{3}{c}{(30.04-31.19)} & \multicolumn{3}{c}{(15.10-38.76)} & \multicolumn{3}{c}{(14.39-19.83)} & \multicolumn{3}{c}{(78.5-90.8)} & \multicolumn{3}{c}{(33.51-37.35)} & \multicolumn{3}{c}{(15.15-39.99)} \\
 \rowcolor{tblalt}\cellcolor{tblsub} & SDM
& \multicolumn{3}{c}{23.35} & \multicolumn{3}{c}{87.0} & \multicolumn{3}{c}{17.42} & \multicolumn{3}{c}{\textbf{11.79}} & \multicolumn{3}{c}{21.55} & \multicolumn{3}{c}{86.6} & \multicolumn{3}{c}{21.20} & \multicolumn{3}{c}{\textbf{11.20}} \\
 \rowcolor{tblalt}\cellcolor{tblsub} &
& \multicolumn{3}{c}{(19.79-26.90)} & \multicolumn{3}{c}{(80.5-93.4)} & \multicolumn{3}{c}{(17.08-17.75)} & \multicolumn{3}{c}{\textbf{(3.76-19.82)}} & \multicolumn{3}{c}{(17.87-25.21)} & \multicolumn{3}{c}{(80.0-93.1)} & \multicolumn{3}{c}{(20.07-22.34)} & \multicolumn{3}{c}{\textbf{(3.89-18.51)}} \\
\cellcolor{tblsub} & FlowMatching
& \multicolumn{3}{c}{23.63} & \multicolumn{3}{c}{86.6} & \multicolumn{3}{c}{16.90} & \multicolumn{3}{c}{13.97} & \multicolumn{3}{c}{24.10} & \multicolumn{3}{c}{86.3} & \multicolumn{3}{c}{16.05} & \multicolumn{3}{c}{15.02} \\
\cellcolor{tblsub} &
& \multicolumn{3}{c}{(20.34-26.92)} & \multicolumn{3}{c}{(79.6-93.6)} & \multicolumn{3}{c}{(16.57-17.23)} & \multicolumn{3}{c}{(1.75-26.19)} & \multicolumn{3}{c}{(20.74-27.46)} & \multicolumn{3}{c}{(78.8-93.7)} & \multicolumn{3}{c}{(15.13-16.96)} & \multicolumn{3}{c}{(2.43-27.61)} \\
 \rowcolor{tblalt}\cellcolor{tblsub} & CAR
& \multicolumn{3}{c}{21.35} & \multicolumn{3}{c}{79.6} & \multicolumn{3}{c}{21.92} & \multicolumn{3}{c}{14.90} & \multicolumn{3}{c}{20.83} & \multicolumn{3}{c}{79.0} & \multicolumn{3}{c}{23.10} & \multicolumn{3}{c}{15.61} \\
 \rowcolor{tblalt}\cellcolor{tblsub} &
& \multicolumn{3}{c}{(18.72-23.99)} & \multicolumn{3}{c}{(73.5-85.7)} & \multicolumn{3}{c}{(21.50-22.34)} & \multicolumn{3}{c}{(7.38-22.42)} & \multicolumn{3}{c}{(18.25-23.41)} & \multicolumn{3}{c}{(73.2-84.8)} & \multicolumn{3}{c}{(21.82-24.38)} & \multicolumn{3}{c}{(8.62-22.60)} \\
\cellcolor{tblsub} & \textbf{BrainGFM}
& \multicolumn{3}{c}{\textbf{26.63}} & \multicolumn{3}{c}{\textbf{90.7}} & \multicolumn{3}{c}{\textbf{11.94}} & \multicolumn{3}{c}{12.43} & \multicolumn{3}{c}{\textbf{26.90}} & \multicolumn{3}{c}{\textbf{89.6}} & \multicolumn{3}{c}{\textbf{11.65}} & \multicolumn{3}{c}{13.43} \\
\cellcolor{tblsub} \multirow{-16}{*}{\rotatebox{90}{\textbf{MRI sequence transformation}}} &
& \multicolumn{3}{c}{\textbf{(23.83-29.43)}} & \multicolumn{3}{c}{\textbf{(86.5-94.9)}} & \multicolumn{3}{c}{\textbf{(11.71-12.17)}} & \multicolumn{3}{c}{(6.71-18.15)} & \multicolumn{3}{c}{\textbf{(23.93-29.87)}} & \multicolumn{3}{c}{\textbf{(85.7-93.4)}} & \multicolumn{3}{c}{\textbf{(10.97-12.33)}} & \multicolumn{3}{c}{(7.31-19.55)} \\
\hline
\rowcolor{tblhead}
\textbf{Task}
& \textbf{Method}
& \multicolumn{6}{c}{\textbf{Dice $\uparrow$}}
& \multicolumn{6}{c}{\textbf{HD95 $\downarrow$}}
& \multicolumn{6}{c}{\textbf{Dice $\uparrow$}}
& \multicolumn{6}{c}{\textbf{HD95 $\downarrow$}}
\\
\hline
 \rowcolor{tblalt}\cellcolor{tblsub} & UNet (few)
& \multicolumn{6}{c}{58.0} & \multicolumn{6}{c}{31.6} & \multicolumn{6}{c}{50.8} & \multicolumn{6}{c}{33.0} \\
 \rowcolor{tblalt}\cellcolor{tblsub} & \textit{60 cases}
& \multicolumn{6}{c}{(57.3-58.7)} & \multicolumn{6}{c}{(31.0-32.2)} & \multicolumn{6}{c}{(48.3-53.3)} & \multicolumn{6}{c}{(30.9-35.1)} \\
\cellcolor{tblsub} & CAR
& \multicolumn{6}{c}{69.6} & \multicolumn{6}{c}{19.9} & \multicolumn{6}{c}{67.2} & \multicolumn{6}{c}{23.1} \\
\cellcolor{tblsub} & 
& \multicolumn{6}{c}{(68.8-70.3)} & \multicolumn{6}{c}{(19.0-20.9)} & \multicolumn{6}{c}{(66.4-67.9)} & \multicolumn{6}{c}{(22.0-24.2)} \\
 \rowcolor{tblalt}\cellcolor{tblsub} & MedSAM
& \multicolumn{6}{c}{72.0} & \multicolumn{6}{c}{13.0} & \multicolumn{6}{c}{70.3} & \multicolumn{6}{c}{18.8} \\
 \rowcolor{tblalt}\cellcolor{tblsub} &
& \multicolumn{6}{c}{(71.2-72.9)} & \multicolumn{6}{c}{(12.4-13.6)} & \multicolumn{6}{c}{(68.2-72.5)} & \multicolumn{6}{c}{(17.0-20.5)} \\
\cellcolor{tblsub} & MAE
& \multicolumn{6}{c}{70.5} & \multicolumn{6}{c}{15.3} & \multicolumn{6}{c}{69.9} & \multicolumn{6}{c}{20.5} \\
\cellcolor{tblsub} &
& \multicolumn{6}{c}{(69.6-71.3)} & \multicolumn{6}{c}{(14.8-15.9)} & \multicolumn{6}{c}{(67.6-72.2)} & \multicolumn{6}{c}{(18.6-22.5)} \\
 \rowcolor{tblalt}\cellcolor{tblsub} & BrainIAC
& \multicolumn{6}{c}{69.9} & \multicolumn{6}{c}{15.2} & \multicolumn{6}{c}{70.6} & \multicolumn{6}{c}{19.8} \\
 \rowcolor{tblalt}\cellcolor{tblsub} &
& \multicolumn{6}{c}{(69.2-70.6)} & \multicolumn{6}{c}{(14.7-15.7)} & \multicolumn{6}{c}{(68.4-72.8)} & \multicolumn{6}{c}{(18.4-21.2)} \\
\cellcolor{tblsub} & \textbf{BrainGFM}
& \multicolumn{6}{c}{\textbf{78.1}} & \multicolumn{6}{c}{\textbf{10.0}} & \multicolumn{6}{c}{\textbf{76.9}} & \multicolumn{6}{c}{\textbf{11.7}} \\
\cellcolor{tblsub} & 
& \multicolumn{6}{c}{\textbf{(77.4-78.8)}} & \multicolumn{6}{c}{\textbf{(9.6-10.4)}} & \multicolumn{6}{c}{\textbf{(74.7-79.0)}} & \multicolumn{6}{c}{\textbf{(10.4-13.0)}} \\
\cdashline{2-26}[0.8pt/2pt]
 \rowcolor{tblalt}\cellcolor{tblsub} & {\color{gray}UNet (full)}
& \multicolumn{6}{c}{{\color{gray}69.3}} & \multicolumn{6}{c}{{\color{gray}14.2}} & \multicolumn{6}{c}{{\color{gray}65.1}} & \multicolumn{6}{c}{{\color{gray}24.1}} \\
 \rowcolor{tblalt}\cellcolor{tblsub} \multirow{-14}{*}{\rotatebox{90}{\textbf{Lesion segmentation}}}
& {\color{gray}\textit{200 cases}}
& \multicolumn{6}{c}{{\color{gray}(68.5-70.2)}} & \multicolumn{6}{c}{{\color{gray}(13.6-14.7)}} & \multicolumn{6}{c}{{\color{gray}(62.7-67.6)}} & \multicolumn{6}{c}{{\color{gray}(22.0-26.3)}} \\
\hline
\rowcolor{tblhead}
\textbf{Task}
& \textbf{Method}
& \multicolumn{4}{c}{\textbf{ACC $\uparrow$}}
& \multicolumn{4}{c}{\textbf{F1 $\uparrow$}}
& \multicolumn{4}{c}{\textbf{AUC $\uparrow$}}
& \multicolumn{4}{c}{\textbf{ACC $\uparrow$}}
& \multicolumn{4}{c}{\textbf{F1 $\uparrow$}}
& \multicolumn{4}{c}{\textbf{AUC $\uparrow$}}
\\
\hline
 \rowcolor{tblalt}\cellcolor{tblsub} & ResNet50 (few)
& \multicolumn{4}{c}{58.0} & \multicolumn{4}{c}{47.1} & \multicolumn{4}{c}{62.6} & \multicolumn{4}{c}{53.1} & \multicolumn{4}{c}{45.5} & \multicolumn{4}{c}{61.6} \\
 \rowcolor{tblalt}\cellcolor{tblsub} & \textit{60 cases}
& \multicolumn{4}{c}{(55.6-60.5)} & \multicolumn{4}{c}{(43.8-50.4)} & \multicolumn{4}{c}{(58.8-66.4)} & \multicolumn{4}{c}{(50.2-56.0)} & \multicolumn{4}{c}{(43.0-48.0)} & \multicolumn{4}{c}{(58.4-64.8)} \\
\cellcolor{tblsub} & CAR
& \multicolumn{4}{c}{68.9} & \multicolumn{4}{c}{65.3} & \multicolumn{4}{c}{81.5} & \multicolumn{4}{c}{65.3} & \multicolumn{4}{c}{60.9} & \multicolumn{4}{c}{77.5} \\
\cellcolor{tblsub} &
& \multicolumn{4}{c}{(65.3-72.4)} & \multicolumn{4}{c}{(62.7-67.9)} & \multicolumn{4}{c}{(79.7-83.4)} & \multicolumn{4}{c}{(63.0-67.7)} & \multicolumn{4}{c}{(57.2-64.5)} & \multicolumn{4}{c}{(74.7-80.3)} \\
 \rowcolor{tblalt}\cellcolor{tblsub} & MedSAM
& \multicolumn{4}{c}{67.9} & \multicolumn{4}{c}{62.1} & \multicolumn{4}{c}{78.4} & \multicolumn{4}{c}{66.0} & \multicolumn{4}{c}{56.9} & \multicolumn{4}{c}{70.8} \\
 \rowcolor{tblalt}\cellcolor{tblsub} &
& \multicolumn{4}{c}{(65.9-70.0)} & \multicolumn{4}{c}{(58.5-65.6)} & \multicolumn{4}{c}{(75.4-81.4)} & \multicolumn{4}{c}{(63.9-68.0)} & \multicolumn{4}{c}{(53.5-60.4)} & \multicolumn{4}{c}{(67.8-73.7)} \\
\cellcolor{tblsub} & MAE
& \multicolumn{4}{c}{76.1} & \multicolumn{4}{c}{68.2} & \multicolumn{4}{c}{82.5} & \multicolumn{4}{c}{71.5} & \multicolumn{4}{c}{64.3} & \multicolumn{4}{c}{79.6} \\
\cellcolor{tblsub} & 
& \multicolumn{4}{c}{(73.0-79.3)} & \multicolumn{4}{c}{(64.4-72.0)} & \multicolumn{4}{c}{(79.9-85.1)} & \multicolumn{4}{c}{(68.4-74.6)} & \multicolumn{4}{c}{(62.2-66.5)} & \multicolumn{4}{c}{(76.1-83.1)} \\
 \rowcolor{tblalt} \cellcolor{tblsub} & BrainIAC
& \multicolumn{4}{c}{71.6} & \multicolumn{4}{c}{66.2} & \multicolumn{4}{c}{79.7} & \multicolumn{4}{c}{68.8} & \multicolumn{4}{c}{62.2} & \multicolumn{4}{c}{74.7} \\
\rowcolor{tblalt}\cellcolor{tblsub} & 
& \multicolumn{4}{c}{(67.9-75.3)} & \multicolumn{4}{c}{(63.7-68.8)} & \multicolumn{4}{c}{(77.6-81.7)} & \multicolumn{4}{c}{(66.8-70.8)} & \multicolumn{4}{c}{(60.4-64.1)} & \multicolumn{4}{c}{(72.4-77.0)} \\
\cellcolor{tblsub} & \textbf{BrainGFM}
& \multicolumn{4}{c}{\textbf{81.2}} & \multicolumn{4}{c}{\textbf{73.8}} & \multicolumn{4}{c}{\textbf{87.4}} & \multicolumn{4}{c}{\textbf{77.0}} & \multicolumn{4}{c}{\textbf{70.4}} & \multicolumn{4}{c}{\textbf{84.5}} \\
\cellcolor{tblsub} &
& \multicolumn{4}{c}{\textbf{(79.1-83.4)}} & \multicolumn{4}{c}{\textbf{(70.3-77.2)}} & \multicolumn{4}{c}{\textbf{(85.0-89.8)}} & \multicolumn{4}{c}{\textbf{(74.5-79.5)}} & \multicolumn{4}{c}{\textbf{(67.1-73.6)}} & \multicolumn{4}{c}{\textbf{(82.3-86.6)}} \\
\cdashline{2-26}[0.8pt/2pt]
\rowcolor{tblalt} \cellcolor{tblsub} & {\color{gray}ResNet50 (full)}
& \multicolumn{4}{c}{{\color{gray}73.4}} & \multicolumn{4}{c}{{\color{gray}63.5}} & \multicolumn{4}{c}{{\color{gray}77.5}} & \multicolumn{4}{c}{{\color{gray}71.6}} & \multicolumn{4}{c}{{\color{gray}60.0}} & \multicolumn{4}{c}{{\color{gray}75.6}} \\
 \rowcolor{tblalt} \cellcolor{tblsub} \multirow{-14}{*}{\rotatebox{90}{\textbf{TOAST stroke subtyping}}}
&{\color{gray}\textit{200 cases}}
& \multicolumn{4}{c}{{\color{gray}(70.2-76.6)}} & \multicolumn{4}{c}{{\color{gray}(60.7-66.2)}} & \multicolumn{4}{c}{{\color{gray}(75.2-79.8)}} & \multicolumn{4}{c}{{\color{gray}(68.5-74.7)}} & \multicolumn{4}{c}{{\color{gray}(58.1-62.0)}} & \multicolumn{4}{c}{{\color{gray}(72.4-78.9)}} \\
\hline
\end{tabular}
}
\end{table*}

\begin{table*}[!t]
\centering
\caption{\textbf{Performance of BrainGFM in low-data segmentation across multiple datasets.} Values are Dice scores (\%) with 95\% CIs. For datasets with multiple target regions, scores are averaged across regions. Higher Dice indicates better performance. The best low-data result is shown in bold, and the strongest competing low-data baseline is underlined. UNet (full) is reported as a full-data reference and is not included in the low-data ranking.}
\label{tab:segmentation_multidataset}
\resizebox{\textwidth}{!}{
\begin{tabular}{lccccc}
\hline
\rowcolor{tblhead}
& \multicolumn{1}{c}{TuS-BraTS21-T1} & \multicolumn{1}{c}{TuS-TCIA-T1} & \multicolumn{1}{c}{TuS-TCGA-T1} & \multicolumn{1}{c}{TuS-BraTS21-T1ce} & \multicolumn{1}{c}{TuS-TCIA-T1ce} \\
\hhline{~-----}
\rowcolor{tblhead}
Method & Dice $\uparrow$ & Dice $\uparrow$ & Dice $\uparrow$ & Dice $\uparrow$ & Dice $\uparrow$ \\\rowcolor{tblalt}
\hline
UNet (few)
& 39.6 (36.1-43.1) & 43.1 (39.8-46.4) & 43.2 (39.6-46.8) & 58.2 (55.7-60.7) & 60.4 (58.1-62.7) \\
CAR
& 46.3 (42.9-49.7) & 49.0 (45.8-52.2) & 45.1 (41.6-48.6) & 70.0 (67.9-72.1) & 68.9 (66.9-70.9) \\\rowcolor{tblalt}
MedSAM
& 48.7 (45.6-51.8) & \underline{51.2 (48.2-54.2)} & 46.8 (43.5-50.1) & \underline{73.5 (71.6-75.4)} & 70.9 (69.0-72.8) \\
MAE
& 47.9 (44.6-51.2) & 49.8 (46.7-52.9) & 45.9 (42.4-49.4) & 71.4 (69.3-73.5) & 69.8 (67.7-71.9) \\\rowcolor{tblalt}
BrainIAC
& \underline{49.6 (46.6-52.6)} & 50.6 (47.7-53.5) & \underline{47.6 (44.4-50.8)} & 72.9 (71.1-74.7) & \underline{71.4 (69.6-73.2)} \\
BrainGFM
& \textbf{51.5 (48.6-54.4)} & \textbf{52.9 (50.1-55.7)} & \textbf{49.8 (46.7-52.9)} & \textbf{75.4 (73.6-77.2)} & \textbf{74.9 (73.1-76.7)} \\\rowcolor{tblalt}
\hline
{\color{gray}UNet (full)}
& {\color{gray}52.5 (49.4-55.6)} & {\color{gray}54.6 (51.7-57.5)} & {\color{gray}48.4 (45.1-51.7)} & {\color{gray}73.6 (71.7-75.5)} & {\color{gray}73.0 (71.1-74.9)} \\
\hline
\rowcolor{tblhead}
& \multicolumn{1}{c}{TuS-TCGA-T1ce} & \multicolumn{1}{c}{TuS-BraTS21-FLAIR} & \multicolumn{1}{c}{TuS-TCIA-FLAIR} & \multicolumn{1}{c}{TuS-TCGA-FLAIR} & \multicolumn{1}{c}{TuS-BraTS21-Multi} \\
\hhline{~-----}
\rowcolor{tblhead}
Method & Dice $\uparrow$ & Dice $\uparrow$ & Dice $\uparrow$ & Dice $\uparrow$ & Dice $\uparrow$ \\\rowcolor{tblalt}
\hline
UNet (few)
& 57.1 (54.6-59.6) & 47.3 (44.3-50.3) & 51.7 (48.8-54.6) & 48.7 (45.6-51.8) & 55.2 (53.3-57.1) \\
CAR
& 66.6 (64.3-68.9) & 53.6 (50.7-56.5) & 56.3 (53.7-58.9) & 53.2 (50.3-56.1) & 61.9 (60.1-63.7) \\\rowcolor{tblalt}
MedSAM
& \underline{72.2 (70.2-74.2)} & 58.6 (56.1-61.1) & \underline{64.3 (62.0-66.6)} & 60.8 (58.3-63.3) & 77.0 (75.5-78.5) \\
MAE
& 69.7 (67.6-71.8) & 57.9 (55.2-60.6) & 62.2 (59.7-64.7) & \underline{61.2 (58.6-63.8)} & 73.1 (71.4-74.8) \\\rowcolor{tblalt}
BrainIAC
& 71.7 (69.8-73.6) & \underline{59.6 (57.2-62.0)} & 63.5 (61.3-65.7) & 60.5 (58.1-62.9) & \textbf{\underline{77.6 (76.2-79.0)}} \\
BrainGFM
& \textbf{73.4 (71.5-75.3)} & \textbf{61.5 (59.1-63.9)} & \textbf{65.9 (63.7-68.1)} & \textbf{63.9 (61.5-66.3)} & 76.4 (75.0-77.8) \\\rowcolor{tblalt}
\hline
{\color{gray}UNet (full)}
& {\color{gray}72.7 (70.7-74.7)} & {\color{gray}63.3 (61.0-65.6)} & {\color{gray}65.8 (63.6-68.0)} & {\color{gray}63.5 (61.1-65.9)} & {\color{gray}76.1 (74.6-77.6)} \\
\hline
\rowcolor{tblhead}
& \multicolumn{1}{c}{TuS-TCIA-Multi} & \multicolumn{1}{c}{TuS-TCGA-Multi} & \multicolumn{1}{c}{TuS-BRISC25-T1} & \multicolumn{1}{c}{TiS-MRBrainS13-Multi} & \multicolumn{1}{c}{TiS-IBSR18-T1} \\
\hhline{~-----}
\rowcolor{tblhead}
Method & Dice $\uparrow$ & Dice $\uparrow$ & Dice $\uparrow$ & Dice $\uparrow$ & Dice $\uparrow$ \\\rowcolor{tblalt}
\hline
UNet (few)
& 56.9 (55.0-58.8) & 55.1 (53.1-57.1) & 57.4 (54.6-60.2) & 72.3 (69.2-75.4) & 74.5 (71.6-77.4) \\
CAR
& 63.9 (61.6-66.2) & 62.1 (60.2-64.0) & 68.1 (65.7-70.5) & 79.6 (78.0-81.2) & 81.8 (80.0-83.6) \\\rowcolor{tblalt}
MedSAM
& \underline{78.0 (76.6-79.4)} & 74.3 (72.6-76.0) & \underline{75.3 (73.4-77.2)} & 87.2 (85.8-88.6) & 89.0 (87.3-90.7) \\
MAE
& 73.3 (71.6-75.0) & 71.5 (69.7-73.3) & 72.5 (70.3-74.7) & \underline{88.6 (87.3-89.9)} & \underline{91.2 (89.6-92.8)} \\\rowcolor{tblalt}
BrainIAC
& 77.4 (76.1-78.7) & \textbf{\underline{74.6 (73.0-76.2)}} & - & 88.2 (86.5-89.9) & 90.5 (88.6-92.4) \\
BrainGFM
& \textbf{78.1 (76.8-79.4)} & 74.1 (72.6-75.6) & \textbf{77.9 (75.8-80.0)} & \textbf{90.4 (88.6-92.2)} & \textbf{93.5 (91.4-95.6)} \\\rowcolor{tblalt}
\hline
{\color{gray}UNet (full)}
& {\color{gray}81.3 (79.8-82.8)} & {\color{gray}77.9 (76.2-79.6)} & {\color{gray}73.6 (71.5-75.7)} & {\color{gray}85.1 (83.5-86.7)} & {\color{gray}89.1 (87.5-90.7)} \\
\hline
\rowcolor{tblhead}
& \multicolumn{1}{c}{TiS-iSeg17-T1} & \multicolumn{1}{c}{HVS-BHSD-CT} & \multicolumn{1}{c}{HVS-INSTANCE22-CT} & \multicolumn{1}{c}{HVS-TopBrain-CT} & \multicolumn{1}{c}{HVS-CT-ICH20-CT} \\
\hhline{~-----}
\rowcolor{tblhead}
Method & Dice $\uparrow$ & Dice $\uparrow$ & Dice $\uparrow$ & Dice $\uparrow$ & Dice $\uparrow$ \\\rowcolor{tblalt}
\hline
UNet (few)
& 76.3 (73.9-78.7) & 46.2 (42.4-50.0) & 48.5 (44.9-52.1) & 50.8 (47.3-54.3) & 45.2 (43.1-47.3) \\
CAR
& 83.1 (81.7-84.5) & 65.4 (62.2-68.6) & 67.2 (64.1-70.3) & 68.9 (65.9-71.9) & 65.9 (62.5-69.3) \\\rowcolor{tblalt}
MedSAM
& 91.7 (90.2-93.2) & \underline{69.1 (66.2-72.0)} & \underline{70.0 (67.2-72.8)} & 70.8 (68.1-73.5) & \underline{69.7 (67.4-72.0)} \\
MAE
& 91.4 (90.1-92.7) & 67.0 (63.9-70.1) & 68.9 (65.9-71.9) & \underline{71.2 (68.3-74.1)} & 68.2 (64.7-71.7) \\\rowcolor{tblalt}
BrainIAC
& \underline{92.3 (90.6-94.0)} & 61.3 (58.6-64.0) & 62.6 (60.0-65.2) & 63.4 (60.9-65.9) & 59.1 (56.0-62.2) \\
BrainGFM
& \textbf{94.3 (92.6-96.0)} & \textbf{71.5 (68.8-74.2)} & \textbf{75.1 (72.5-77.7)} & \textbf{76.3 (73.8-78.8)} & \textbf{70.6 (67.5-73.7)} \\\rowcolor{tblalt}
\hline
{\color{gray}UNet (full)}
& {\color{gray}90.9 (88.8-93.0)} & {\color{gray}70.1 (67.3-72.9)} & {\color{gray}71.4 (68.7-74.1)} & {\color{gray}72.8 (70.2-75.4)} & {\color{gray}72.2 (69.4-75.0)} \\
\hline
\end{tabular}%
}
\end{table*}

\begin{table*}[!t]
\centering
\caption{\textbf{Performance of BrainGFM in low-data classification across multiple datasets.} Values are ACC, F1, and AUC (\%) with 95\% CIs. Multiclass AUC is macro-averaged using one-vs-rest. Bold and underlined values indicate the best low-data result and strongest competing baseline. ResNet50 (full) is a full-data reference excluded from ranking. Dashes indicate unavailable results.}
\label{tab:classification_multidataset}
\resizebox{\textwidth}{!}{%
\begin{tabular}{lcccccc}
\hline
\rowcolor{tblhead}
& \multicolumn{3}{c}{SeC-BraTS23-\#} & \multicolumn{3}{c}{TuC-Br35H-Unknow} \\
\hhline{~------}
\rowcolor{tblhead}
Method & ACC $\uparrow$ & F1 $\uparrow$ & AUC $\uparrow$ & ACC $\uparrow$ & F1 $\uparrow$ & AUC $\uparrow$ \\\rowcolor{tblalt}
\hline
ResNet50 (few) & 83.9 (81.7-86.0) & 84.0 (81.9-86.1) & 91.6 (89.9-93.3) & 87.3 (85.4-89.3) & 87.2 (85.2-89.3) & 93.9 (92.2-95.7) \\
CAR & 85.1 (83.1-87.0) & 85.1 (83.1-87.1) & 92.7 (91.1-94.3) & 89.4 (87.3-91.6) & 88.7 (86.5-90.8) & 94.8 (93.4-96.3) \\\rowcolor{tblalt}
MedSAM & 85.3 (83.1-87.5) & 85.4 (83.4-87.4) & 92.8 (91.3-94.4) & 89.2 (87.2-91.2) & 89.0 (87.0-91.0) & 94.77 (93.4-96.1) \\
MAE & 87.1 (85.1-89.2) & 86.5 (84.6-88.4) & 93.76 (92.3-95.2) & \underline{89.8} \underline{(87.8-91.7)} & \underline{89.7} \underline{(87.6-91.9)} & \underline{95.6} \underline{(94.4-96.9)} \\\rowcolor{tblalt}
BrainIAC & \textbf{\underline{91.0}} \textbf{\underline{(89.1-93.0)}} & \textbf{\underline{90.3}} \textbf{\underline{(88.2-92.3)}} & \textbf{\underline{97.8}} \textbf{\underline{(97.0-98.7)}} & 89.6 (87.5-91.8) & 89.2 (87.1-91.2) & 95.2 (94.1-96.3) \\
\textbf{BrainGFM} & 90.2 (88.3-92.2) & 90.19 (88.2-92.2) & 97.7 (97.0-98.4) & \textbf{91.1} \textbf{(89.0-93.2)} & \textbf{90.5} \textbf{(88.4-92.6)} & \textbf{96.4} \textbf{(95.6-97.2)} \\\rowcolor{tblalt}
\hline
{\color{gray}ResNet50 (full)} & {\color{gray}94.0 (92.1-95.9)} & {\color{gray}90.8 (88.9-92.7)} & {\color{gray}98.3 (97.0-99.6)} & {\color{gray}90.6 (88.2-92.9)} & {\color{gray}89.9 (87.5-92.3)} & {\color{gray}95.8 (94.2-97.5)} \\
\hline
\rowcolor{tblhead}
& \multicolumn{3}{c}{TuC-BRISC25-T1} & \multicolumn{3}{c}{HaC-INSTANCE22-CT} \\
\hhline{~------}
\rowcolor{tblhead}
Method & ACC $\uparrow$ & F1 $\uparrow$ & AUC $\uparrow$ & ACC $\uparrow$ & F1 $\uparrow$ & AUC $\uparrow$ \\\rowcolor{tblalt}
\hline
ResNet50 (few) & 81.95 (79.9-84.0) & 82.0 (79.9-84.1) & 90.5 (88.7-92.4) & 84.7 (82.6-86.8) & 84.2 (82.0-86.3) & 92.2 (90.5-93.8) \\
CAR & 84.1 (81.9-86.3) & 83.6 (81.5-85.6) & 92.2 (90.5-93.8) & 85.6 (83.4-87.9) & 85.0 (82.8-87.3) & 93.4 (91.9-94.9) \\\rowcolor{tblalt}
MedSAM & 83.8 (81.6-86.1) & 83.3 (81.2-85.5) & 91.68 (90.0-93.4) & \underline{87.4} \underline{(85.3-89.4)} & \underline{87.0} \underline{(85.0-89.0)} & \underline{94.8} \underline{(93.6-96.1)} \\
MAE & \underline{85.3} \underline{(83.2-87.5)} & \underline{84.9} \underline{(82.9-86.8)} & \underline{93.2} \underline{(91.7-94.7)} & 86.7 (84.6-88.9) & 86.25 (84.1-88.4) & 94.3 (92.9-95.6) \\\rowcolor{tblalt}
BrainIAC & - & - & - & 85.6 (83.5-87.8) & 85.6 (83.5-87.7) & 93.6 (92.6-94.6) \\
\textbf{BrainGFM} & \textbf{89.4} \textbf{(87.4-91.3)} & \textbf{89.0} \textbf{(86.8-91.2)} & \textbf{97.1} \textbf{(96.8-97.5)} & \textbf{88.1} \textbf{(85.9-90.3)} & \textbf{87.9} \textbf{(85.9-90.0)} & \textbf{95.7} \textbf{(94.7-96.6)} \\\rowcolor{tblalt}
\hline
{\color{gray}ResNet50 (full)} & {\color{gray}93.5 (92.4-94.7)} & {\color{gray}92.4 (91.3-93.6)} & {\color{gray}97.9 (96.7-99.2)} & {\color{gray}87.5 (85.4-89.7)} & {\color{gray}87.4 (85.2-89.5)} & {\color{gray}95.1 (93.1-97.1)} \\
\hline
\rowcolor{tblhead}
& \multicolumn{3}{c}{HaC-Head CT-hemorrhage-CT} & \multicolumn{3}{c}{HaC-RSNA ICH-CT} \\
\hhline{~------}
\rowcolor{tblhead}
Method & ACC $\uparrow$ & F1 $\uparrow$ & AUC $\uparrow$ & ACC $\uparrow$ & F1 $\uparrow$ & AUC $\uparrow$ \\\rowcolor{tblalt}
\hline
ResNet50 (few) & 86.4 (84.6-88.2) & 85.7 (83.9-87.5) & 93.1 (91.4-94.8) & 81.2 (78.8-83.5) & 81.3 (79.0-83.6) & 90.3 (88.3-92.2) \\
CAR & 87.9 (85.9-89.8) & 87.42 (85.6-89.3) & 94.6 (93.2-95.9) & 84.17 (81.8-86.5) & 83.43 (81.1-85.7) & 91.75 (90.0-93.5) \\\rowcolor{tblalt}
MedSAM & \textbf{\underline{89.0}} \textbf{\underline{(87.0-91.0)}} & \underline{88.4} \underline{(86.6-90.2)} & \underline{95.1} \underline{(94.0-96.3)} & \underline{85.4} \underline{(83.3-87.5)} & \underline{85.3} \underline{(83.1-87.5)} & \underline{93.9} \underline{(92.5-95.4)} \\
MAE & 88.6 (86.8-90.4) & 87.9 (85.9-89.8) & 94.9 (93.8-96.1) & 84.8 (82.5-87.2) & 84.8 (82.7-87.0) & 93.2 (91.7-94.8) \\\rowcolor{tblalt}
BrainIAC & 87.8 (86.0-89.5) & 87.6 (85.6-89.5) & 94.7 (93.4-96.0) & 83.9 (81.5-86.3) & 83.8 (81.5-86.1) & 91.9 (90.2-93.6) \\
\textbf{BrainGFM} & 88.6 (86.7-90.5) & \textbf{88.4} \textbf{(86.5-90.4)} & \textbf{95.4} \textbf{(94.2-96.5)} & \textbf{88.0} \textbf{(85.8-90.3)} & \textbf{87.7} \textbf{(85.5-89.8)} & \textbf{95.9} \textbf{(94.7-97.0)} \\\rowcolor{tblalt}
\hline
{\color{gray}ResNet50 (full)} & {\color{gray}91.9 (89.1-94.8)} & {\color{gray}92.0 (87.9-96.2)} & {\color{gray}96.1 (93.8-98.5)} & {\color{gray}87.5 (85.4-89.5)} & {\color{gray}87.1 (85.1-89.2)} & {\color{gray}95.3 (93.9-96.7)} \\
\hline
\rowcolor{tblhead}
& \multicolumn{3}{c}{NeC-ABIDE-II-3D-T1} & \multicolumn{3}{c}{NeC-Alzheimer-3D-T1} \\
\hhline{~------}
\rowcolor{tblhead}
Method & ACC $\uparrow$ & F1 $\uparrow$ & AUC $\uparrow$ & ACC $\uparrow$ & F1 $\uparrow$ & AUC $\uparrow$ \\\rowcolor{tblalt}
\hline
ResNet50 (fell) & 70.2 (66.8-73.6) & 69.5 (66.1-72.9) & 79.9 (76.6-83.1) & 73.8 (70.4-77.3) & 74.0 (70.5-77.5) & 83.6 (80.5-86.7) \\
CAR & 72.6 (69.0-76.2) & 72.0 (68.4-75.5) & 82.2 (78.9-85.5) & 79.0 (75.7-82.3) & 78.2 (74.9-81.6) & 87.3 (83.9-90.7) \\\rowcolor{tblalt}
MedSAM & 71.7 (68.2-75.1) & 71.0 (67.5-74.6) & 81.2 (78.0-84.5) & 77.1 (73.7-80.5) & 76.7 (73.3-80.0) & 85.44 (82.1-88.8) \\
MAE & 72.5 (68.9-76.0) & 72.7 (69.3-76.1) & 82.7 (79.5-86.0) & 78.7 (75.3-82.2) & 78.79 (75.4-82.2) & 87.41 (84.0-90.8) \\\rowcolor{tblalt}
BrainIAC & \textbf{\underline{75.5}} \textbf{\underline{(72.2-78.8)}} & \textbf{\underline{75.0}} \textbf{\underline{(71.6-78.4)}} & \textbf{\underline{85.4}} \textbf{\underline{(82.1-88.7)}} & \textbf{\underline{80.2}} \textbf{\underline{(77.0-83.5)}} & \textbf{\underline{80.2}} \textbf{\underline{(76.9-83.4)}} & \textbf{\underline{89.1}} \textbf{\underline{(85.7-92.6)}} \\
\textbf{BrainGFM} & 74.9 (71.5-78.4) & 74.7 (71.3-78.2) & 85.2 (81.9-88.5) & 79.6 (76.3-82.9) & 79.7 (76.3-83.0) & 88.8 (85.3-92.2) \\\rowcolor{tblalt}
\hline
{\color{gray}ResNet50 (full)} & {\color{gray}75.9 (72.5-79.4)} & {\color{gray}75.5 (72.0-79.1)} & {\color{gray}85.9 (82.2-89.5)} & {\color{gray}87.0 (83.4-90.7)} & {\color{gray}85.2 (81.3-89.0)} & {\color{gray}91.4 (88.0-94.7)} \\
\hline
\end{tabular}%
}

\end{table*}

\begin{table*}[!t]
\centering
\caption{\textbf{Performance of BrainGFM in medical-image enhancement across multiple datasets.} Values are reported with 95\% confidence intervals. Higher PSNR and SSIM indicate better performance, whereas lower RMSE and LPIPS indicate better performance. The best result is shown in bold, and the strongest competing baseline is underlined.}
\label{tab:imaging_multidataset}
\resizebox{\textwidth}{!}{%
\begin{tabular}{lcccccccc}
\hline
\rowcolor{tblhead}
& \multicolumn{4}{c}{LFMR-BraTS21-T1} & \multicolumn{4}{c}{LFMR-BraTS21-T2} \\
\hhline{~--------}
\rowcolor{tblhead}
Method & PSNR $\uparrow$ & SSIM $\uparrow$ & RMSE $\downarrow$ & LPIPS $\downarrow$ & PSNR $\uparrow$ & SSIM $\uparrow$ & RMSE $\downarrow$ & LPIPS $\downarrow$ \\\rowcolor{tblalt}
\hline
RegGAN & 33.66 (29.11-38.21) & 94.4 (90.6-98.2) & 29.11 (0.27-57.95) & 5.17 (1.12-9.22) & 34.03 (29.45-38.61) & 94.0 (90.2-97.8) & 37.21 (7.62-66.80) & 6.46 (1.99-10.93) \\
CycleGAN & 37.78 (34.10-41.46) & 97.2 (95.4-99.0) & 17.28 (2.45-32.11) & 1.71 (0.30-3.12) & 38.16 (34.54-41.78) & 97.4 (95.9-98.9) & 22.27 (5.17-39.37) & 1.31 (0.36-2.26) \\\rowcolor{tblalt}
MAE & \underline{38.09} \underline{(34.69-41.49)} & \underline{97.4} \underline{(95.8-99.0)} & 16.49 (1.49-31.49) & 1.88 (0.67-3.09) & 38.19 (34.84-41.54) & \underline{97.6} \underline{(96.2-98.9)} & 22.08 (5.26-38.90) & 1.65 (0.69-2.61) \\
DDIM & 37.10 (33.48-40.72) & 96.7 (94.7-98.7) & 18.59 (3.01-34.17) & \underline{1.28} \underline{(0.35-2.21)} & 37.88 (34.40-41.36) & 97.4 (95.9-98.9) & 22.95 (5.13-40.77) & \underline{1.08} \underline{(0.45-1.71)} \\\rowcolor{tblalt}
SDM & 37.75 (34.94-40.56) & 96.9 (95.7-98.1) & \underline{16.08} \underline{(0.36-31.80)} & 1.50 (0.58-2.42) & 37.90 (35.08-40.72) & 97.0 (95.8-98.1) & 22.01 (7.13-36.89) & 1.65 (0.85-2.45) \\
Flow Matching & 37.79 (34.17-41.41) & 96.9 (95.1-98.8) & 17.15 (1.37-32.93) & 2.11 (0.57-3.65) & \underline{38.45} \underline{(34.76-42.14)} & 97.3 (95.8-98.9) & \underline{21.64} \underline{(4.82-38.46)} & 1.82 (0.77-2.87) \\\rowcolor{tblalt}
CAR & 36.81 (34.11-39.51) & 96.8 (95.5-98.0) & 17.08 (0.39-33.77) & 1.52 (0.58-2.46) & 36.95 (34.18-39.72) & 96.8 (95.7-98.0) & 23.36 (7.55-39.17) & 1.67 (0.85-2.49) \\
\textbf{BrainGFM} & \textbf{40.42} \textbf{(38.27-42.57)} & \textbf{98.3} \textbf{(97.3-99.2)} & \textbf{13.85} \textbf{(1.18-26.52)} & \textbf{1.14} \textbf{(0.46-1.82)} & \textbf{39.83} \textbf{(37.62-42.04)} & \textbf{98.4} \textbf{(97.5-99.2)} & \textbf{18.21} \textbf{(5.06-31.36)} & \textbf{0.97} \textbf{(0.38-1.56)} \\\rowcolor{tblalt}
\hline
\rowcolor{tblhead}
& \multicolumn{4}{c}{LFMR-ABIDE-II} & \multicolumn{4}{c}{LFMR-IBSR-T1} \\
\hhline{~--------}
\rowcolor{tblhead}
Method & PSNR $\uparrow$ & SSIM $\uparrow$ & RMSE $\downarrow$ & LPIPS $\downarrow$ & PSNR $\uparrow$ & SSIM $\uparrow$ & RMSE $\downarrow$ & LPIPS $\downarrow$ \\\rowcolor{tblalt}
\hline
RegGAN & 33.21 (28.88-37.54) & 94.8 (90.9-98.6) & 23.92 (15.84-32.00) & 4.24 (0.43-8.05) & 32.15 (26.93-37.37) & 92.4 (87.5-97.2) & 28.75 (2.09-55.41) & 7.25 (2.13-12.37) \\
CycleGAN & 35.23 (31.46-39.00) & 96.4 (93.8-99.0) & 18.60 (12.88-24.32) & 2.46 (0.34-4.58) & 34.22 (29.37-39.07) & 94.1 (89.9-98.4) & 24.52 (1.52-47.52) & 5.12 (0.87-9.37) \\\rowcolor{tblalt}
MAE & \underline{36.84} \underline{(33.20-40.48)} & \underline{97.3} \underline{(95.3-99.2)} & \underline{15.37} \underline{(10.82-19.92)} & \underline{1.57} \underline{(0.34-2.80)} & 35.85 (31.33-40.37) & 95.4 (91.5-99.2) & 21.25 (1.61-40.89) & 3.25 (0.13-6.37) \\
DDIM & 34.76 (31.23-38.29) & 95.8 (92.9-98.6) & 19.48 (13.79-25.17) & 2.06 (0.27-3.85) & 34.15 (29.40-38.90) & 94.1 (89.7-98.4) & 24.85 (3.00-46.70) & 5.25 (0.90-9.60) \\\rowcolor{tblalt}
SDM & 33.60 (30.21-36.99) & 94.9 (91.8-98.0) & 22.10 (15.79-28.41) & 3.66 (1.24-6.08) & 33.05 (28.10-38.00) & 93.3 (88.7-97.8) & 26.15 (1.69-50.61) & 6.05 (1.40-10.70) \\
Flow Matching & 34.88 (31.07-38.69) & 95.7 (92.7-98.7) & 19.37 (13.56-25.18) & 2.59 (0.79-4.39) & \underline{35.95} \underline{(31.50-40.40)} & \underline{95.5} \underline{(91.7-99.2)} & \underline{21.05} \underline{(2.28-39.82)} & \underline{3.15} \underline{(0.10-6.20)} \\\rowcolor{tblalt}
CAR & 33.54 (30.12-36.96) & 94.9 (91.8-98.1) & 22.15 (15.80-28.50) & 3.68 (1.23-6.13) & 33.12 (28.20-38.04) & 93.2 (88.6-97.8) & 26.22 (1.51-50.93) & 6.12 (1.44-10.80) \\
\textbf{BrainGFM} & \textbf{37.68} \textbf{(35.62-39.74)} & \textbf{97.9} \textbf{(97.1-98.8)} & \textbf{14.12} \textbf{(10.97-17.27)} & \textbf{1.42} \textbf{(0.21-2.63)} & \textbf{37.25} \textbf{(33.40-41.10)} & \textbf{96.9} \textbf{(94.2-99.5)} & \textbf{18.75} \textbf{(2.62-34.88)} & \textbf{2.25} \textbf{(0.13-4.37)} \\\rowcolor{tblalt}
\hline
\rowcolor{tblhead}
& \multicolumn{4}{c}{LFMR-MSD-T1} & \multicolumn{4}{c}{LFMR-MSD-T2} \\
\hhline{~--------}
\rowcolor{tblhead}
Method & PSNR $\uparrow$ & SSIM $\uparrow$ & RMSE $\downarrow$ & LPIPS $\downarrow$ & PSNR $\uparrow$ & SSIM $\uparrow$ & RMSE $\downarrow$ & LPIPS $\downarrow$ \\\rowcolor{tblalt}
\hline
RegGAN & 33.99 (29.93-38.05) & 93.7 (89.7-97.7) & 24.80 (10.72-38.88) & 6.65 (1.74-11.56) & 34.08 (30.19-37.97) & 94.0 (90.5-97.5) & 28.79 (9.67-47.91) & 7.02 (2.45-11.59) \\
CycleGAN & 37.68 (34.45-40.91) & 96.7 (94.7-98.6) & 15.51 (7.66-23.36) & 2.28 (0.41-4.15) & 38.14 (35.08-41.20) & 97.4 (96.0-98.8) & 17.41 (6.83-27.99) & 1.87 (0.66-3.08) \\\rowcolor{tblalt}
MAE & \underline{39.47} \underline{(36.49-42.45)} & \underline{97.7} \underline{(96.4-99.0)} & \textbf{\underline{12.46}} \textbf{\underline{(6.36-18.56)}} & \underline{1.67} \underline{(0.59-2.75)} & \underline{39.21} \underline{(36.11-42.31)} & \underline{97.9} \underline{(96.8-99.0)} & \underline{15.37} \underline{(6.18-24.56)} & \underline{1.51} \underline{(0.70-2.32)} \\
DDIM & 37.63 (34.69-40.57) & 96.5 (94.6-98.4) & 15.47 (7.71-23.23) & 1.71 (0.56-2.86) & 37.41 (34.41-40.41) & 96.8 (95.3-98.4) & 18.88 (7.46-30.30) & 1.71 (0.79-2.63) \\\rowcolor{tblalt}
SDM & 36.05 (32.84-39.26) & 96.1 (93.7-98.5) & 20.10 (3.82-36.38) & 2.35 (0.13-4.57) & 36.88 (33.62-40.14) & 96.9 (94.5-99.4) & 19.18 (1.34-37.02) & 2.48 (1.18-3.78) \\
Flow Matching & 38.31 (35.07-41.55) & 96.9 (95.1-98.6) & 14.39 (7.21-21.57) & 1.74 (0.72-2.76) & 38.16 (34.85-41.47) & 97.2 (95.7-98.7) & 17.45 (6.92-27.98) & 1.68 (0.79-2.57) \\\rowcolor{tblalt}
CAR & 33.12 (30.17-36.07) & 95.8 (93.3-98.2) & 23.95 (4.55-43.35) & 2.38 (0.13-4.63) & 34.95 (31.85-38.05) & 96.7 (94.2-99.2) & 21.60 (2.92-40.28) & 2.52 (1.20-3.84) \\
\textbf{BrainGFM} & \textbf{40.25} \textbf{(38.03-42.47)} & \textbf{98.1} \textbf{(97.1-99.1)} & 14.25 (0.99-27.51) & \textbf{1.18} \textbf{(0.46-1.90)} & \textbf{40.12} \textbf{(37.84-42.40)} & \textbf{98.1} \textbf{(97.0-99.1)} & \textbf{14.65} \textbf{(2.21-27.09)} & \textbf{1.22} \textbf{(0.47-1.97)} \\\rowcolor{tblalt}
\hline
\rowcolor{tblhead}
& \multicolumn{4}{c}{LFMR-IXI-T1} & \multicolumn{4}{c}{LFMR-IXI-T2} \\
\hhline{~--------}
\rowcolor{tblhead}
Method & PSNR $\uparrow$ & SSIM $\uparrow$ & RMSE $\downarrow$ & LPIPS $\downarrow$ & PSNR $\uparrow$ & SSIM $\uparrow$ & RMSE $\downarrow$ & LPIPS $\downarrow$ \\\rowcolor{tblalt}
\hline
RegGAN & 37.01 (32.15-41.87) & 97.42 (95.6-99.3) & 16.34 (1.57-31.11) & 2.21 (0.39-4.03) & 34.41 (28.89-39.93) & 94.1 (90.1-98.1) & 22.99 (4.19-41.79) & 5.49 (1.18-9.80) \\
CycleGAN & 39.07 (34.72-43.42) & 98.3 (97.0-99.5) & 12.47 (0.40-24.54) & 1.05 (0.07-2.03) & 37.32 (32.71-41.93) & 97.0 (95.1-98.8) & 15.56 (3.51-27.61) & 1.81 (0.41-3.21) \\\rowcolor{tblalt}
MAE & 38.41 (33.92-42.90) & 98.0 (96.6-99.4) & 13.50 (2.16-24.84) & 1.06 (0.15-1.97) & 37.34 (32.72-41.96) & 96.9 (94.9-98.8) & 15.72 (3.45-27.99) & 2.09 (0.66-3.52) \\
DDIM & 38.67 (35.08-42.26) & 98.1 (97.0-99.2) & 12.48 (1.05-23.91) & \underline{0.74} \underline{(0.26-1.22)} & 37.04 (33.29-40.79) & 96.6 (94.7-98.5) & 15.92 (3.43-28.41) & 1.56 (0.63-2.49) \\\rowcolor{tblalt}
SDM & 38.02 (34.22-41.82) & 97.4 (96.2-98.5) & 13.12 (0.64-25.60) & 1.10 (0.39-1.81) & 36.99 (32.89-41.09) & 96.7 (95.1-98.4) & 15.31 (4.47-26.15) & 1.76 (0.63-2.89) \\
Flow Matching & \textbf{\underline{39.40}} \textbf{\underline{(35.35-43.45)}} & \textbf{\underline{98.3}} \textbf{\underline{(97.2-99.3)}} & \textbf{\underline{11.52}} \textbf{\underline{(0.30-22.74)}} & 1.24 (0.47-2.01) & \underline{38.48} \underline{(32.97-43.99)} & \underline{97.2} \underline{(95.5-98.9)} & \underline{14.00} \underline{(2.92-25.08)} & \textbf{\underline{1.37}} \textbf{\underline{(0.47-2.27)}} \\\rowcolor{tblalt}
CAR & 38.05 (34.23-41.87) & 97.3 (96.2-98.5) & 13.17 (0.61-25.73) & 1.13 (0.40-1.86) & 37.02 (32.91-41.13) & 96.7 (95.0-98.4) & 15.35 (4.48-26.22) & 1.78 (0.63-2.93) \\
\textbf{BrainGFM} & 39.29 (36.05-42.53) & 98.3 (97.3-99.2) & 12.89 (2.16-23.62) & \textbf{0.71} \textbf{(0.26-1.16)} & \textbf{38.69} \textbf{(35.18-42.20)} & \textbf{97.7} \textbf{(96.4-99.0)} & \textbf{13.92} \textbf{(4.89-22.95)} & 1.42 (0.53-2.31) \\\rowcolor{tblalt}
\hline

\rowcolor{tblhead}
& \multicolumn{4}{c}{LDCT-TopBrain-CT} & \multicolumn{4}{c}{LDCT-CQ500-CT} \\
\hhline{~--------}
\rowcolor{tblhead}
Method & PSNR $\uparrow$ & SSIM $\uparrow$ & RMSE $\downarrow$ & LPIPS $\downarrow$ & PSNR $\uparrow$ & SSIM $\uparrow$ & RMSE $\downarrow$ & LPIPS $\downarrow$ \\\rowcolor{tblalt}
\hline
RegGAN & 38.21 (36.56-39.86) & 97.5 (96.0-98.9) & 11.25 (8.39-14.11) & 1.58 (0.39-2.77) & 40.74 (40.57-40.91) & 98.4 (98.0-98.8) & 19.53 (18.96-20.11) & 0.71 (0.66-0.77) \\
CycleGAN & 36.86 (35.02-38.70) & 97.9 (96.6-99.3) & 14.66 (11.71-17.61) & 2.79 (1.24-4.34) & 38.06 (37.94-38.19) & 98.6 (98.5-98.7) & 26.97 (25.77-28.17) & 1.20 (1.15-1.25) \\\rowcolor{tblalt}
MAE & 39.29 (37.46-41.12) & 99.1 (98.6-99.7) & 11.07 (9.07-13.07) & 0.99 (0.74-1.24) & 38.26 (38.12-38.39) & 98.8 (98.8-98.9) & 25.94 (25.11-26.76) & 1.67 (1.60-1.74) \\
DDIM & 36.87 (35.94-37.80) & 98.5 (98.0-99.0) & 14.42 (12.92-15.92) & 2.89 (2.00-3.78) & 41.81 (41.70-41.92) & 99.4 (99.3-99.4) & 17.26 (16.69-17.83) & 0.63 (0.59-0.66) \\\rowcolor{tblalt}
SDM & 38.15 (36.43-39.87) & 97.3 (95.8-98.8) & 11.18 (8.06-14.30) & 1.65 (0.18-3.12) & 37.44 (37.24-37.63) & 99.1 (99.1-99.2) & 28.82 (27.48-30.16) & 1.32 (1.24-1.40) \\
Flow Matching & \textbf{\underline{50.68}} \textbf{\underline{(48.25-53.11)}} & \underline{99.7} \underline{(99.5-99.9)} & \underline{3.04} \underline{(2.24-3.84)} & \underline{0.19} \underline{(0.02-0.36)} & \underline{50.69} \underline{(50.49-50.89)} & \underline{99.9} \underline{(99.9-99.9)} & \underline{6.24} \underline{(5.95-6.54)} & \textbf{\underline{0.11}} \textbf{\underline{(0.11-0.12)}} \\\rowcolor{tblalt}
CAR & 36.42 (34.75-38.09) & 97.0 (95.4-98.6) & 12.15 (8.50-15.80) & 1.62 (0.37-2.87) & 39.40 (39.25-39.55) & 98.5 (98.3-98.8) & 23.25 (22.37-24.14) & 0.96 (0.90-1.01) \\
\textbf{BrainGFM} & 50.35 (48.17-52.53) & \textbf{99.7} \textbf{(99.6-99.9)} & \textbf{2.15} \textbf{(1.19-3.11)} & \textbf{0.12} \textbf{(0.04-0.20)} & \textbf{51.28} \textbf{(51.04-51.52)} & \textbf{99.9} \textbf{(99.9-99.9)} & \textbf{5.84} \textbf{(5.54-6.13)} & 0.21 (0.19-0.23) \\\rowcolor{tblalt}
\hline
\rowcolor{tblhead}
& \multicolumn{4}{c}{LDCT-INSTANCE22-CT} & \multicolumn{4}{c}{LDCT-BHSD-CT} \\
\hhline{~--------}
\rowcolor{tblhead}
Method & PSNR $\uparrow$ & SSIM $\uparrow$ & RMSE $\downarrow$ & LPIPS $\downarrow$ & PSNR $\uparrow$ & SSIM $\uparrow$ & RMSE $\downarrow$ & LPIPS $\downarrow$ \\\rowcolor{tblalt}
\hline
RegGAN & 40.13 (39.55-40.71) & 98.9 (98.6-99.2) & 21.58 (19.24-23.92) & 0.79 (0.67-0.91) & 40.54 (39.91-41.17) & 98.8 (98.3-99.2) & 23.00 (18.60-27.40) & 1.07 (0.55-1.59) \\
CycleGAN & 35.68 (35.14-36.22) & 93.7 (90.9-96.4) & 35.95 (32.26-39.64) & 1.93 (1.64-2.22) & 38.40 (37.81-38.99) & 97.8 (97.2-98.3) & 29.47 (23.77-35.17) & 1.58 (0.97-2.19) \\\rowcolor{tblalt}
MAE & 38.31 (37.71-38.91) & 98.8 (98.6-99.0) & 26.58 (23.87-29.29) & 1.62 (1.37-1.87) & 38.89 (38.39-39.39) & 98.9 (98.8-99.0) & 26.13 (24.12-28.14) & 1.74 (1.47-2.01) \\
DDIM & 39.24 (38.76-39.72) & 98.5 (98.1-98.8) & 23.81 (21.49-26.13) & 1.34 (1.04-1.64) & 43.45 (42.74-44.16) & 99.2 (98.9-99.5) & 17.15 (12.33-21.97) & 0.74 (0.05-1.43) \\\rowcolor{tblalt}
SDM & 37.31 (36.72-37.90) & 99.0 (98.6-99.3) & 30.03 (26.61-33.45) & 1.38 (1.22-1.54) & 37.12 (35.64-38.60) & 97.5 (95.3-99.7) & 50.12 (14.99-85.25) & 1.74 (1.08-2.40) \\
Flow Matching & \underline{48.96} \underline{(48.36-49.56)} & \underline{99.6} \underline{(99.5-99.6)} & \underline{7.58} \underline{(6.88-8.28)} & \underline{0.24} \underline{(0.18-0.30)} & \underline{50.11} \underline{(49.20-51.02)} & \textbf{\underline{99.8}} \textbf{\underline{(99.7-99.9)}} & \underline{8.26} \underline{(5.61-10.91)} & \textbf{\underline{0.27}} \textbf{\underline{(0.11-0.43)}} \\\rowcolor{tblalt}
CAR & 38.16 (37.57-38.75) & 97.5 (97.0-98.0) & 29.36 (26.41-32.31) & 1.41 (1.20-1.62) & 39.08 (38.43-39.73) & 98.3 (97.9-98.8) & 25.62 (22.05-29.19) & 1.44 (0.94-1.94) \\
\textbf{BrainGFM} & \textbf{49.32} \textbf{(48.72-49.92)} & \textbf{99.7} \textbf{(99.6-99.7)} & \textbf{7.16} \textbf{(6.46-7.86)} & \textbf{0.20} \textbf{(0.14-0.26)} & \textbf{50.32} \textbf{(49.41-51.23)} & 99.8 (99.7-99.8) & \textbf{8.03} \textbf{(5.38-10.68)} & 0.31 (0.03-0.59) \\\rowcolor{tblalt}
\hline
\rowcolor{tblhead}
& \multicolumn{4}{c}{NCMR-BraTS21-T1,Tce} & \multicolumn{4}{c}{NCMR-TCGA-T1,T1ce} \\
\hhline{~--------}
\rowcolor{tblhead}
Method & PSNR $\uparrow$ & SSIM $\uparrow$ & RMSE $\downarrow$ & LPIPS $\downarrow$ & PSNR $\uparrow$ & SSIM $\uparrow$ & RMSE $\downarrow$ & LPIPS $\downarrow$ \\\rowcolor{tblalt}
\hline
RegGAN & 26.22 (23.89-28.55) & 90.7 (86.4-95.1) & 44.03 (8.61-79.45) & 6.92 (2.71-11.13) & 27.34 (25.14-29.54) & 91.5 (87.6-95.4) & 45.03 (8.40-81.66) & 7.18 (2.56-11.80) \\
CycleGAN & 28.33 (26.12-30.54) & 95.3 (93.2-97.4) & 26.40 (7.11-45.69) & 2.16 (1.05-3.27) & 28.72 (26.21-31.23) & 95.8 (93.7-98.0) & 24.19 (5.24-43.14) & 2.04 (0.96-3.12) \\\rowcolor{tblalt}
MAE & 29.42 (27.17-31.67) & 95.8 (94.0-97.7) & 24.27 (6.12-42.42) & 2.31 (1.25-3.37) & 27.84 (25.21-30.47) & 95.7 (93.9-97.6) & 24.68 (5.94-43.42) & 2.24 (1.12-3.36) \\
DDIM & 30.12 (27.77-32.47) & 95.2 (93.1-97.4) & 24.72 (6.76-42.68) & 1.71 (0.89-2.53) & 30.31 (27.95-32.67) & 95.2 (92.9-97.4) & 25.12 (5.51-44.73) & \underline{1.66} \underline{(0.87-2.45)} \\\rowcolor{tblalt}
SDM & 27.26 (25.24-29.28) & 94.8 (93.1-96.4) & 26.86 (7.27-46.45) & 1.83 (0.95-2.71) & 30.46 (28.15-32.77) & 95.4 (93.1-97.8) & 24.07 (5.11-43.03) & 1.76 (0.92-2.60) \\
Flow Matching & \underline{31.07} \underline{(28.65-33.49)} & \underline{96.0} \underline{(94.1-97.9)} & \underline{22.26} \underline{(5.35-39.17)} & \underline{1.62} \underline{(0.86-2.38)} & \underline{30.96} \underline{(28.72-33.20)} & \underline{96.0} \underline{(94.1-98.0)} & \underline{23.94} \underline{(7.15-40.73)} & 1.85 (0.94-2.76) \\\rowcolor{tblalt}
CAR & 26.28 (24.34-28.22) & 94.5 (92.8-96.2) & 28.51 (7.72-49.30) & 1.84 (0.94-2.74) & 27.31 (24.85-29.77) & 94.6 (92.0-97.2) & 27.73 (5.74-49.72) & 1.69 (0.83-2.55) \\
\textbf{BrainGFM} & \textbf{31.42} \textbf{(29.28-33.56)} & \textbf{97.2} \textbf{(95.9-98.5)} & \textbf{19.12} \textbf{(5.24-33.00)} & \textbf{1.36} \textbf{(0.64-2.08)} & \textbf{31.09} \textbf{(28.82-33.36)} & \textbf{96.5} \textbf{(95.1-98.0)} & \textbf{20.03} \textbf{(5.27-34.79)} & \textbf{1.49} \textbf{(0.69-2.29)} \\\rowcolor{tblalt}
\hline
\end{tabular}
}
\end{table*}

\begin{table*}[!t]
\noindent\textit{Tab.~\ref{tab:imaging_multidataset} continued}\par
\centering
\resizebox{\textwidth}{!}{
\begin{tabular}{lcccccccc}
\hline
\rowcolor{tblhead}
& \multicolumn{4}{c}{NCMR-TCIA-T1,T1ce} & \multicolumn{4}{c}{NCMR-MSDBrain-T1,T1ce} \\
\hhline{~--------}
\rowcolor{tblhead}
Method & PSNR $\uparrow$ & SSIM $\uparrow$ & RMSE $\downarrow$ & LPIPS $\downarrow$ & PSNR $\uparrow$ & SSIM $\uparrow$ & RMSE $\downarrow$ & LPIPS $\downarrow$ \\\rowcolor{tblalt}
\hline
RegGAN & 26.91 (24.56-29.26) & 91.1 (86.6-95.7) & 46.07 (7.72-84.42) & 7.42 (2.51-12.33) & 25.98 (23.16-28.80) & 90.9 (86.2-95.6) & 42.58 (8.14-77.02) & 7.61 (2.53-12.69) \\
CycleGAN & 28.14 (25.67-30.61) & 95.3 (92.9-97.7) & 25.46 (5.62-45.30) & 2.29 (1.12-3.46) & 28.79 (26.20-31.38) & 94.8 (92.4-97.2) & 25.51 (6.36-44.66) & 2.35 (1.11-3.59) \\\rowcolor{tblalt}
MAE & 26.56 (23.98-29.14) & \underline{95.6} \underline{(93.9-97.3)} & 25.53 (6.17-44.89) & \underline{1.74} \underline{(0.92-2.56)} & \underline{29.83} \underline{(27.27-32.39)} & \underline{95.4} \underline{(93.4-97.3)} & \textbf{\underline{21.22}} \textbf{\underline{(5.31-37.13)}} & 2.57 (1.31-3.83) \\
DDIM & 29.84 (27.43-32.25) & 94.8 (92.3-97.4) & 25.91 (5.73-46.09) & 1.81 (0.94-2.68) & 28.51 (25.97-31.05) & 94.5 (92.0-96.9) & 26.61 (6.67-46.55) & 1.88 (0.97-2.79) \\\rowcolor{tblalt}
SDM & 30.18 (27.81-32.55) & 95.1 (92.8-97.4) & \underline{24.76} \underline{(5.43-44.09)} & 1.91 (0.98-2.84) & 27.75 (25.13-30.37) & 94.6 (91.4-97.7) & 30.96 (4.38-57.54) & 1.97 (0.96-2.98) \\
Flow Matching & \underline{30.34} \underline{(28.03-32.65)} & 94.9 (92.8-97.1) & 25.03 (7.57-42.49) & 2.05 (1.03-3.07) & 29.03 (26.40-31.66) & 94.9 (92.6-97.2) & 24.55 (6.14-42.96) & \underline{1.81} \underline{(0.93-2.69)} \\\rowcolor{tblalt}
CAR & 27.88 (25.26-30.50) & 94.2 (91.5-96.9) & 26.82 (5.45-48.19) & 1.86 (0.95-2.77) & 27.80 (25.18-30.42) & 94.5 (91.4-97.7) & 31.06 (3.07-59.05) & 1.94 (1.00-2.88) \\
\textbf{BrainGFM} & \textbf{30.72} \textbf{(28.61-32.83)} & \textbf{96.7} \textbf{(95.4-98.1)} & \textbf{20.42} \textbf{(6.10-34.74)} & \textbf{1.45} \textbf{(0.70-2.20)} & \textbf{30.44} \textbf{(28.10-32.78)} & \textbf{95.9} \textbf{(94.2-97.6)} & 22.25 (4.36-40.14) & \textbf{1.66} \textbf{(0.80-2.52)} \\\rowcolor{tblalt}
\hline
\rowcolor{tblhead}
& \multicolumn{4}{c}{NCMR-BraTS-METS-T1,T1ce} & \multicolumn{4}{c}{NCMR-BraTS-PEDs-T1,T1ce} \\
\hhline{~--------}
\rowcolor{tblhead}
Method & PSNR $\uparrow$ & SSIM $\uparrow$ & RMSE $\downarrow$ & LPIPS $\downarrow$ & PSNR $\uparrow$ & SSIM $\uparrow$ & RMSE $\downarrow$ & LPIPS $\downarrow$ \\\rowcolor{tblalt}
\hline
RegGAN & 25.74 (23.31-28.17) & 90.4 (85.3-95.5) & 50.59 (8.78-92.40) & 7.84 (2.55-13.13) & 25.31 (22.81-27.81) & 90.1 (84.8-95.2) & 51.03 (8.37-93.69) & 8.17 (2.64-13.70) \\
CycleGAN & 28.73 (26.03-31.43) & 94.4 (91.6-97.2) & 25.56 (5.62-45.50) & 2.53 (1.22-3.84) & 28.26 (25.36-31.16) & 94.0 (91.0-97.0) & 26.01 (5.47-46.55) & 2.61 (1.23-3.99) \\\rowcolor{tblalt}
MAE & 28.81 (25.58-32.04) & \underline{95.0} \underline{(92.7-97.3)} & \underline{22.74} \underline{(4.99-40.49)} & 2.76 (1.43-4.09) & \underline{28.31} \underline{(25.02-31.60)} & \underline{95.1} \underline{(92.9-97.3)} & \underline{23.46} \underline{(5.08-41.84)} & 2.83 (1.42-4.24) \\
DDIM & 28.37 (25.88-30.86) & 93.8 (90.8-96.8) & 29.28 (5.97-52.59) & \underline{2.02} \underline{(1.06-2.98)} & 27.87 (25.36-30.38) & 93.1 (89.8-96.3) & 30.11 (5.80-54.42) & 2.13 (1.11-3.15) \\\rowcolor{tblalt}
SDM & 28.59 (26.11-31.07) & 93.9 (91.1-96.8) & 27.55 (5.28-49.82) & 2.16 (1.09-3.23) & 28.08 (25.58-30.58) & 93.4 (90.4-96.5) & 28.96 (5.64-52.28) & 2.24 (1.13-3.35) \\
Flow Matching & \underline{28.82} \underline{(26.45-31.19)} & 93.9 (91.4-96.3) & 29.24 (7.87-50.61) & 2.31 (1.18-3.44) & \underline{28.31} \underline{(25.88-30.74)} & 93.8 (91.3-96.3) & 30.11 (8.08-52.14) & \underline{2.05} \underline{(1.07-3.03)} \\\rowcolor{tblalt}
CAR & 26.63 (23.94-29.32) & 93.4 (90.3-96.5) & 29.26 (6.10-52.42) & 2.07 (1.06-3.08) & 27.13 (24.39-29.87) & 93.3 (90.2-96.5) & 28.31 (5.68-50.94) & 2.19 (1.13-3.25) \\
\textbf{BrainGFM} & \textbf{30.22} \textbf{(27.71-32.73)} & \textbf{95.5} \textbf{(93.6-97.3)} & \textbf{21.98} \textbf{(5.68-38.28)} & \textbf{1.79} \textbf{(0.87-2.71)} & \textbf{29.68} \textbf{(27.23-32.13)} & \textbf{95.7} \textbf{(93.9-97.4)} & \textbf{22.79} \textbf{(5.79-39.79)} & \textbf{1.74} \textbf{(0.86-2.62)} \\\rowcolor{tblalt}
\hline
\rowcolor{tblhead}
& \multicolumn{4}{c}{NCMR-UPenn-GBM-T1,T1ce} & \multicolumn{4}{c}{NCMR-Ivy GAP-T1,T1ce} \\
\hhline{~--------}
\rowcolor{tblhead}
Method & PSNR $\uparrow$ & SSIM $\uparrow$ & RMSE $\downarrow$ & LPIPS $\downarrow$ & PSNR $\uparrow$ & SSIM $\uparrow$ & RMSE $\downarrow$ & LPIPS $\downarrow$ \\\rowcolor{tblalt}
\hline
RegGAN & 26.96 (24.10-29.82) & 90.2 (85.0-95.4) & 47.32 (7.84-86.80) & 8.49 (2.71-14.27) & 26.38 (23.59-29.17) & 90.6 (85.5-95.7) & 46.85 (8.09-85.61) & 8.63 (2.72-14.54) \\
CycleGAN & 26.54 (23.63-29.45) & 93.6 (90.5-96.7) & 29.04 (5.98-52.10) & \textbf{\underline{2.24}} \textbf{\underline{(1.17-3.31)}} & 26.07 (23.23-28.91) & \underline{93.6} \underline{(90.6-96.6)} & \underline{28.72} \underline{(5.89-51.55)} & 2.69 (1.30-4.08) \\\rowcolor{tblalt}
MAE & 24.76 (21.71-27.81) & \underline{93.8} \underline{(91.0-96.6)} & 29.22 (5.75-52.69) & 3.04 (1.54-4.54) & 24.37 (21.39-27.35) & 93.3 (90.4-96.2) & 30.08 (6.06-54.10) & 3.18 (1.71-4.65) \\
DDIM & 28.18 (25.45-30.91) & 92.8 (89.4-96.1) & 29.53 (5.81-53.25) & 2.29 (1.20-3.38) & 28.41 (25.73-31.09) & 92.5 (89.2-95.8) & 29.21 (5.83-52.59) & 2.21 (1.17-3.25) \\\rowcolor{tblalt}
SDM & 28.61 (25.92-31.30) & 93.2 (90.0-96.4) & \underline{28.56} \underline{(5.64-51.48)} & 2.34 (1.18-3.50) & 28.29 (25.65-30.93) & 93.0 (90.0-96.1) & 29.48 (5.91-53.05) & \textbf{\underline{2.18}} \textbf{\underline{(1.15-3.21)}} \\
Flow Matching & \textbf{\underline{29.37}} \textbf{\underline{(26.74-32.00)}} & 93.0 (90.4-95.7) & 28.69 (7.55-49.83) & 2.51 (1.26-3.76) & \underline{28.74} \underline{(26.15-31.33)} & 92.74 (90.0-95.5) & 29.82 (7.91-51.73) & 2.44 (1.22-3.66) \\\rowcolor{tblalt}
CAR & 25.74 (22.91-28.57) & 92.8 (89.4-96.1) & 30.61 (6.35-54.87) & 2.42 (1.30-3.54) & 25.48 (22.71-28.25) & 92.4 (88.9-95.8) & 31.18 (6.13-56.23) & 2.39 (1.27-3.51) \\
\textbf{BrainGFM} & 29.33 (26.71-31.95) & \textbf{94.6} \textbf{(92.4-96.7)} & \textbf{23.82} \textbf{(5.34-42.30)} & 2.31 (1.26-3.36) & \textbf{29.21} \textbf{(26.63-31.79)} & \textbf{94.0} \textbf{(91.7-96.3)} & \textbf{24.11} \textbf{(5.15-43.07)} & 2.27 (1.28-3.26) \\\rowcolor{tblalt}
\hline
\end{tabular}%
}
\end{table*}


\begin{figure*}
    \centering
    \includegraphics[width=1\linewidth]{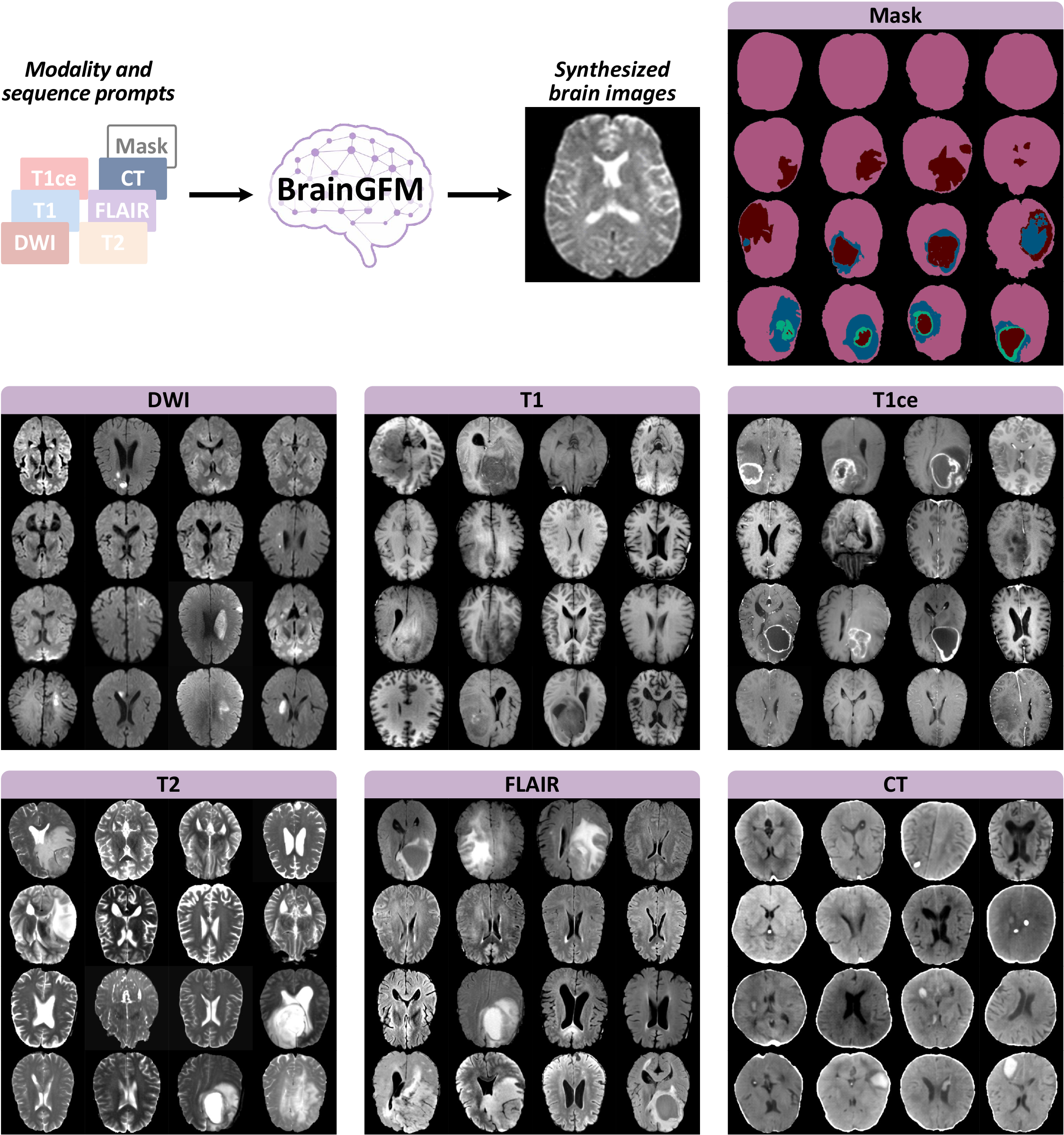}
    \caption{\textbf{Modality- and sequence-conditioned brain image and segmentation mask generation by BrainGFM.} Top, schematic illustration of prompt-guided generation, in which modality, MRI sequence, and mask prompts guide BrainGFM to synthesize brain images and segmentation masks. Representative generated segmentation masks are displayed in pseudocolor to distinguish different anatomical or pathological regions. Bottom, representative synthetic brain images generated under six imaging conditions: DWI, T1, T1ce, T2, FLAIR, and CT. The examples illustrate the diversity of anatomical structures, tissue contrast, and imaging appearances captured by BrainGFM across heterogeneous modalities and sequences.}
    \label{fig:visual_multimodal_syn}
\end{figure*}

\begin{figure*}
    \centering
    \includegraphics[width=1\linewidth]{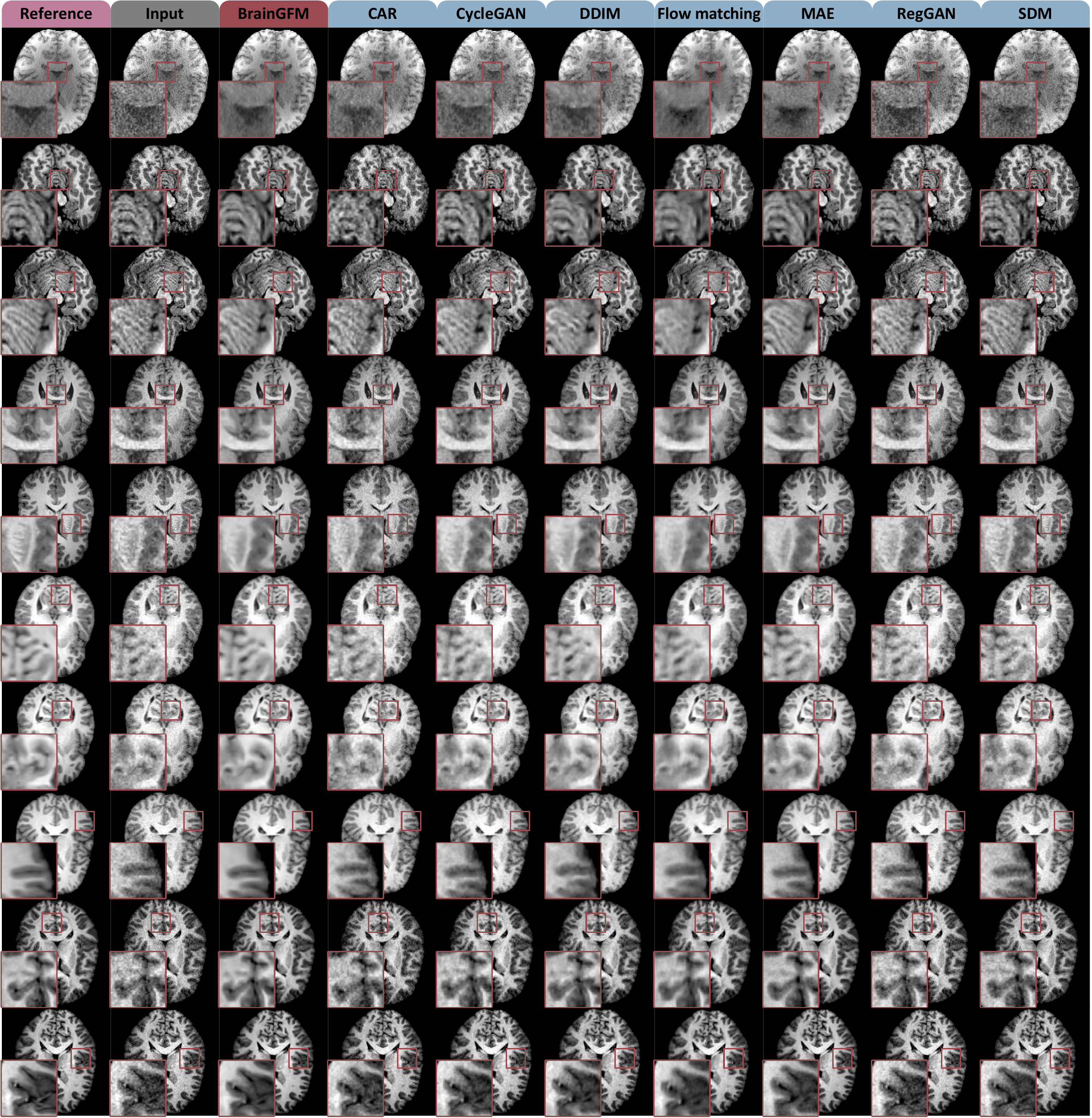}
    \caption{\textbf{Qualitative comparison of low-field brain MRI enhancement across different methods.}}
    \label{fig:visual_LFMR}
\end{figure*}

\begin{figure*}
    \centering
    \includegraphics[width=1\linewidth]{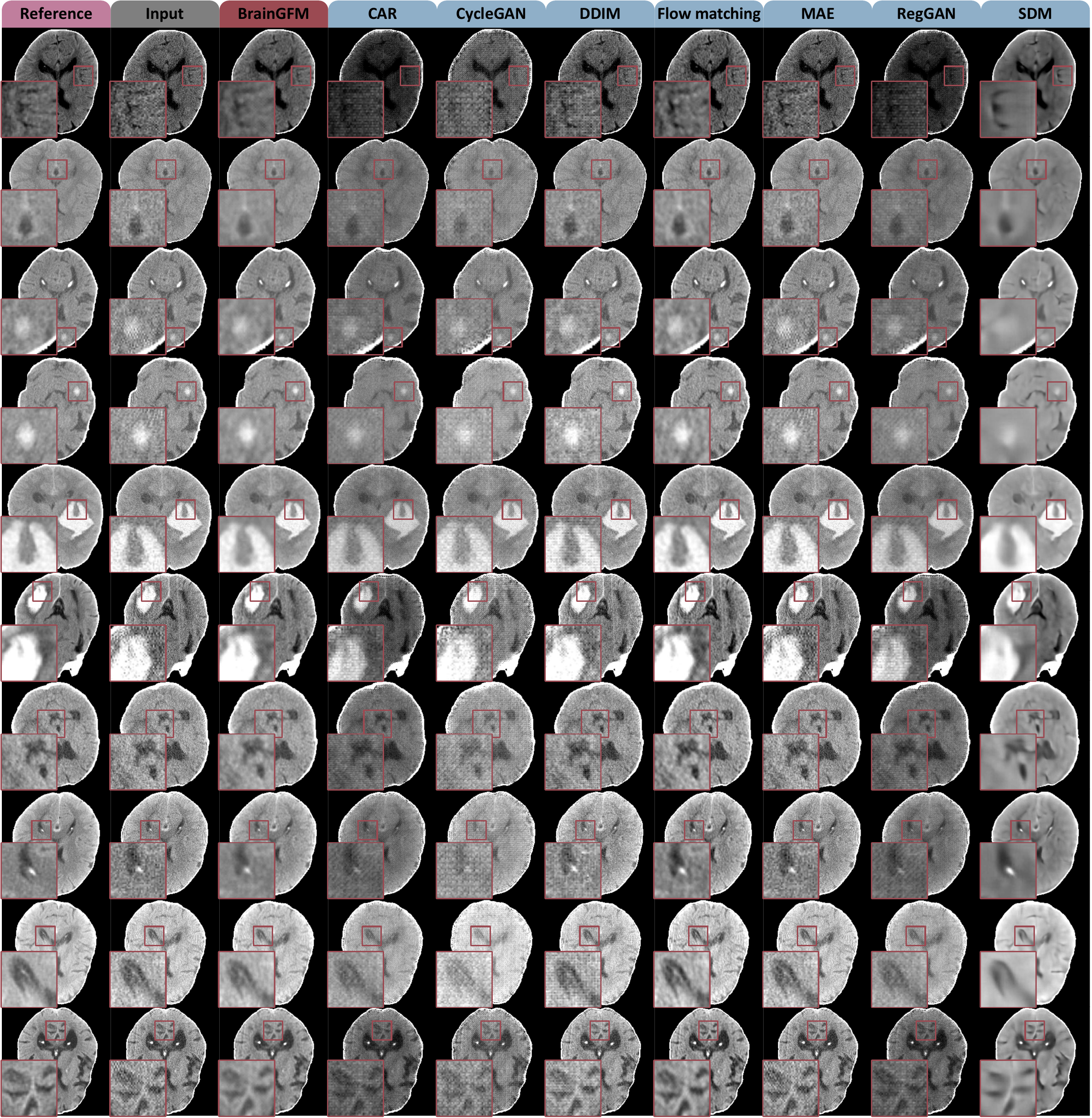}
    \caption{\textbf{Qualitative comparison of low-dose brain CT enhancement across different methods.}}
    \label{fig:visual_LDCT}
\end{figure*}

\begin{figure*}
    \centering
    \includegraphics[width=1\linewidth]{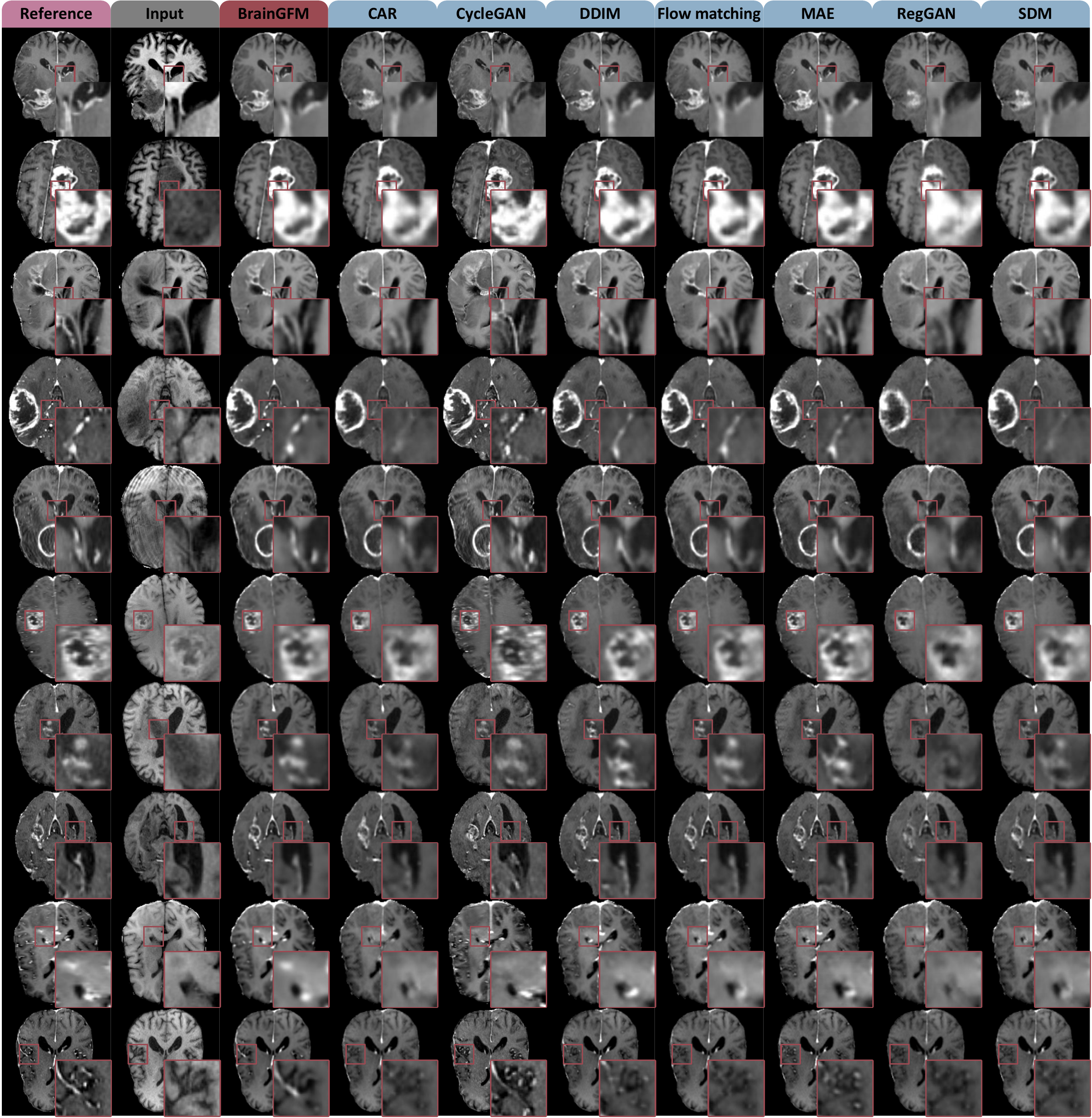}
    \caption{\textbf{Qualitative comparison of non-contrast brain MRI enhancement across different methods.}}
    \label{fig:visual_NCMR}
\end{figure*}

\begin{figure*}
    \centering
    \includegraphics[width=1\linewidth]{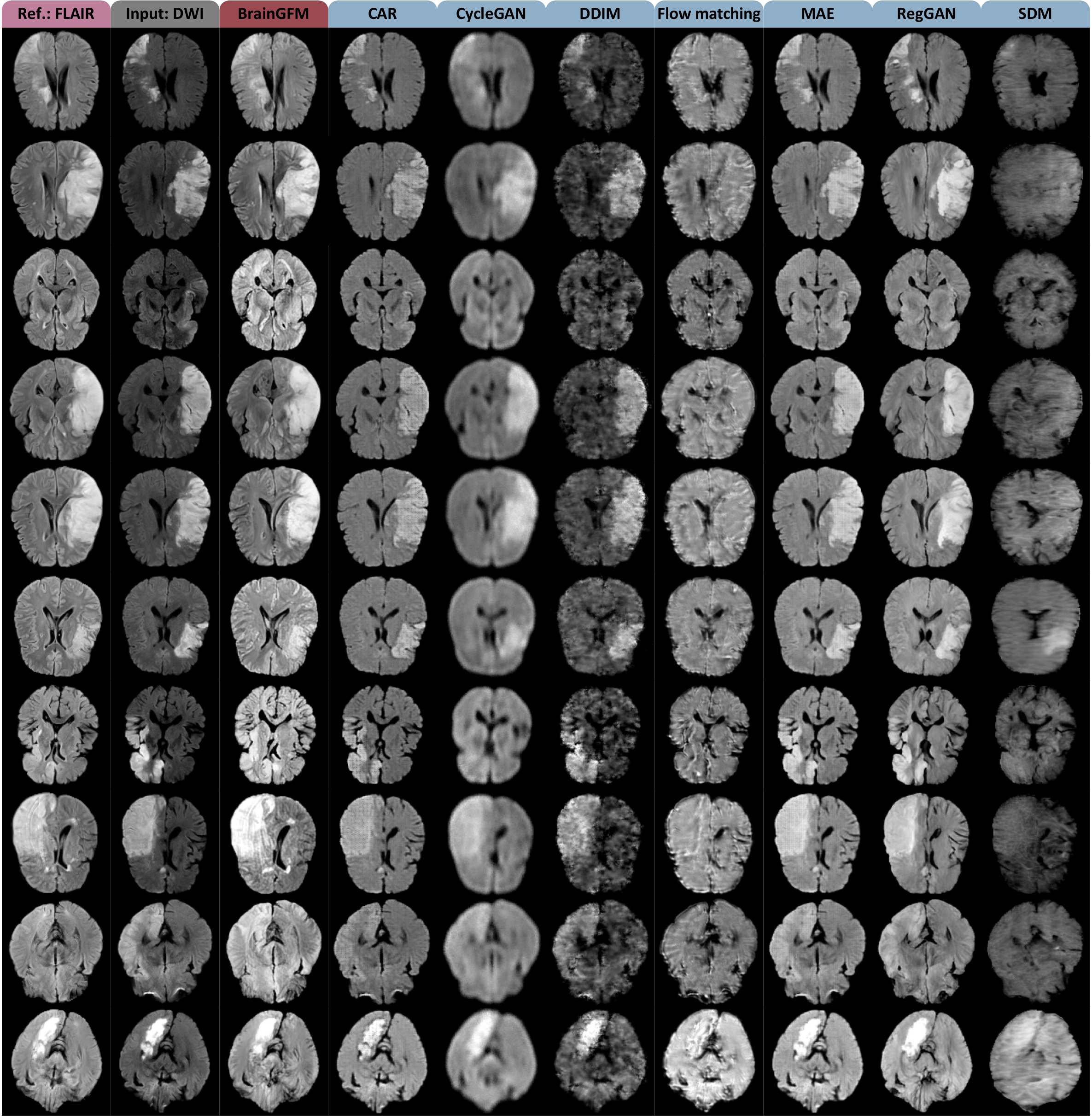}
    \caption{\textbf{Qualitative comparison of DWI-to-FLAIR brain MRI transformation across different methods.}}
    \label{fig:visual_NCMR}
\end{figure*}

\begin{figure*}
    \centering
    \includegraphics[width=1\linewidth]{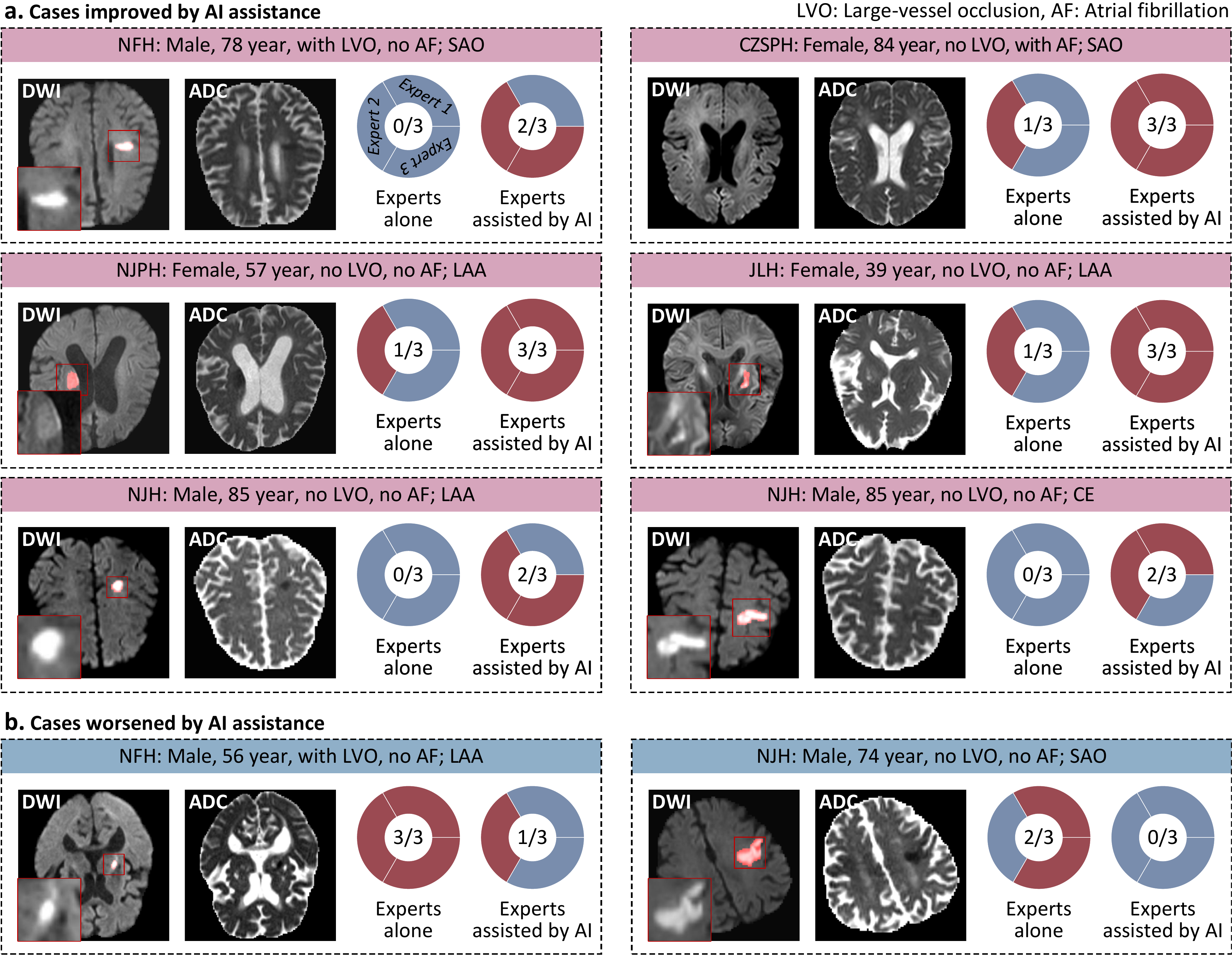}
    \caption{\textbf{Representative cases illustrating diagnostic improvement and deterioration following BrainGFM assistance.} \textbf{a.} Six ischemic stroke cases in which BrainGFM assistance improved radiologists' stroke-subtyping accuracy. \textbf{b.} Two cases in which diagnostic accuracy deteriorated following AI assistance. For each case, DWI and ADC maps are shown alongside patient age, sex, large-vessel occlusion status, and atrial fibrillation status. Donut charts show the proportion of the three radiologists who correctly identified the stroke subtype before (experts alone) and after (experts assisted by AI) BrainGFM assistance, with red and blue sectors indicating correct and incorrect diagnoses. These examples illustrate the potential diagnostic benefits of BrainGFM assistance and the risk of deterioration in initially correct clinical judgments.}
    \label{fig:human_AI_examples}
\end{figure*}

\end{document}